\pdfoutput=1
\PassOptionsToPackage{table,dvipsnames}{xcolor}
\documentclass[11pt]{article}

\usepackage[final]{acl}

\usepackage{times}
\usepackage{latexsym}

\usepackage[T1]{fontenc}

\usepackage[utf8]{inputenc}

\usepackage{microtype}

\usepackage{inconsolata}

\usepackage{graphicx}
\usepackage{soul}
\usepackage{bm} 
\usepackage{amsmath} 
\usepackage{amssymb}
\usepackage{booktabs}
\usepackage{multirow}

\usepackage{pgfplots}
\pgfplotsset{compat=newest}
\definecolor{tiffanyblue}{RGB}{129,216,208}
\definecolor{bangdiblue}{RGB}{0,149,182}
\definecolor{kleinblue}{RGB}{0,47,167}
\definecolor{purple}{RGB}{138,43,226}
\usepackage{amssymb}
\usepackage{pgfplotstable}
\usetikzlibrary{shapes}
\usepackage{circledtext}
\usetikzlibrary{arrows,decorations.pathmorphing,backgrounds,positioning,fit,petri}
\usetikzlibrary{arrows.meta,fit,shapes.arrows}
\usepackage{caption} 
\usepackage{xcolor}
\usepackage{pgfplots}
\usepackage{tikz}
\usepackage{subfig}
\usetikzlibrary{positioning,shapes,arrows,shadows,patterns}
\usetikzlibrary{calc}
\usepackage{pgfplots}
\usepackage{makecell}
\usepackage{CJKutf8}
\usepackage{tabularx}
\usepackage{xcolor}
\usepackage[table]{xcolor} 
\usepackage{xspace}
\usepackage{CJKutf8}  
\usepackage[super]{nth}
\newcommand{\ours}{\textsc{Plast}\xspace}
\newcommand{\headercolorSLAM}{\rowcolor[RGB]{221, 232, 250}}

\newcommand{\eg}{\emph{e.g.}}

\definecolor{mycolor_green}{HTML}{D5E8D4}
\definecolor{mycolor_red}{HTML}{F8CECC}
\def\mylogo{\resizebox{0.45cm}{!}{\includegraphics{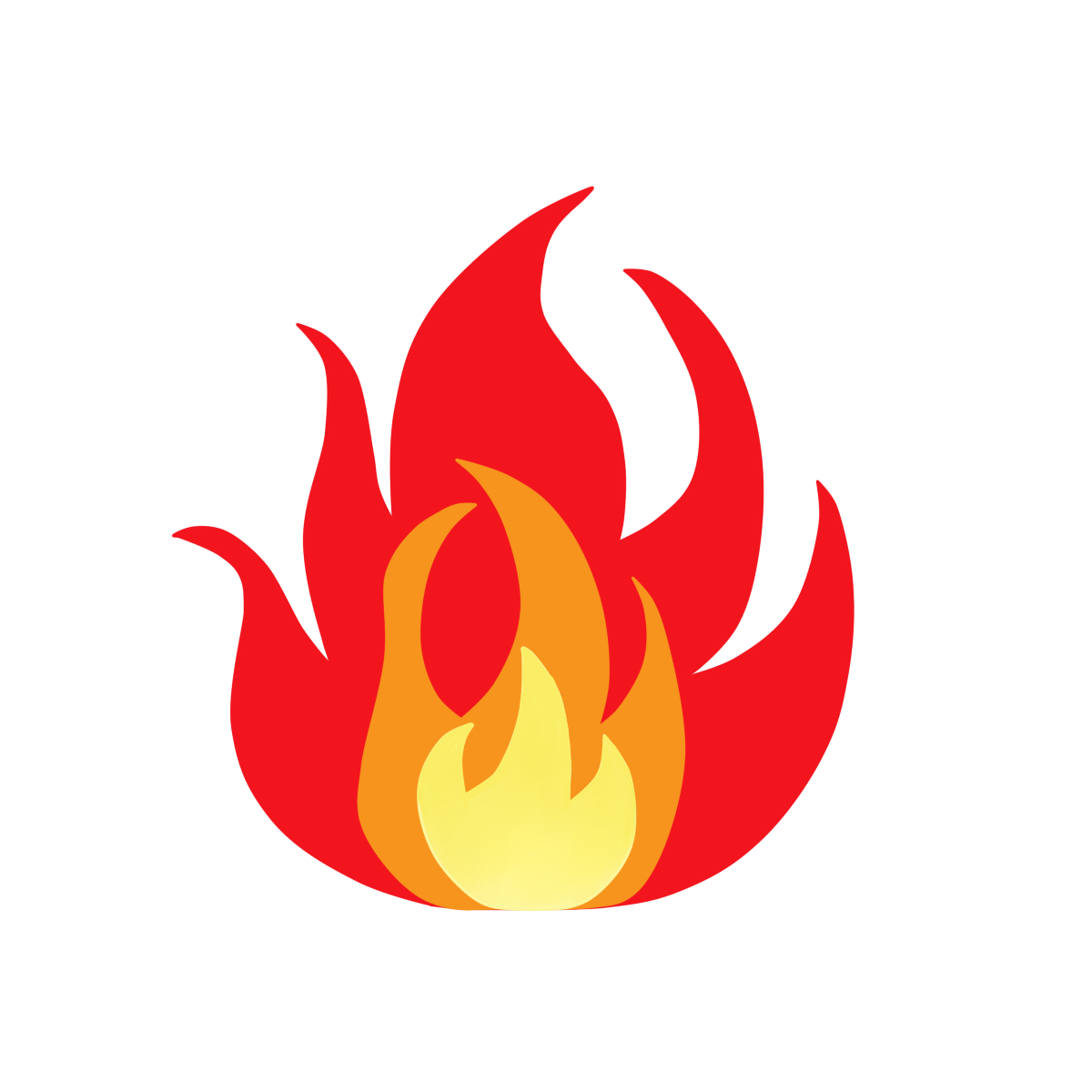}}} 
\def\mylogosnow{\resizebox{0.45cm}{!}{\includegraphics{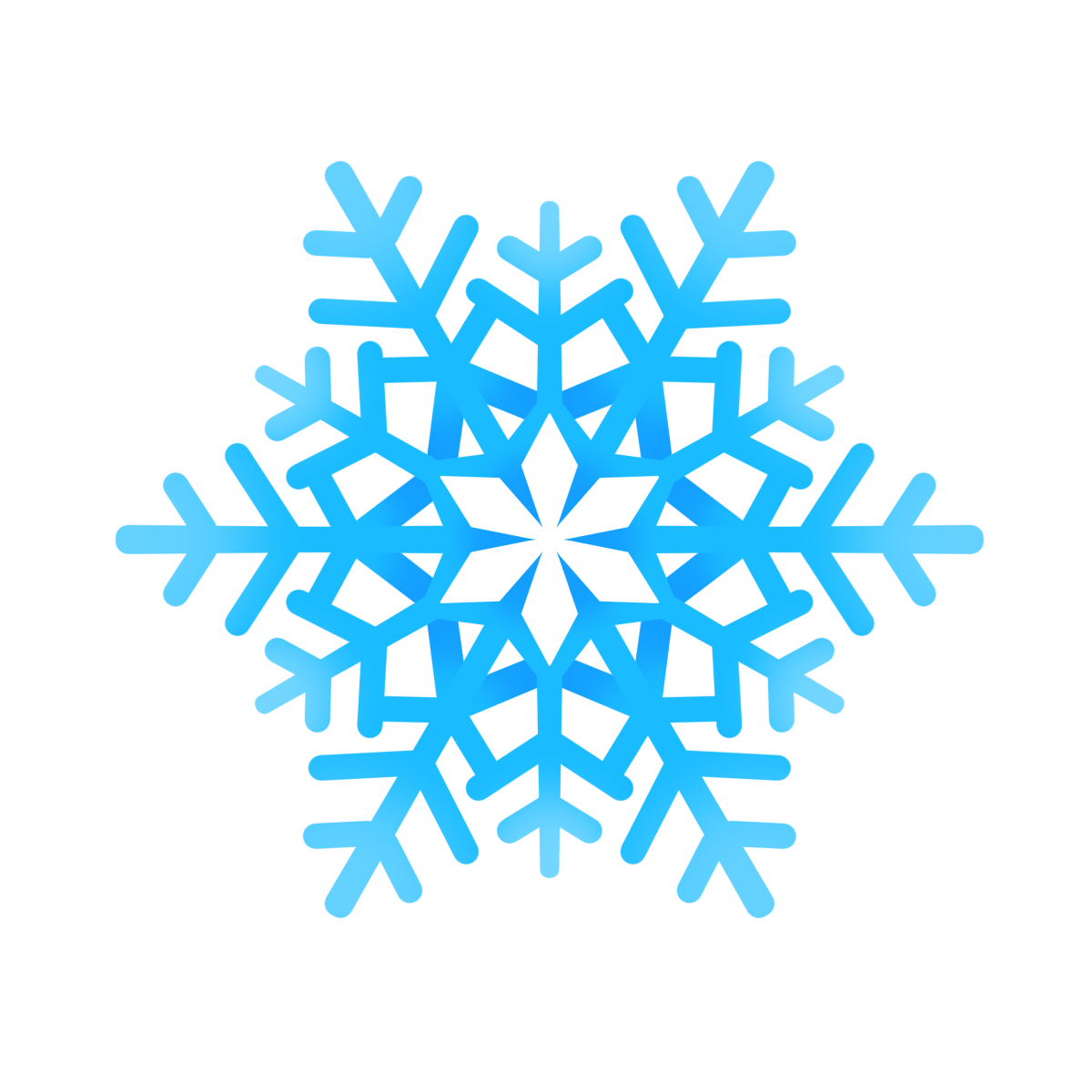}}} 
\def\mylogotranslate{\resizebox{0.45cm}{!}{\includegraphics{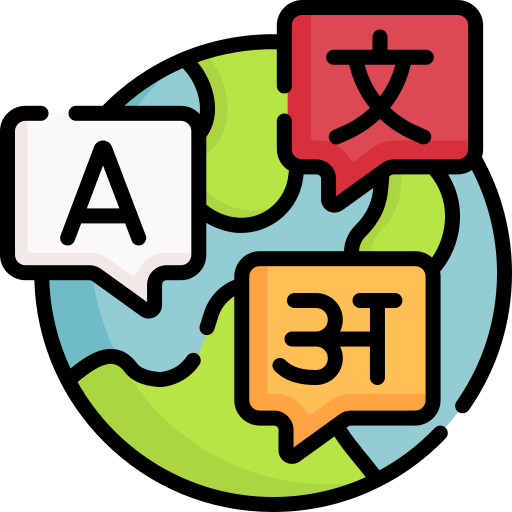}}} 

\def\mylogogear{\resizebox{0.6cm}{!}{\includegraphics{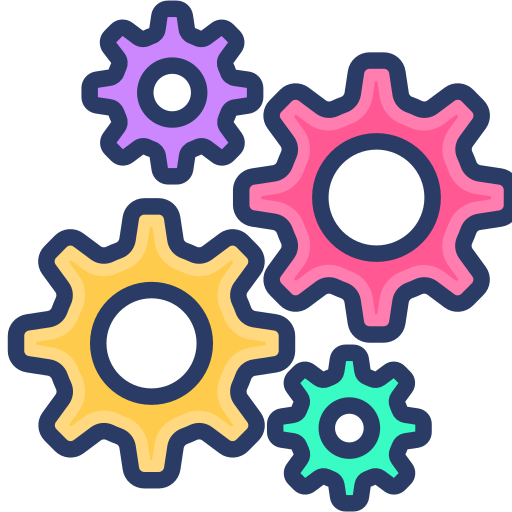}}} 

\def\mylogoidea{\resizebox{0.35cm}{!}{\includegraphics{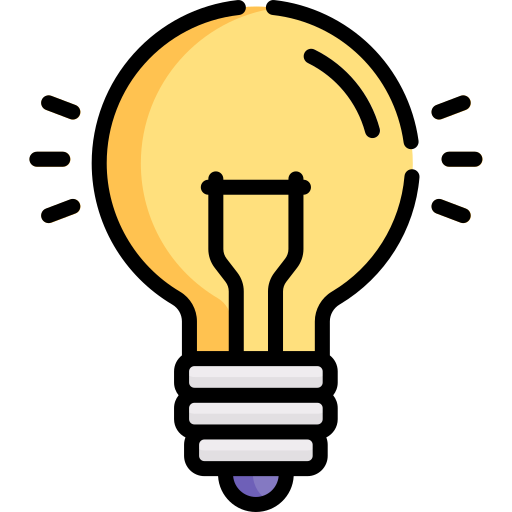}}} 

\def\mylogofig{\resizebox{1.6cm}{!}{\includegraphics{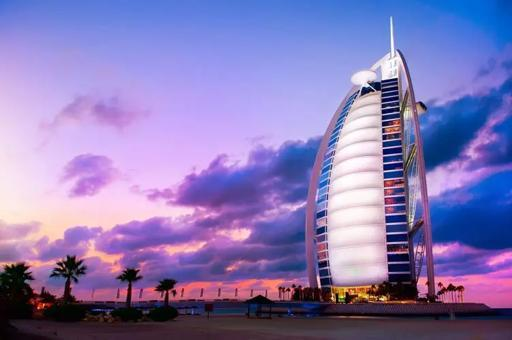}}} 

\usepackage{tikz}
\newcommand{\texthidden}[2]{ 
    \draw[thick, fill=blue!30] 
        (#1, #2) -- (#1+0.10, #2+0.2) -- (#1+0.35, #2+0.2) -- 
        (#1+0.45, #2) -- (#1+0.45, #2-0.24) -- (#1, #2-0.24) -- cycle;
}
\newcommand{\visionhidden}[2]{ 
    \draw[thick, fill=rou] 
        (#1, #2) -- (#1+0.10, #2+0.2) -- (#1+0.35, #2+0.2) -- 
        (#1+0.45, #2) -- (#1+0.45, #2-0.24) -- (#1, #2-0.24) -- cycle;
}
\newcommand{\textoutput}[2]{ 
    \draw[thick, fill=emb-inside] 
        (#1, #2) -- (#1+0.10, #2+0.2) -- (#1+0.35, #2+0.2) -- 
        (#1+0.45, #2) -- (#1+0.45, #2-0.24) -- (#1, #2-0.24) -- cycle;
}

\newcommand{\inlinecolorboxone}{
    \tikz[baseline=-0.1cm]{
        \node[draw=black,thick,fill=emb-inside, inner sep=0.1cm, rounded corners=2pt, minimum width=0.4cm,minimum height=0.4cm] (char1) {};
    }
}

\title{Language-Specific Layer Matters: Efficient Multilingual Enhancement for Large Vision-Language Models}

\author{Yuchun Fan\textsuperscript{1}, Yilin Wang\textsuperscript{1}, Yongyu Mu\textsuperscript{1},  Lei Huang\textsuperscript{4}, Bei Li\textsuperscript{3}, Xiaocheng Feng\textsuperscript{4},\\
{\bf Tong Xiao\textsuperscript{1,2}\thanks{\xspace\xspace Corresponding author.},  \and Jingbo Zhu\textsuperscript{1,2}} \\
	\textsuperscript{1}NLP Lab, School of Computer Science and Engineering, Northeastern University, Shenyang, China\\
	\textsuperscript{2}NiuTrans Research, Shenyang, China\\
        \textsuperscript{3}Meituan Inc.
        \textsuperscript{4}Harbin Institute of Technology, Harbin, China\\
	\ttfamily{yuchunfan\_neu@outlook.com} \ttfamily{\{xiaotong,zhujingbo\}@mail.neu.edu.cn}
}

\begin{document}
\maketitle
\begin{abstract}
Large vision-language models (LVLMs) have demonstrated exceptional capabilities in understanding visual information with human languages but also exhibit an imbalance in multilingual capabilities.
In this work, we delve into the multilingual working pattern of LVLMs and identify a salient correlation between the multilingual understanding ability of LVLMs and language-specific neuron activations in shallow layers. 
Building on this insight, we introduce \ours, a training recipe that achieves efficient multilingual enhancement for LVLMs by \textbf{P}recise \textbf{LA}nguage-\textbf{S}pecific layers fine-\textbf{T}uning.
\ours first identifies layers involved in multilingual understanding by monitoring language-specific neuron activations. These layers are then precisely fine-tuned with question-translation pairs to achieve multilingual alignment. Our empirical results on MMBench and MMMB demonstrate that \ours effectively improves the multilingual capabilities of LVLMs and achieves significant efficiency with only 14\% of the parameters tuned.
Further analysis reveals that \ours can be generalized to low-resource and complex visual reasoning tasks, facilitating the language-specific visual information engagement in shallow layers\footnote{The project will be available at: \url{https://github.com/fmm170/PLAST}}.
\end{abstract}

\section{Introduction}

Large vision-language models (LVLMs) have made remarkable progress in understanding visual information with human languages, achieving impressive performance in multimodal tasks like visual question answering (VQA)~\citep{2024-Liu,2024-LiuYuan,2024-Lin}. However, they still struggle in multilingual scenarios because the training corpora are predominantly English-centric \citep{2023-Chen-Xi}. This imbalance poses significant challenges for their real-world applications across the globe.

Recent efforts to enhance the multilingual abilities of LVLMs primarily focus on two aspects. One line of research~\citep{qin2023cross, zhang2024multimodal,guo2023connecting,mu-acl2023} involves incorporating translation, either implicitly or explicitly, into the design of prompts, encouraging the model to first comprehend questions in English and then solve them step by step. However, the inadequate multilingual understanding capability of LVLMs often triggers cascading errors, leading to inferior performance. Another approach expands this idea by adopting a \textit{translate-then-train} strategy~\citep{2024-Sun-hailong, qu2024mitigating}, where English training data are first translated into multiple languages via machine translation, followed by fine-tuning for multilingual instruction alignment. Despite effectiveness, these approaches rely on accurately translating large-scale complex multimodal data and full-parameter fine-tuning, which limits their applicability in data-deficiency and resource-constrained scenarios.
\begin{figure}[t!]
    \centering
    \includegraphics[width=0.48\textwidth]{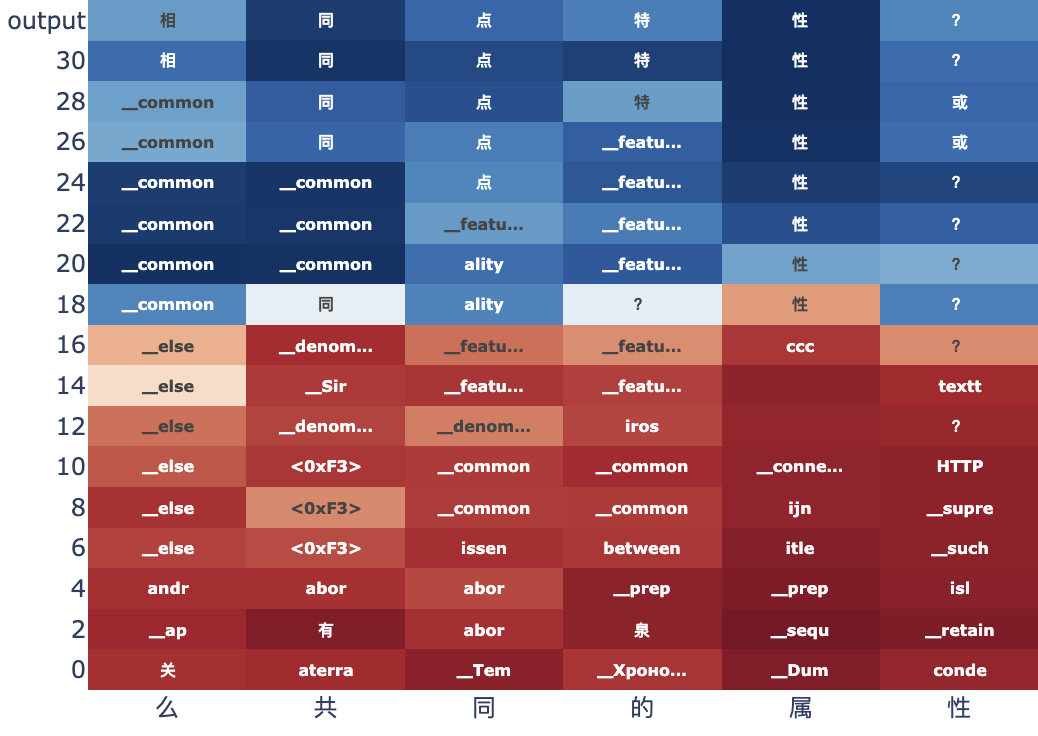}
    \caption{The logit lens \citep{nostalgebraist2020logitlens} applies the language modeling head (LLaVA-1.5-7B in our case) to intermediate layer embeddings, producing one next-token distribution per position (x-axis) and per layer (y-axis). The blue-to-red gradient indicates entropy levels, ranging from low to high. Full visualizations for all languages are shown in Figure~\ref{fig:arabic}–\ref{fig:chinese}.}
    \label{fig:three-stage-logits-lens}
\end{figure}

When considering multilingual models, one might think of capturing language-specific representation in certain parts of these models.
Inspired by recent studies~\citep{Tang2024LanguageSpecific,zhao2024how,zhao2024large} on the multilingual mechanisms in large language models (LLMs), which have verified that the multilingual working pattern of LLMs can be divided into three stages: language understanding, task-solving, and language converting. Moreover, during this process, the activation of language-specific neurons~\citep{Fan2025SLAM} can serve as a crucial indicator for distinguishing each working stage. We further investigate \textit{whether this pattern persists in LVLMs} (\S\ref{sec:pilot-study}). Our pilot study reveals that LVLMs also follow this three-stage process for handling multilingualism, with lower-level layers more engaged in learning language-specific representations. As shown in Figure~\ref{fig:three-stage-logits-lens}, the Chinese query undergoes an explicit process of language understanding ($\texttt{Zh} \rightarrow \texttt{En}$) and language converting ($\texttt{En} \rightarrow \texttt{Zh}$).
This motivates us to unlock the multilingual capabilities more efficiently by precisely enhancing language-specific representations.

\begin{figure}[t!]
    \centering
    \definecolor{rou}{RGB}{248, 214, 207}
\definecolor{proj-inside}{RGB}{253, 243, 231}
\definecolor{proj-side}{RGB}{249, 219, 183}
\definecolor{encoder-inside}{RGB}{230, 246, 253}
\definecolor{encoder-side}{RGB}{160, 224, 246}
\definecolor{embedding-inside}{RGB}{217, 217, 252}
\definecolor{embedding-side}{RGB}{159, 165, 247}
\definecolor{emb-inside}{RGB}{223, 246, 240}
\definecolor{LLM-inside}{RGB}{232, 242, 223}
\definecolor{LLM-side}{RGB}{167, 210, 165}
\begin{tikzpicture}

\tikzstyle{dash-re} = [draw, dashed, line width=0.5pt, inner sep=0.4em, rounded corners=4pt, minimum width=2.8cm, minimum height=0.8cm]

\tikzstyle{proj} = [draw=proj-side,line width=0.5pt,fill=proj-inside, inner sep=0.1cm, rounded corners=2pt, minimum width=2.8cm,minimum height=0.3cm]

\tikzstyle{encoder} = [draw=encoder-side,line width=0.5pt,fill=encoder-inside, inner sep=0.1cm, rounded corners=2pt, minimum width=2.8cm,minimum height=0.4cm]

\tikzstyle{embedding} = [draw=embedding-side,line width=0.5pt,fill=embedding-inside, inner sep=0.1cm, rounded corners=2pt, minimum width=2.8cm,minimum height=0.76cm]

\tikzstyle{embtoken} = [draw=black,thick,fill=emb-inside, inner sep=0.1cm, rounded corners=2pt, minimum width=0.4cm,minimum height=0.4cm]

\tikzstyle{LLM} = [draw=LLM-side,thick,fill=LLM-inside, inner sep=0.1cm, rounded corners=2pt, minimum width=7cm,minimum height=1.4cm]

\tikzstyle{examples-node} = [draw, dashed,fill=gray!10, dash pattern=on 1.5pt off 1.5pt,line width=0.5pt, rounded corners=1.5pt,minimum width=3.3cm,minimum height=0.36cm]

\node[anchor=center] (start) at (2.18, -0.1cm) {...}; 
\node[dash-re,anchor=center] (dash1) at ([xshift=-0.95cm,yshift=0.25cm]start.south){};
\node[proj,anchor=center] (proj) at ([xshift=-0cm,yshift=-0.4cm]dash1.south){};
\node[anchor = south,font=\scriptsize] (proj-char) at ([xshift = 0em,yshift=-0.25em]proj.south) {Projection};
\node[anchor = south,font=\footnotesize] (fire1) at ([xshift = -1.1cm,yshift=-0.2cm]proj.south) {\mylogo};
\node[encoder,anchor=center] (encoder) at ([xshift=0cm,yshift=-0.25cm]proj.south){};
\node[anchor = south,font=\scriptsize] (encoder-char) at ([xshift = 0em,yshift=-0.05em]encoder.south) {Vision Encoder};
\node[anchor = south,font=\footnotesize] (fire1) at ([xshift = -1.1cm,yshift=-0.16cm]encoder.south) {\mylogosnow};
\node[dash-re,anchor=center] (dash2) at ([xshift=2cm,yshift=0cm]dash1.east){};
\node[embedding,anchor=center] (embedding) at ([xshift=0cm,yshift=-0.63cm]dash2.south){};
\node[anchor = south,font=\scriptsize] (embedding-char) at ([xshift = 0em,yshift=0.25em]embedding.south) {Word Embedding};
\node[dash-re,anchor=center] (dash3) at ([xshift=0cm,yshift=-0.8cm]embedding.south){};

\node[embtoken,anchor=center] (mul-in1) at ([xshift=-1.05cm,yshift=-0.5cm]dash3.north){};
\node[anchor = south,font=\scriptsize] (mul-in1-char) at ([xshift = 0em,yshift=-1.1em]mul-in1.north) {Ar};
\node[embtoken,anchor=center] (mul-in2) at ([xshift=0.31cm,yshift=0cm]mul-in1.east){};
\node[anchor = south,font=\scriptsize] (mul-in2-char) at ([xshift = 0em,yshift=-1.1em]mul-in2.north) {Tr};
\node[embtoken,anchor=center] (mul-in3) at ([xshift=0.31cm,yshift=0cm]mul-in2.east){};
\node[anchor = south,font=\scriptsize] (mul-in3-char) at ([xshift = 0em,yshift=-1.1em]mul-in3.north) {Ru};
\node[embtoken,anchor=center] (mul-in4) at ([xshift=0.31cm,yshift=0cm]mul-in3.east){};
\node[anchor = south,font=\scriptsize] (mul-in4-char) at ([xshift = 0em,yshift=-1.1em]mul-in4.north) {Pt};
\node[embtoken,anchor=center] (mul-in5) at ([xshift=0.31cm,yshift=0cm]mul-in4.east){};
\node[anchor = south,font=\scriptsize] (mul-in5-char) at ([xshift = 0em,yshift=-1.1em]mul-in5.north) {Zh};
\node[anchor = south,font=\footnotesize] (trans) at ([xshift = -1.2cm,yshift=0.45cm]dash3.south) {\mylogotranslate};
\node[anchor = south,font=\footnotesize] (text-input) at ([xshift = 0em,yshift=-3.6em]dash3.north) {\textbf{Translation Instructions}};
\node[anchor = south,font=\footnotesize] (text-input) at (4.1,2.1) {\textbf{English Response}};
\node[anchor = south,font=\footnotesize] (fire1) at ([xshift = -1.15cm,yshift=0cm]embedding.south) {\mylogosnow};

\node[anchor = south,font=\footnotesize] (fig) at ([xshift = 0cm,yshift=-1.5cm]encoder.south) {\mylogofig};

\draw[->,line width=0.5pt,>=stealth,draw=gray!50] (0.225,0.2) -- (5.625,2.3);
\draw[->,line width=0.5pt,>=stealth,draw=gray!50] (0.775,0.2) -- (5.625,2.3);
\draw[->,line width=0.5pt,>=stealth,draw=gray!50] (1.325,0.2) -- (5.625,2.3);
\draw[->,line width=0.5pt,>=stealth,draw=gray!50] (2.225,0.2) -- (5.625,2.3);

\draw[->,line width=0.5pt,>=stealth,draw=gray!50] (0.225,0.2) -- (6.15,2.3);
\draw[->,line width=0.5pt,>=stealth,draw=gray!50] (0.775,0.2) -- (6.15,2.3);
\draw[->,line width=0.5pt,>=stealth,draw=gray!50] (1.325,0.2) -- (6.15,2.3);
\draw[->,line width=0.5pt,>=stealth,draw=gray!50] (2.225,0.2) -- (6.15,2.3);

\draw[->,line width=0.5pt,>=stealth,draw=gray!50] (3.625,0.2) -- (5.625,2.3);
\draw[->,line width=0.5pt,>=stealth,draw=gray!50] (4.175,0.2) -- (5.625,2.3);
\draw[->,line width=0.5pt,>=stealth,draw=gray!50] (4.725,0.2) -- (5.625,2.3);
\draw[->,line width=0.5pt,>=stealth,draw=gray!50] (5.625,0.2) -- (5.625,2.3);

\draw[->,line width=0.5pt,>=stealth,draw=gray!50] (3.625,0.2) -- (6.15,2.3);
\draw[->,line width=0.5pt,>=stealth,draw=gray!50] (4.175,0.2) -- (6.15,2.3);
\draw[->,line width=0.5pt,>=stealth,draw=gray!50] (4.725,0.2) -- (6.15,2.3);
\draw[->,line width=0.5pt,>=stealth,draw=gray!50] (5.625,0.2) -- (6.15,2.3);

\draw[->,line width=0.5pt,>=stealth] (dash3.north) -- (embedding.south);
\draw[->,line width=0.5pt,>=stealth] (embedding.north) -- (dash2.south);
\draw[->,line width=0.5pt,>=stealth] (proj.north) -- (dash1.south);
\draw[->,line width=0.5pt,>=stealth] ([xshift = 0cm,yshift=-0.3cm]encoder.south) -- (encoder.south);

\node[LLM,anchor=west,fill opacity=0.9] (LLM) at ([xshift=-0cm,yshift=1.3cm]dash1.west){};
\draw[-,thick,draw=LLM-side] (-0.18,1) -- (6.85,1);
\node[anchor = south,font=\footnotesize] (fire1) at ([xshift = -3.2cm,yshift=-0.15cm]LLM.south) {\mylogo};
\node[anchor = south,font=\footnotesize] (fire1) at ([xshift = -3.2cm,yshift=0.5cm]LLM.south) {\mylogosnow};

\node[anchor = south,font=\footnotesize] (gear) at ([xshift = 2.5cm,yshift=0.5cm]LLM.south) {\mylogogear};
\node[anchor = south,font=\footnotesize] (fire1) at ([xshift = 0cm,yshift=-0.6cm]gear.south) {\mylogoidea};

\texthidden{3.4}{0}
\texthidden{3.95}{0}
\texthidden{4.5}{0}
\node[anchor=center] (start2) at (5.16, -0.1cm) {...}; 
\node[anchor=center] (startn) at (1.76, -0.1cm) {...}; 
\texthidden{5.4}{0}
\textoutput{5.4}{2.5}
\textoutput{5.95}{2.5}

\node[anchor=center] (start3) at (6.6, 2.35) {...}; 

\visionhidden{0}{0}         
\visionhidden{0.55}{0} 
\visionhidden{1.1}{0} 
\visionhidden{2}{0} 

\node[examples-node, anchor=south] (examples-node) at ([xshift=-1.94cm,yshift=0.7cm]LLM.north) {};
\node[anchor=south](Reasoning-data) at ([xshift=-1.35cm,yshift=-0.17cm]examples-node.south){\mylogo};

\node[anchor = west,font=\scriptsize] (Reasoning-data-input) at ([xshift = -0.17cm,yshift=0cm]Reasoning-data.east) {Fine-tune};

\node[anchor=west](Reasoning-data) at ([xshift=1.05cm,yshift=0cm]Reasoning-data.east){\mylogosnow};
\node[anchor = west,font=\scriptsize] (Reasoning-data-input) at ([xshift = -0.1cm,yshift=0cm]Reasoning-data.east) {Freeze};
\node[anchor = south,font=\footnotesize] (text-input) at ([xshift = 0em,yshift=-7.4em]dash1.south) {\textbf{Visual Input}};

\node[anchor = south,font=\footnotesize] (text-input) at ([xshift = 0em,yshift=-2.3em]LLM.north) {Task Solving};

\node[anchor = south,font=\footnotesize] (text-input) at ([xshift = 0em,yshift=-3.9em]LLM.north) {Multilingual Understanding};

\end{tikzpicture}
    \caption{An overview of our method, \ours.  \protect\inlinecolorboxone\ denotes the question-translation data. For instance, the instruction for training is: ``Translate this from Chinese to English: \begin{CJK}{UTF8}{gbsn} \footnotesize{这座建筑是什么样子的?}\end{CJK}''.}
    \label{fig:Framework}
\end{figure}
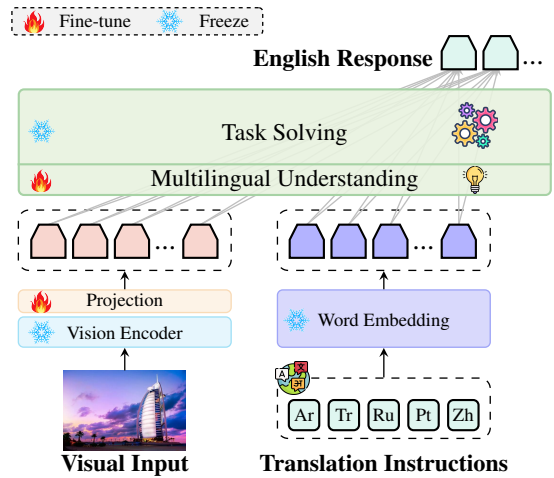

To this end, we propose \ours, a training recipe designed to achieve efficient multilingual enhancement for LVLMs, as outlined in Figure~\ref{fig:Framework}. Specifically, starting with a visual instruction-tuned model, we first identify decoder layers mostly involved in language understanding by monitoring the activation of language-specific neurons using multilingual image-text pairs (\S\ref{sec:method-step2}). More concretely, we calculate the mean squared deviation (MSD) of the numbers of neurons activated by different languages and select the layers with higher MSD scores. Subsequently, \ours precisely fine-tunes only these identified layers using \textit{a small amount of} image-question translation pairs (\S\ref{sec:method-step3}). This improves its multilingual understanding ability by precisely enhancing the language-specific representation while not affecting task-solving abilities at higher layers.

To evaluate the effectiveness of \ours, we conduct extensive experiments on two multilingual VQA benchmarks, MMBench~\citep{2024-LiuYuan} and MMMB~\citep{2024-Sun-hailong} across three LVLMs.
Experimental results demonstrate the effectiveness of \ours in improving multilingual performance, with average gains of 8.0\% and 4.0\% on MMBench and MMMB.
Furthermore, compared to full-parameter fine-tuning approaches, \ours achieves superior efficiency with only 15.6\% and 12.5\% of the parameters within 7B and 13B models tuned.
Further analysis reveals that \ours can be generalized to low-resource and complex visual reasoning tasks, facilitating the language-specific visual information engagement in shallow layers.

\section{Preliminaries}

\label{sec:pilot-study}
Recent studies~\citep{Tang2024LanguageSpecific,Fan2025SLAM} have underscored the pivotal role of language-specific neurons within LLMs for multilingual processing. Drawing from these findings, this section starts with a preliminary analysis aimed at exploring whether neurons within LVLMs exhibit language-specific behavior and understanding their significance for multilingual visual understanding.

\subsection{Notation}
LVLMs, represented by the LLaVA series~\citep{liu2023llava}, typically comprise a vision encoder with a pre-trained LLM. 
Given a visual input $\bm{X}_{v}$ and a textual input $\bm{X}_{t}$, the visual content is encoded into vision tokens $\bm{H_{v}}$ by a vision encoder (\eg, CLIP ViT-L/14~\citep{Radford2021Learning}) and a projection layer. Concurrently, the textual input $\bm{X}_{t}$ is converted into textual embeddings $\bm{H}_{t}$ via the LLM's embedding layer.
These vision tokens and textual embeddings, concatenated into $\bm{H}$, are subsequently processed by the LLM (\eg, Vicuna~\citep{vicuna2023}) to generate the output.

More precisely, the feed-forward network (FFN) sub-layer within the $i$-th layer of the LLM backbone is formulated as:
\begin{equation}
\label{eq:neuron}
\text{FFN}(\bm{H}^{i}) = \left[ f(\bm{W}_\text{gate}^{i}\bm{H}^{i}) \otimes (\bm{W}_\text{up}^{i}\bm{H}^{i}) \right] \bm{W}_\text{down}^{i},
\end{equation}
where $\bm{W}_\text{gate}^{i}$, $\bm{W}_\text{up}^{i}$ $\in$ $\mathbb{R}^{d_\text{model} \times d_\text{inter}}$ are weight matrices for the gate and up-projection respectively, and $\bm{W}_\text{down}^{i}$ $\in$ $\mathbb{R}^{d_\text{inter} \times d_\text{model}}$ is the down-projecting matrix. $f(\cdot)$ denotes a non-linear activation function. 

\begin{figure}[t!]
    \centering
    \definecolor{upurple}{RGB}{155,89,182}
\definecolor{ured}{RGB}{231,76,60}
\definecolor{udark}{RGB}{77,153,77}
\definecolor{udpdark}{HTML}{85827a}
\definecolor{usemidark}{HTML}{8c564b}
\definecolor{ublue}{RGB}{52,152,219}
\definecolor{udpblue}{HTML}{0419fb}
\definecolor{usemiblue}{HTML}{17becf}
\definecolor{uorange}{HTML}{ffcc00}
\definecolor{udporange}{HTML}{bcbd22}
\definecolor{udpcyonAR}{RGB}{070,120,142}
\definecolor{ucyonTr}{RGB}{80,160,180}
\definecolor{uyellowRu}{RGB}{255,210,100}
\definecolor{uorangePt}{RGB}{255,140,000}
\definecolor{ugreenZh}{RGB}{121,149,025}
\tikzset{global scale/.style={
    scale=#1,
    every node/.append style={scale=#1}
  }
}

\pgfplotsset{
    width=0.685\textwidth,
   height=0.28\textheight,
   grid=major,
   major grid style={dotted},
   enlarge y limits={upper,value=0.05},
   legend style={
      fill,
      at={(0.50,16.5em)},
      legend columns=3,
      legend cell align=left,
      anchor=south
      },
   }
\begin{tikzpicture}
\useasboundingbox (13em,-1em) rectangle (3.5em,10em);
\begin{axis}[
global scale = 0.63,
   at={(1em,1em)},
    legend entries={Ar,Tr,Ru,Pt,Zh,AVG},
    ymajorgrids=true,
    xmajorgrids=true,
    legend style={draw=none,
    line width=1pt,
    at={(13.7em,11em)},
    legend cell align=left,
    column sep=3pt,
    yshift=-5pt,
    anchor=south
    },
    enlarge x limits={abs=1},
   xtick={1,2,3,4,5,6,7,8,9,10},
   xticklabels={1,2,3,4,5,6,7,8,9,10},
   ymin=0.2, ymax=0.55,
   ytick={0.20,0.25,0.3,0.35,0.4,0.45,0.5,0.55},
   yticklabels={0.20,0.25,0.30,0.35,0.40,0.45,0.50,0.55},
   yticklabel style={/pgf/number format/fixed,/pgf/number format/fixed zerofill,/pgf/number format/precision=2,rotate=0}
   ]

   \addplot[sharp plot,orange!65,smooth,thick,line width=0.5pt,mark=pentagon*,mark size=2pt,thick,mark options={fill=orange!55,draw=orange!65,line width=0.5pt}] plot coordinates {
(1, 0.487) (2, 0.386) (3, 0.361) (4, 0.369) (5, 0.308) (6, 0.278) (7, 0.269) (8, 0.263) (9, 0.270) (10, 0.279)
      };
   \addplot[sharp plot,udark!65,smooth,thick,line width=0.5pt,mark=square*,mark size=1.6pt,thick,mark options={fill=udark!65,draw=udark!75,line width=0.5pt}] plot coordinates {
(1, 0.440) (2, 0.377) (3, 0.326) (4, 0.327) (5, 0.321) (6, 0.271) (7, 0.247) (8, 0.241) (9, 0.262) (10, 0.271)
      };

   \addplot[sharp plot,udpdark!65,smooth,thick,line width=0.5pt,mark=triangle*,,mark size=2pt,thick,mark options={fill=udpdark!65,draw=udpdark!75,line width=0.5pt}] plot coordinates {
(1, 0.489) (2, 0.433) (3, 0.338) (4, 0.338) (5, 0.307) (6, 0.255) (7, 0.245) (8, 0.246) (9, 0.270) (10, 0.280)
          };
   \addplot[sharp plot,udpblue!40,smooth,thick,line width=0.5pt,mark=pentagon*,mark size=2pt,thick,mark options={fill=udpblue!30,draw=udpblue!40,line width=0.5pt}] plot coordinates{
(1, 0.451) (2, 0.381) (3, 0.312) (4, 0.317) (5, 0.288) (6, 0.244) (7, 0.233) (8, 0.237) (9, 0.266) (10, 0.274)
          };

   \addplot[sharp plot,usemiblue!65,smooth,thick,line width=0.5pt,mark=triangle*,mark size=2pt,thick,mark options={fill=usemiblue!55,draw=usemiblue!65,line width=0.5pt}] plot coordinates{
(1, 0.502) (2, 0.419) (3, 0.342) (4, 0.342) (5, 0.305) (6, 0.256) (7, 0.240) (8, 0.240) (9, 0.265) (10, 0.276)
          };

    \addplot[sharp plot,ured!65,smooth,thick,line width=0.5pt,mark=*,mark size=1.6pt,thick,mark options={fill=ured!65,draw=ured!75,line width=0.5pt}] plot coordinates {
(1, 0.473) (2, 0.399) (3, 0.335) (4, 0.338) (5, 0.305) (6, 0.260) (7, 0.246) (8, 0.245) (9, 0.266) (10, 0.276)
          };

   \end{axis}
\node at (8.5em,-0.4em){\footnotesize Layer Index};

   \node  [rotate=90, align=center] at(-1.3em,5em){\scriptsize Neuron activation numbers };
\end{tikzpicture}
 \caption{The number of activated neurons $R_{l}^{i}$ across all non-English languages. “AVG” indicates the average activation level computed over these languages.}
 
    \label{fig:neurons}
\end{figure}
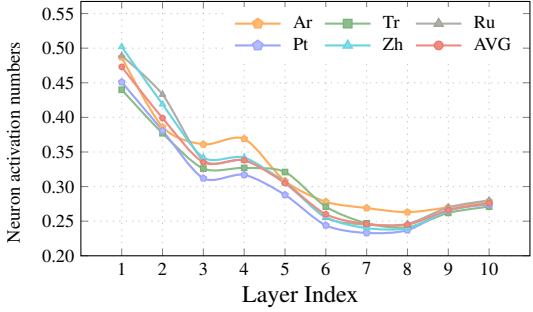
In the context of an FFN sub-layer, a neuron is defined as a single column within $\bm{W}_\text{up}^{i}$, implying that each FFN sub-layer contains $d_\text{inter}$ neurons. A specific neuron in the $i$-th FFN sub-layer is considered \textit{activated} if its corresponding value within the  $f(\bm{W}_\text{gate}^{i}\bm{H}^{i})$ exceeds zero~\citep{Tang2024LanguageSpecific}.

\subsection{Experimental Setups for Pilot Study}
To investigate the influence of language-specific neurons on the multilingual understanding capabilities of LVLMs, we conduct our pilot study using the MMBench dataset~\citep{2024-Sun-hailong}.
MMBench is a comprehensive multilingual VQA dataset, featuring image-question pairs annotated across six distinct languages. For our analysis, we sample $n$ instances per language.
We denote the language of the question in each pair as $l$. 
For each language $l$, we quantify the activation of neurons by the image-question pair in the $i$-th layer as follows:
\begin{equation}
  \label{eq:activation}  
  \bm{A}_{l}^{i} = \mathbb{I}[f(\bm{W}_\text{gate}^{i}\bm{H}^{i})>0 ],
\end{equation} 
where $\mathbb{I}$ is the element-wise indicator function. 
We then normalize the number of activated neurons by the total number of neurons in layer $i$ to compute the proportion of activated neurons as follows:
\begin{equation}
  \label{eq:activated-normalization}  
  R_{l}^{i} = \frac{\sum_{k=1}^{d_\text{inter}}(\bm{A}_{l}^i)_k}{d_\text{inter}}.
\end{equation}

To analyze language-specific activation patterns, we categorize activated neurons sets $\mathcal{N}^i$ in layer $i$ by English and non-English. Let \( \mathcal{L}_{\mathrm{non\text{-}eng}} \) denote the set of all non-English languages. The overlap ratio of activated neuron sets between each non-English \(l \in \mathcal{L}_{\mathrm{non\text{-}eng}} \) and English is computed as:
\begin{equation}
  \label{eq:neuron-overlap}  
  O^i_{l} = \frac{|\mathcal{N}^i_l \cap \mathcal{N}^i_{\mathrm{eng}}|}{|\mathcal{N}^i_{\mathrm{eng}}|}.
\end{equation}

\subsection{Observations}
\begin{figure}[t!]
    \centering
    \definecolor{upurple}{RGB}{155,89,182}
\definecolor{ured}{RGB}{231,76,60}
\definecolor{udark}{RGB}{77,153,77}
\definecolor{udpdark}{HTML}{85827a}
\definecolor{usemidark}{HTML}{8c564b}
\definecolor{ublue}{RGB}{52,152,219}
\definecolor{udpblue}{HTML}{0419fb}
\definecolor{usemiblue}{HTML}{17becf}
\definecolor{uorange}{HTML}{ffcc00}
\definecolor{udporange}{HTML}{bcbd22}

\definecolor{udpcyonAR}{RGB}{070,120,142}
\definecolor{ucyonTr}{RGB}{80,160,180}
\definecolor{uyellowRu}{RGB}{255,210,100}
\definecolor{uorangePt}{RGB}{255,140,000}
\definecolor{ugreenZh}{RGB}{121,149,025}
\tikzset{global scale/.style={
    scale=#1,
    every node/.append style={scale=#1}
  }
}

\pgfplotsset{
    width=0.685\textwidth,
   height=0.28\textheight,
   grid=major,
   major grid style={dotted},
   symbolic x coords={1,2,3,4,5,6,7,8,9,10,11},
   enlarge y limits={upper,value=0.1},
   legend style={
      fill,
      at={(0.50,16.5em)},
      legend columns=3,
      legend cell align=left,
      anchor=south
      },
   }
\begin{tikzpicture}
\useasboundingbox (13em,-1em) rectangle (3.5em,10em);
\begin{axis}[
global scale = 0.63,
   at={(1em,1em)},
    legend entries={Ar,Tr,Ru,Pt,Zh,AVG},
    ymajorgrids=true,
    xmajorgrids=true,
    legend style={draw=none,
    line width=1pt,
    at={(6.5em,11em)},
    legend cell align=left,
    column sep=3pt,
    yshift=-5pt,
    anchor=south
    },
   xtick={1,2,3,4,5,6,7,8,9,10},
   xticklabels={1,2,3,4,5,6,7,8,9,10},
   ymin=0.65, ymax=0.95,
   ytick={0.65,0.675,0.7,0.725,0.75,0.775,0.8,0.825,0.85,0.875,0.9,0.925,0.95},
   yticklabels={0.65,,0.70,,0.75,,0.80,,0.85,,0.90,,0.95},
   yticklabel style={/pgf/number format/fixed,/pgf/number format/fixed zerofill,/pgf/number format/precision=2,rotate=0}
   ]
   \addplot[sharp plot,orange!65,smooth,thick,line width=0.5pt,mark=pentagon*,mark size=2pt,thick,mark options={fill=white,draw=orange!65,line width=0.5pt}] plot coordinates {
(1, 0.695) (2, 0.736) (3, 0.723) (4, 0.736) (5, 0.708) (6, 0.707) (7, 0.742) (8, 0.766) (9, 0.799) (10, 0.784)
      };
   \addplot[sharp plot,udark!65,smooth,thick,line width=0.5pt,mark=square*,mark size=1.6pt,thick,mark options={fill=white,draw=udark!65,line width=0.5pt}] plot coordinates {
(1, 0.761) (2, 0.833) (3, 0.757) (4, 0.790) (5, 0.761) (6, 0.773) (7, 0.807) (8, 0.833) (9, 0.853) (10, 0.846)
      };
      \addplot[sharp plot,udpdark!65,smooth,thick,line width=0.5pt,mark=triangle*,,mark size=2pt,thick,mark options={fill=white,draw=udpdark!65,line width=0.5pt}] plot coordinates {
(1, 0.840) (2, 0.861) (3, 0.837) (4, 0.847) (5, 0.830) (6, 0.835) (7, 0.857) (8, 0.875) (9, 0.892) (10, 0.884)

      };
   \addplot[sharp plot,udpblue!40,smooth,thick,line width=0.5pt,mark=pentagon*,mark size=2pt,thick,mark options={fill=white,draw=udpblue!40,line width=0.5pt}] plot coordinates{
(1, 0.856) (2, 0.867) (3, 0.823) (4, 0.861) (5, 0.867) (6, 0.866) (7, 0.883) (8, 0.902) (9, 0.924) (10, 0.909)

      };
   \addplot[sharp plot,usemiblue!65,smooth,thick,line width=0.5pt,mark=triangle*,mark size=2pt,thick,mark options={fill=white,draw=usemiblue!65,line width=0.5pt}] plot coordinates{
(1, 0.853) (2, 0.874) (3, 0.857) (4, 0.863) (5, 0.851) (6, 0.850) (7, 0.870) (8, 0.885) (9, 0.914) (10, 0.898)
      };

       \addplot[sharp plot,red!80,smooth,thick,line width=0.5pt,mark=oplus,mark size=1.6pt,thick,mark options={fill=white,draw=red!80,line width=0.5pt}] plot coordinates {
(1, 0.801) (2, 0.834) (3, 0.799) (4, 0.819) (5, 0.803) (6, 0.806) (7, 0.832) (8, 0.852) (9, 0.876) (10, 0.864)
          };

   \end{axis}
  \draw[red!80, dashed] (13.62em, 7em) -- (13.62em, 1em);
  \draw[red!80, dashed] (13.62em, 7em) -- (1em, 7em);
\node at (8.5em,-0.4em){\footnotesize Layer Index};
\draw[->, >=stealth, semithick,black!80] (4.7em,1.4em) -- (2.3em,1.4em);
\draw[->, >=stealth, semithick,black!80] (10.65em,1.4em) -- (13.62em,1.4em);
\node at (7.7em,1.4em){\tiny Language-specific Layers};
   \node  [rotate=90] at(-1.3em,5.1em){\scriptsize Overlap Ratio };
\end{tikzpicture}
    \caption{The overlap ratio $O_{l}^{i}$ between non-English and English activated neurons. “AVG” indicates the average overlap ratio among all non-English languages.}
    \label{overlap}
\end{figure}
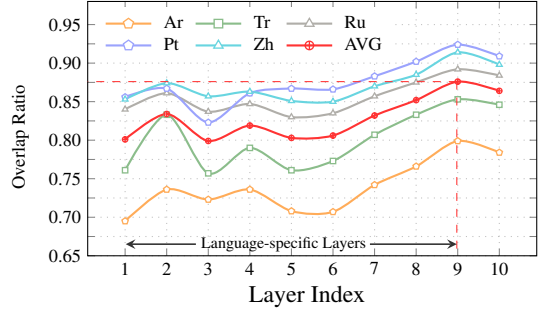

Our analysis reveals significant language-specific neurons activity throughout the decoder layers of LVLMs.
Illustrative data for the LLaVA-1.5-7B are presented in Figure~\ref{fig:neurons}, which shows the count of activated neurons per layer, and Figure~\ref{overlap}, which depicts the overlap ratio of activated neurons between non-English and English. 
We observe a progressive decline in the total number of activated neurons across non-English languages with increasing layer depth. Conversely, the overlap ratio of activated neurons between non-English and English increases, ultimately reaching a distinct peak.

This pattern suggests that language representations are initially distinct and independent, while progressively converging toward an English-centric pattern at greater depths. This confirms that \textbf{certain layers within the decoder are primarily responsible for processing language-specific representation}.
Building upon this, we classify the layers preceding the point where the average overlap ratio among all non-English languages attains its maximum as \textit{language-specific layers}. Full visualization of all layers is available in Appendix~\ref{sec:full_visualization}.

\section{Methodology}

Building upon our findings, we propose a training method designed to achieve \textit{efficient} multilingual enhancement of LVLMs, which includes two processes: (1) identifying layers responsible for multilingual understanding; (2) precisely fine-tuning the selected layers to enhance multilingual capabilities.

\subsection{Select Multilingual Understanding Layers}
\label{sec:method-step2}

Prior research \citep{zhang2024redundancy} suggests that LVLMs predominantly extract crucial information from visual representations in shallow to intermediate layers to facilitate the core task-solving process.
Indiscriminately fine-tuning all language-specific layers might compromise inherent general capabilities embedded within them.
Consequently, we introduce a more granular layer selection algorithm designed to strike a balance between enhancing multilingual understanding and maintaining the model's general abilities.
To this end, we utilize the mean squared deviation (MSD) to precisely measure the stability of neuron activation across different languages.
For each layer $i$ within language-specific layers $\mathcal{K}$, we compute its $\mathrm{MSD^{\textit{i}}}$ as follows: 
\begin{align}
  \label{eq:mean}  
  \mu^{i} &= \frac{1}{|L|} \sum_{l \in L} R_{l}^{i},  \\
  \label{eq:msd}
  \mathrm{MSD^{\textit{i}}} &= \frac{1}{|L|} \sum_{l \in L} (R_{l}^{i} - \mu^{i})^2, 
\end{align}
where $L$ denotes the collection of all languages, and $R_{l}^{i}$ represents the normalized count of activated neurons, as calculated by Equation (\ref{eq:activated-normalization}). 

A higher $\mathrm{MSD^{\textit{i}}}$ for a given layer indicates a greater divergence in its activation pattern across different languages, suggesting a more substantial engagement in multilingual understanding rather than general abilities. 
To quantify the average engagement in multilingual understanding across language-specific layers, we calculate as follows:
\begin{equation}
  \label{eq:theta}
\theta = \frac{1}{|\mathcal{K}|} \sum_{i \in \mathcal{K}}\mathrm{MSD^{\textit{i}}}.
\end{equation} 
Layers whose $\mathrm{MSD^{\textit{i}}}$ exceeds the threshold $\theta$ are ultimately selected for fine-tuning, as these layers contribute more significantly to multilingual understanding than to general capabilities.

\subsection{Supervised Fine-Tuning for Enhancing Multilingual Understanding}
\label{sec:method-step3}
Recent studies~\citep{basu2024understanding,zhang2024Crossmodal,Zhang2025LLaVAMiniEI} have shown that text tokens effectively integrate visual information from vision tokens via attention sub-layers, particularly within the shallow layers of LVLMs.
In light of this, we undertake fine-tuning across all modules within the selected layers. This not only preserves the model's fundamental ability to process visual information within these crucial layers, but also significantly enhances its capacity to comprehend multilingual questions. 
Given the visual inputs $X_{v}$, non-English questions $X_{t,l}$ and their English counterparts $X_{t,eng}$, the loss function is formulated as:
\begin{equation}
  \label{eq:theta-loss}
\mathcal{L} = -\sum_{ l \in \mathcal{L}_{\mathrm{non\text{-}eng}}} \log P(X_{t,\mathrm{eng}} \mid X_{t,l}, X_{v}; \theta),
\end{equation}
where $\theta$ represents the parameters of the selected layers actively involved in the fine-tuning process.
\section{Experiment Settings}
\subsection{Datasets}
To assess the efficacy of \ours, our main experiments are conducted on two multilingual VQA datasets covering six languages, including Arabic (Ar), Turkish (Tr), Russian (Ru), Portuguese (Pt), Chinese (Zh), and English (En).

\paragraph{MMBench} \citep{2024-Sun-hailong} consists of data in six languages, translated from the original MMBench \citep{2024-LiuYuan} using GPT-4 \citep{2023-ChatGPT}. These translations are subsequently verified manually to ensure their accuracy.

\paragraph{MMMB} \citep{2024-Sun-hailong} is compiled by sampling items from the ScienceQA \citep{Lu2022learn}, MME \citep{fu2023MME}, and SEED-Bench \citep{Li2024SEEDBenchBM} datasets, which are then translated into five other languages using GPT-4.
\subsection{Evaluation Metrics}
We follow the evaluation settings from \citet{2024-Sun-hailong}, primarily focusing on \textit{accuracy}. During our assessments, we employ the VLMEvalKit from OpenCompass \citep{2023opencompass}, and ensure consistent configuration settings across all compared methods to facilitate a fair comparison.
\subsection{Baselines}
We compare our method with two types of baselines: \textit{prompting-based} and \textit{training-based}.
To validate the generalizability of \ours, we select three representative LVLMs with different sizes for evaluation: LLaVA-1.5-7B/13B~\citep{2024-Liu} and LLaVA-1.6-7B/13B~\citep{liu2024llavanext}, and Qwen-VL-Chat~\citep{bai2024qwenvl}. Comprehensive baseline details are provided in Appendix~\ref{sec:all_baselines}.

\begin{table*}[t!]
    \centering
    \resizebox{\linewidth}{!}{
%
%
\begin{tabular}{lccc|ccccccc|ccccccc}
\toprule[1.0pt]
\multirow{2}[2]{*}{\textbf{Method}} & \multirow{2}[2]{*}{ \shortstack{\\ \textbf{Training} \\ \textbf{Cost}}} & \multirow{2}[2]{*}{ \shortstack{\\ \textbf{Training} \\ \textbf{Layers}}} & \multicolumn{1}{c}{\multirow{2}[2]{*}{ \shortstack{\\ \textbf{Trained} \\ \textbf{Param.}}}} & \multicolumn{7}{c}{\textbf{MMBench}} & \multicolumn{7}{c}{\textbf{MMMB}}\\
    \cmidrule(lr){5-11}
    \cmidrule(lr){12-18}
&  & & \multicolumn{1}{c}{} & \multicolumn{1}{c}{\multirow{1}{*}{Ar}} & \multicolumn{1}{c}{\multirow{1}{*}{Tr}} & \multicolumn{1}{c}{\multirow{1}{*}{Ru}}& \multicolumn{1}{c}{\multirow{1}{*}{Pt}}& \multicolumn{1}{c}{\multirow{1}{*}{Zh}} & \multicolumn{1}{c}{\multirow{1}{*}{En}}& \multicolumn{1}{c}{\multirow{1}{*}{Avg.}} 
& \multicolumn{1}{c}{\multirow{1}{*}{Ar}} & \multicolumn{1}{c}{\multirow{1}{*}{Tr}} & \multicolumn{1}{c}{\multirow{1}{*}{Ru}}& \multicolumn{1}{c}{\multirow{1}{*}{Pt}}& \multicolumn{1}{c}{\multirow{1}{*}{Zh}} & \multicolumn{1}{c}{\multirow{1}{*}{En}}& \multicolumn{1}{c}{\multirow{1}{*}{Avg.}} \\ 
\midrule

   \textbf{\textit{LLaVA-1.5-7B}}        &   &  &  &34.6  &42.4  &54.8  &\ul{61.1}  &58.1  &  64.7 & 52.6 &41.7  &43.1  &55.1  &\textbf{59.2}  &\textbf{57.7}  &  66.2 & 53.8\\
\ \ \ \ \ + ITP &   -&-  &-  &22.2 & 39.0 & 49.6 & 55.6 & 48.1 & 64.7 & 46.5 & 30.6 & 33.6 & 42.5 & 46.8 & 45.7 & 66.2 & 44.2  \\ 
\ \ \ \ \ + ETP &  -&-  &-  &35.2 & 40.6 & \ul{58.1} & 58.0 & 52.7 & \textbf{64.7} & 51.6 & 42.4 & 45.3 & \ul{57.5} & \ul{58.7} & \ul{57.6} & \textbf{66.2} & \ul{54.6} \\ 
\ \ \ \ \ + M-SFT  & \phantom{0}7.3$\times$  & 1-32 &100.0\%  &\ul{39.3} & \ul{50.2} & 54.9 & 57.6 & \ul{58.4} & 63.7 & \ul{54.0} & \ul{44.3} & \ul{47.3} & 56.4 & 58.4 & 55.7 & 63.4 & 54.2 \\ 
\ \ \ \ \ + \textsc{QAlign}  & \phantom{0}2.8$\times$  & 1-32   &100.0\% &24.3 & 29.0 & 39.7 & 39.7 & 38.7 & 44.6 & 36.0 & 36.3 & 40.7 & 43.8 & 41.3 & 41.2 & 49.6 & 42.2 \\ 
\headercolorSLAM
\ \ \ \ \ + \ours  & \phantom{0}1.0$\times$  & 1-5\phantom{0}  & \phantom{0}15.6\%  &\textbf{44.4} & \textbf{51.9} & \textbf{58.4} & \textbf{62.3} & \textbf{59.4} & \ul{64.2} & \textbf{56.8} & \textbf{46.7} & \textbf{50.1} & \textbf{59.4} & 57.1 & 56.8 & \ul{65.1} & \textbf{55.8} \\
\midrule
   \textbf{\textit{LLaVA-1.5-13B}} &  &   &  &46.6 & 53.2 & 61.6 & 63.0 & \textbf{63.2} & 69.0 & 59.4 & 45.9 & 50.7 & \ul{62.6} & \ul{61.7} & \ul{61.6} & 69.8 & 58.7  \\
\ \ \ \ \ + ITP  & - &- &-  &43.9 & 53.5 & \textbf{62.4} & \textbf{64.7} & 61.0 & 69.0 & 59.1 & 44.5 & 49.3 & 62.2 & 60.9 & 55.4 & 69.8 & 57.0  \\ 
\ \ \ \ \ + ETP &  -&-  &-  &42.8 & 49.4 & 60.2 & 59.5 & 61.0 & \textbf{69.0} & 57.0 & 45.0 & 50.9 & \textbf{62.9} & 58.5 & \textbf{63.8} & \textbf{69.8} & 58.5  \\ 
\ \ \ \ \ + M-SFT   &  12.5$\times$  & 1-40 &100.0\% &\ul{48.4} & \textbf{59.5} & 60.6 & 61.7 & 60.9 & 67.1 & \ul{59.7} & \ul{48.7} & \ul{53.0} & 60.1 & \textbf{63.1} & 59.9 & 68.1 & \ul{58.8} \\ 
\ \ \ \ \ + \textsc{QAlign}   &   \phantom{0}4.9$\times$  &1-40  &100.0\%  &45.9 & 57.4 & 59.1 & 61.8 & 59.2 & 66.9 & 58.4 & 47.5 & 47.2 & 56.8 & 55.5 & 54.6 & 62.7 & 54.0 \\ 
\headercolorSLAM
\ \ \ \ \ + \ours   &  \phantom{0}1.0$\times$  &1-5\phantom{0}  &\phantom{0}12.5\%  &\textbf{51.5} & \ul{58.7} & \ul{62.2} & \ul{64.3} & \ul{62.3} & \ul{67.6} & \textbf{61.1} & \textbf{49.7} & \textbf{53.1} & 61.8 & 61.5 & 60.6 & \ul{69.2} & \textbf{59.3} \\ 
\midrule
   \textbf{\textit{Qwen-VL-Chat}} &  &   &  &36.7 & 40.1 & 47.9 & 49.1 & 56.0 & 57.3 & 47.9 & 43.0 & 44.1 & 51.7 & 46.4 & \ul{57.8} & 56.0 & 49.8  \\
\ \ \ \ \ + ITP  & - &- &- &23.5 & 37.2 & 42.6 & 44.8 & 49.5 & 57.3 & 42.5 & 33.2 & 34.6 & 39.3 & 34.4 & 46.0 & 56.0 & 40.6  \\ 
\ \ \ \ \ + ETP &  -&-  &-  &38.3 & 41.7 & \ul{49.5} & 48.2 & 53.6 & 57.3 & 48.1 & 45.9 & 47.9 & 52.8 & 45.5 & 53.6 & 56.0 & 50.3  \\
\ \ \ \ \ + M-SFT   &  \phantom{0}7.3$\times$  &1-32  &100.0\%  &\ul{41.6} & \textbf{51.2} & 48.7 & \ul{52.4} & \ul{58.1} & \ul{58.2} & \ul{51.7} &\ul{47.4} & \ul{48.2} & \ul{54.6} & \ul{48.5} & \textbf{58.0} & \ul{58.5} & \ul{52.5} \\ 
\ \ \ \ \ + \textsc{QAlign}    & \phantom{0}2.8$\times$ &1-32  &100.0\%  &22.3 & 31.6 & 31.5 & 34.7 & 38.2 & 37.4 & 32.6 & 32.0 & 39.1 & 45.3 & 36.9 & 41.4 & 41.2 & 39.3 \\ 
\headercolorSLAM
\ \ \ \ \ + \ours    &  \phantom{0}1.0$\times$  &1-5\phantom{0} &\phantom{0}15.6\%  &\textbf{47.9} & \ul{50.6} & \textbf{51.8} & \textbf{54.6} & \textbf{58.7} & \textbf{59.0} & \textbf{53.8} & \textbf{48.4} & \textbf{52.6} & \textbf{54.8} & \textbf{50.2} & 56.1 & \textbf{60.7} & \textbf{53.8} \\

\bottomrule[1.0pt]
\end{tabular}
  }

  \caption{The Accuracy (\%) on the MMBench and MMMB benchmarks. ``Avg.'' denotes the average accuracy across six languages. ``Training Cost'' refers to the time required to train the models. ``Training Layers'' specifies the decoder layers selected for trainig. ``Trained Param.'' indicates the proportion of trainable parameters in the LLM backbone. \textbf{Bold} and \ul{underline} numbers indicate the best performance and second performance among each group.}
  \label{table:main-exp}
\end{table*}

\subsubsection{Prompting-based Methods}
We select two established prompting strategies that enhance the multilingual capabilities of LVLMs via implicit and explicit translation, respectively. 

\paragraph{Implicit Translation Prompting (ITP)~\citep{shao2024visual}.}
To enhance the model's ability across multilingual scenarios, we prompt the model to \textit{implicitly translate} non-English questions into English, enabling it to \textit{think in English}.

\paragraph{Explicit Translation Prompting (ETP)~\citep{qin2023cross}.}
This approach incorporates a two-stage prompting strategy: first prompting the model to \textit{explicitly translate} non-English questions into English, then solving the task in English. This explicit translation mechanism has shown effectiveness in improving multilingual performance.


  

\subsubsection{Training-based Methods}
\paragraph{Multilingual Supervised Fine-tuning (\textsc{M-SFT})} further fine-tunes the visual instruction-tuned model with multilingual visual-instruction data, aiming to achieve better multilingual alignment.

\paragraph{\textsc{QAlign}~\citep{zhu2024QALign}} incorporates a question alignment stage, where the model is first trained to translate non-English image-question pairs into English. Then, the model is further fine-tuned using English-only visual-instruction data, effectively leveraging the acquired language translation capabilities for visual instruction alignment.

\subsection{Implementation Details}
Due to the scarcity of multilingual visual instruction training data, we first sample English image-text pairs from the ShareGPT4V dataset \citep{chen2024sharegpt4v} and then translate the English questions into five other languages using GPT-4 to construct our training data.
During the fine-tuning process, we only train the first five decoder layers of the visual instruction-tuned models. For more details, please refer to Appendix~\ref{sec:ours_training_details}.

\section{Experimental Results}
We present the main results of LLaVA-1.5-7B/13B and Qwen-VL-Chat on MMBench and MMMB benchmarks in Table~\ref{table:main-exp}. The results of LLaVA-1.6-7B/13B are presented in Table~\ref{table:LLaVA-1.6-7-13B} in Appendix~\ref{sec:appendix_llava_1.6}.

\paragraph{LVLMs exhibit significant performance imbalances in multilingual scenarios.}
As shown in Table~\ref{table:main-exp}, while the LLaVA-1.5-7B model achieves a strong performance of 64.7 on the MMBench and 66.2 on the MMMB in English, its performance significantly declines in moderately low-resource languages, such as Arabic. Specifically, it shows a decrease of \textbf{46.5\%} (64.7 $\rightarrow$ 34.6) on the MMBench and \textbf{37.0\%} (66.2 $\rightarrow$ 41.7) on MMMB. 
These performance imbalances are notably exacerbated in prompting-based strategies.
Both implicit and explicit prompting strategies struggle to improve performance and can even result in substantial decline. Notably, ITP results in an average performance decrease of \textbf{21.7\%} in Arabic and \textbf{10.2\%} in Turkish among the three models, primarily due to the translation inaccuracies triggered by the model's inferior performance in relatively low-resource languages.

\paragraph{\ours enhances multilingual performance and shows robust cross-model generalization.} 
It can be observed that \ours consistently outperforms all baselines on both MMBench and MMMB, achieving \textit{state-of-the-art} performance. Specifically, it achieves an average improvement of  8.0\% and 4.0\% on MMBench and MMMB respectively across the three evaluation models.
These results highlight the effectiveness of \ours in enhancing the multilingual capabilities of LVLMs.
Remarkably, \ours leads to substantial gains for moderately low-resource languages, with improvements of \textbf{12.0\%} (41.7 $\rightarrow$ 46.7) in Arabic and \textbf{16.2\%} (43.1 → 50.1) in Turkish on the MMMB for the LLaVA-1.5-7B model.
This suggests that \ours, by leveraging only a modest amount of question-translation data, effectively bridges language gaps to some extent.
In addition, among all evaluated models of varying architecture and scales, \ours consistently delivers improvements, highlighting its strong generalizability across different models.

\begin{figure}[t!]
    \centering
    \definecolor{upurple}{RGB}{155,89,182}
\definecolor{ublue}{RGB}{52,152,219}
\definecolor{ured}{RGB}{231,76,60}
\definecolor{udark}{RGB}{77,153,77}
\definecolor{ugreen}{RGB}{46,204,113}
\definecolor{c1purple}{HTML}{CE93D8}
\tikzset{global scale/.style={
    scale=#1,
    every node/.append style={scale=#1}
  },
      axis break gap/.initial=0mm
}

\pgfplotsset{
   width=0.4\textwidth,
   height=0.2\textheight,
   symbolic x coords={0,1,2,3,4,5,6,7,8,9,10,11},
   enlarge y limits={upper,value=0.05},
   grid=major,
   legend style={
      fill,
      at={(0.50,16.5em)},
      legend columns=3,
      legend cell align=left,
      anchor=south
      },
   major grid style={dotted},
   }
\begin{tikzpicture}
\useasboundingbox (13em,-3em) rectangle (3.5em,10em);
\begin{axis}[
global scale = 0.63,
   at={(1,1em)},
    name=bottom axis,   
    legend cell align=left,
    width=0.4\textwidth,
   height=0.1\textheight,
    xmin = 0, xmax = 11,     
    xtick={0,1,2,3,4,5,6,7,8,9,10,11},
    xticklabels={,1,2,3,4,5,6,7,8,R,A,},
    ymin = 27, ymax = 29.5,   
    ytick={27,28,29,30,31,32},  
     yticklabel style={/pgf/number format/fixed,rotate=0},
    grid = major,
    axis x line*=bottom,
]
\addplot [sharp plot,ublue,smooth,ultra thick,line width=1pt,mark=pentagon*,mark size=2.5pt,thick,mark options={fill=white,draw=ublue,line width=0.5pt}]
    coordinates {
(1, 51.00)
(2, 53.32)
(3, 53.07)
(4, 54.47)
(5, 56.77)
(6, 56.02)
(7, 48.40)
(8, 48.12)
(9, 37.10)
(10, 28.23)

    };

\end{axis}

\begin{axis}[
    global scale = 0.63,
    at=(bottom axis.north),
    legend entries={MMBench},
    anchor=south, yshift=\pgfkeysvalueof{/tikz/axis break gap},
    width=0.4\textwidth,
   height=0.25\textheight,
    xmin = 0, xmax = 11,     
    xtick={0,1,2,3,4,5,6,7,8,9,10,11},
    xticklabels={,1,2,3,4,5,6,7,8,R,A,},
    ymin = 34, ymax = 60,   
    ytick={36,38,40,42,44,46,48,50,52,54,56,58,60}, 
    yticklabels={36,,40,,44,,48,,52,,56,,60},
    legend style={draw=none,
    line width=0.5pt,
    at={(9.2em,11.5em)},
    anchor=south,
    font=\small,        
    /tikz/mark size=1.2pt,                 
    legend image post style={scale=0.6} },
    grid = major,
    title style={yshift=-1ex, text centered},  
    axis x line*=top,
     yticklabel style={/pgf/number format/fixed,rotate=0},
    xticklabel=\empty,
    after end axis/.code={
         \draw (rel axis cs:0,0) +(-2mm,-1mm) -- +(2mm,1mm)
              ++(0pt,-\pgfkeysvalueof{/tikz/axis break gap})
              +(-2mm,-1mm) -- +(2mm,1mm)
              (rel axis cs:0,0) +(0mm,0mm) -- +(0mm,0mm)
              ++(0pt,-\pgfkeysvalueof{/tikz/axis break gap})
              +(-2mm,-1mm) -- +(2mm,1mm);
              \draw (rel axis cs:0,0) +(-2mm,0mm) -- +(2mm,2mm)
              ++(0pt,-\pgfkeysvalueof{/tikz/axis break gap})
              +(-2mm,-1mm) -- +(2mm,1mm)
              (rel axis cs:0,0) +(0mm,0mm) -- +(0mm,0mm)
              ++(0pt,-\pgfkeysvalueof{/tikz/axis break gap})
              +(-2mm,-1mm) -- +(2mm,1mm);
    }]

\addplot [sharp plot,ublue,ultra thick,line width=0.5pt,mark=pentagon*,mark size=2.5pt,thick,mark options={fill=white,draw=ublue,line width=0.5pt}]
    coordinates {
(1, 51.00)
(2, 53.32)
(3, 54.47)
(4, 56.02)
(5, 56.77)
(6, 53.07)
(7, 48.40)
(8, 48.12)
(9, 52.75)
(10, 30.13)

    };

\end{axis}

\node at (4em,-0.5em){\footnotesize(a) End Training Layer};
\node at (13.6em,-0.5em){\footnotesize (b) End Training Layer};

\begin{axis}[
global scale = 0.63,
   at={(10em,1em)},
    name=bottom axis,   
    legend cell align=left,
    width=0.4\textwidth,
   height=0.1\textheight,
    xmin = 0, xmax = 11,     
    xtick={0,1,2,3,4,5,6,7,8,9,10,11},
    xticklabels={,1,2,3,4,5,6,7,8,R,A},
    ymin = 27, ymax = 29.5,   
    ytick={27,28,29,30,31,32},        
    grid = major,
    axis x line*=bottom,
]
\addplot [sharp plot,ured,ultra thick,line width=0.5pt,mark=pentagon*,mark size=2.5pt,thick,mark options={fill=white,draw=ured,line width=0.5pt}]
    coordinates {
(1, 49.73)
(2, 51.62)
(3, 51.10)
(4, 54.47)
(5, 55.84)
(6, 54.97)
(7, 48.48)
(8, 46.51)
(9, 36.72)
(10, 28.57)
    };

\end{axis}

\begin{axis}[
    global scale = 0.63,
    at=(bottom axis.north),
    legend entries={MMMB},
    anchor=south, yshift=\pgfkeysvalueof{/tikz/axis break gap},
    width=0.4\textwidth,
   height=0.25\textheight,
    xmin = 0, xmax = 11,     
    xtick={0,1,2,3,4,5,6,7,8,9,10,11},
    xticklabels={,1,2,3,4,5,6,7,8,R,A,},
    ymin = 34, ymax = 60,   
    ytick={36,38,40,42,44,46,48,50,52,54,56,58,60}, 
    yticklabels={36,,40,,44,,48,,52,,56,,60},
    grid = major,
    legend style={draw=none,
    line width=0.5pt,
    at={(9.1em,10.9em)},
    anchor=south,
    font=\small,        
    /tikz/mark size=1.2pt,                 
    legend image post style={scale=0.6} },,
    title style={yshift=-1ex, text centered},  
    axis x line*=top,
    xticklabel=\empty,
    after end axis/.code={
         \draw (rel axis cs:0,0) +(-2mm,-1mm) -- +(2mm,1mm)
              ++(0pt,-\pgfkeysvalueof{/tikz/axis break gap})
              +(-2mm,-1mm) -- +(2mm,1mm)
              (rel axis cs:0,0) +(0mm,0mm) -- +(0mm,0mm)
              ++(0pt,-\pgfkeysvalueof{/tikz/axis break gap})
              +(-2mm,-1mm) -- +(2mm,1mm);
              \draw (rel axis cs:0,0) +(-2mm,0mm) -- +(2mm,2mm)
              ++(0pt,-\pgfkeysvalueof{/tikz/axis break gap})
              +(-2mm,-1mm) -- +(2mm,1mm)
              (rel axis cs:0,0) +(0mm,0mm) -- +(0mm,0mm)
              ++(0pt,-\pgfkeysvalueof{/tikz/axis break gap})
              +(-2mm,-1mm) -- +(2mm,1mm);
    }]

\addplot [sharp plot,ured,ultra thick,line width=0.5pt,mark=pentagon*,mark size=2.5pt,thick,mark options={fill=white,draw=ured,line width=0.5pt}]
    coordinates {
(1, 49.73)
(2, 51.62)
(3, 51.10)
(4, 54.47)
(5, 55.84)
(6, 54.97)
(7, 48.48)
(8, 46.51)
(9, 51.66)
(10, 31.57)

    };

\end{axis}
   \node  [rotate=90] at(-1.3em,5em){\scriptsize Accuracy (\%)};
   \node  [rotate=90] at(8.6em,5em){\scriptsize Accuracy (\%)};
\end{tikzpicture}
    \vspace{-1em}
    \caption{ 
  Average accuracy across different training layers.
  The $x$-axis signifies training the decoder layers from the first up to the specified layer. ``R'' denotes randomly selected layers, and ``A'' denotes all layers.}
  \label{fig:ACCperlayer}
\end{figure}
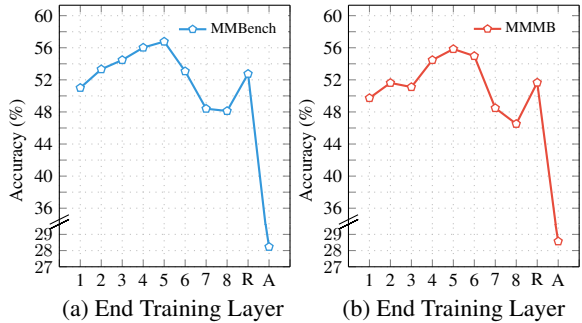

\paragraph{\ours demonstrates superior efficiency.} 
Unlike training-based baselines that require extensive multilingual image-text data for full-parameter fine-tuning, \ours achieves remarkable efficiency by utilizing less than half of the training data (see Table~\ref{table:dataset}) and selectively fine-tunes only shallow decoder layers.
As detailed in Table~\ref{table:main-exp}, \ours fine-tunes merely \textbf{15.6\%} and \textbf{12.5\%} of the LLM backbone's parameters in 7B and 13B models, respectively, resulting in a reduction of the training time by \textbf{7.3$\times$} and \textbf{12.5$\times$}.
Additionally, compared to the two-stage \textsc{QAlign}, \ours effectively mitigates catastrophic forgetting during continual fine-tuning by precisely targeting the layers responsible for multilingual understanding. This targeted fine-tuning leads to significant improvements, with average performance gains of \textbf{42.5\%} and \textbf{26.3\%} on MMBench and MMMB benchmarks, respectively.

\section{Ablation Study and Further Analysis}
We conduct extensive ablation studies and analysis to verify the effectiveness of \ours. For more analysis, please refer to Appendix~\ref{sec:appendix_further_analysis}. 
\paragraph{Effect of layer selection strategy.} 
To validate the necessity of the layer selection, we
conduct ablation studies by training different layers within LVLMs.
Our method identifies the top five layers as multilingual understanding layers by monitoring the dynamics of language-specific neurons using 100 parallel image-question pairs.
As depicted in Figure~\ref{fig:ACCperlayer}, \ours achieves the highest average accuracy on both the MMBench and MMMB. 
Training with an insufficient number of layers hinders the model's ability to understand multilingual questions effectively, whereas excessive layers impair the model's general capabilities, resulting in a decrease in accuracy.
These results highlight that multilingual understanding is predominantly localized in the shallow layers of LVLMs.
Precisely selecting layers that are actively involved in multilingual comprehension is essential for effectively enhancing the multilingual abilities of LVLMs without compromising their general abilities.

\begin{table}[t!]
\small
    \centering
    \resizebox{\linewidth}{!}{

\begin{tabular}{lccccc}
\toprule

\multirow{1}{*}{\textbf{Method}} &  \multirow{1}{*}{\textbf{Hi}} &  \multirow{1}{*}{\textbf{Iw}} &   \multirow{1}{*}{\textbf{Ro}} & \multirow{1}{*}{\textbf{Th}} & \multirow{1}{*}{\textbf{Avg.}}\\

\midrule
\textbf{\textit{LLaVA-1.5-7B}} & \phantom{0}5.8 & \phantom{0}\ul{7.0} & 21.9 & 14.6 & 12.3 \\
\ \ \ \ \ + ITP & \phantom{0}1.7 &\phantom{0}3.5 &\phantom{0}2.8 & 12.3 & \phantom{0}5.1 \\
\ \ \ \ \ + ETP &\phantom{0}9.5 & \phantom{0}6.7 & \phantom{0}4.8 &\phantom{0}8.6 &\phantom{0}7.4 \\
\ \ \ \ \ + M-SFT & \phantom{0}8.3 & \phantom{0}3.0 & \ul{34.3} & \ul{18.9} & \ul{16.1} \\
\ \ \ \ \ + \textsc{QAlign} & \phantom{0}\ul{9.8} & \phantom{0}4.3 & 19.8 & 12.7 & 11.7 \\
\headercolorSLAM
\ \ \ \ \ + \ours (Ours) & \textbf{10.7} & \textbf{10.2} & \textbf{36.7} & \textbf{20.8} & \textbf{19.6} \\

\bottomrule

\end{tabular}
  }
  \caption{The accuracy (\%) on the MaXM benchmark.}
  \label{table:maxm-results}
\end{table}

\paragraph{Extend to truly lower-resource languages.}
Given that MMBench and MMMB primarily encompass medium-resource languages, we further evaluate \ours on MaXM benchmark~\citep{changpinyo2023maxm} to demonstrate its generalizability across a more diverse language families, particularly for low-resource languages. Specifically, we select Hindi (Hi), Hebrew (Iw), Romanian (Ro), and Thai (Th) for evaluation, with results presented in Table~\ref{table:maxm-results}. The empirical results reveal that \ours achieves substantial performance improvements of 45.7\% and 67.6\% improvements for Iw and Ro, respectively, compared to the baseline models. These significant gains demonstrate the strong generalizability of our approach in truly low-resource settings. More details refer to Appendix~\ref{ssec:appendix_maxm}.

\paragraph{Generalization to complex reasoning tasks.}
To evaluate the cross-task transferability of \ours, we conduct experiments on M5-VGR~\citep{schneider2024m5benchmark}, a challenging visually grounded reasoning benchmark spanning multiple languages. As illustrated in Table~\ref{table:reasoning_results}, \ours shows substantial performance enhancements on complex reasoning tasks, particularly for low-resource languages, \eg, Hi and Th, achieving an average improvement of 51.3\% over the LLaVA-1.5-7B model. These findings confirm that the benefits of \ours extend beyond fundamental VQA capabilities to more complex multimodal reasoning scenarios, highlighting its versatility and broader applicability across diverse multilingual task domains. For more experimental details, please refer to Appendix~\ref{ssec:appendix_m5}.

\begin{table}[t!]
\small
    \centering
    \resizebox{\linewidth}{!}{

\begin{tabular}{lccccc}
\toprule

\multirow{1}{*}{\textbf{Method}} &  \multirow{1}{*}{\textbf{Hi}} &  \multirow{1}{*}{\textbf{Th}} &   \multirow{1}{*}{\textbf{Ru}} & \multirow{1}{*}{\textbf{En}} & \multirow{1}{*}{\textbf{Avg.}}\\

\midrule
\textbf{\textit{LLaVA-1.5-7B}} & 38.1 & 38.8 & 41.3 & \textbf{47.8} & 41.5 \\
\ \ \ \ \ + ITP & 45.1 &51.7 &55.8 & 47.8 & 50.1 \\
\ \ \ \ \ + ETP &\ul{50.0} & \ul{55.0} & \textbf{59.2} &47.8 &\ul{53.0} \\
\ \ \ \ \ + M-SFT & 42.8 & 47.2 & 48.8 & 45.1 & 46.0 \\
\ \ \ \ \ + \textsc{QAlign} & 26.6 & 28.5 & 25.8 & 28.4 & 27.3 \\
\headercolorSLAM
\ \ \ \ \ + \ours (Ours) & \textbf{56.9} & \textbf{59.5} & \ul{58.4} & \ul{46.2} & \textbf{55.3} \\

\bottomrule

\end{tabular}
  }
  \caption{The accuracy (\%) on M5-VGR benchmark.}
  \label{table:reasoning_results}
\end{table}

\paragraph{\ours indeed outperforms other parameter-efficient fine-tuning strategies.}
To further demonstrate the advantage of \ours over other parameter-efficient approaches, we compare with the Low-Rank Adaptation (LoRA) \citep{Hu2022LoRA,Low-Rank-zhong,zhang-etal-2025-uora} strategy. Unlike \ours that selectively fine-tunes all parameters within specific language-specific layers, LoRA trains a small subset of parameters across all layers. As shown in Table~\ref{table:vanilla-SFT-LoRA}, with rank set to 512, \ours requires \textit{fewer} trainable parameters while simultaneously achieving \textit{superior} performance, with improvements of 7.1\% and 6.0\% on MMBench and MMMB benchmarks, respectively.
These results indicate the significance of language-specific layers, which precisely activate the model's multilingual capabilities, effectively preventing performance degradation associated with full-layer fine-tuning, while maintaining parameter efficiency.
For more details, please refer to Appendix~\ref{sec:LoRA_training-details}. 

\begin{table}[t!]
\small
    \centering
    \resizebox{\linewidth}{!}{
%
%
\begin{tabular}{lcccc}
\toprule[1.0pt]
\multirow{2}{*}{\textbf{Method}} &  \multirow{2}{*}{\shortstack[c]{ \textbf{Training} \\ \textbf{Cost}}} &  \multirow{2}{*}{\shortstack{ \textbf{Trained} \\ \textbf{Param.}}} &   \multirow{2}{*}{\textbf{MMBench}} & \multirow{2}{*}{\textbf{MMMB}} \\

 & & & & \\
\midrule

   \textbf{\textit{LLaVA-1.5-7B}}        &  -  & - & 52.6 & 53.8 \\
\ \ \ \ \ + LoRA (r=512) & \phantom{0}4.9$\times$& 19.8\%   &52.3 & 53.2  \\ 
\ \ \ \ \ + \ours & \phantom{0}1.0$\times$&15.6\%   &\textbf{56.8} & \textbf{55.8} \\ 

\midrule
   \textbf{\textit{LLaVA-1.5-13B}}   & - & -  & 59.4 & 58.7  \\
\ \ \ \ \ + LoRA (r=512) & \phantom{0}6.5$\times$ & 15.8\%   &57.9 & 55.4  \\ 
\ \ \ \ \ + \ours & \phantom{0}1.0$\times$& 12.5\%   &\textbf{61.1} & \textbf{59.3} \\

\bottomrule[1.0pt]
\end{tabular}
  }
  \caption{The average accuracy of LoRA training strategy on the MMBench and MMMB test sets. For accuracy of all languages, refer to Table~\ref{table:Lora-detail-acc}.}
  \label{table:vanilla-SFT-LoRA}
\end{table}

\begin{figure}[t!]
    \centering
    \definecolor{upurple}{RGB}{155,89,182}
\definecolor{ublue}{RGB}{52,152,219}
\definecolor{ured}{RGB}{231,76,60}
\definecolor{udark}{RGB}{77,153,77}
\definecolor{ugreen}{RGB}{46,204,113}
\definecolor{upink}{HTML}{fcd4d4}
\definecolor{ucyan}{HTML}{e3eeff}
\definecolor{uedgecyan}{HTML}{6d97e0}
\definecolor{uedgepink}{HTML}{cc0000}
\tikzset{global scale/.style={
    scale=#1,
    every node/.append style={scale=#1}
  }
}

\pgfplotsset{
    width=0.71\textwidth,
   height=0.25\textheight,
   symbolic x coords={0,1,2,3,4,5,6,7,8},
   enlarge y limits={upper,value=0.06},
   legend style={
      fill,
      at={(0,16.5em)},
      legend columns=2,
      legend cell align=left,
      anchor=south
      },
   }
\begin{tikzpicture}
\useasboundingbox (13em,-1em) rectangle (3.5em,9.35em);
    \begin{axis}[
    global scale = 0.63,
      at={(0.42em,1em)},
      legend style={at={(0.41,1)}, anchor=south west},
      legend cell align={left},
      ybar,
      enlarge x limits=0.08,
      xtick align=inside,
      bar width=0.8em,
      xmax=8,
      xmin=0,
      legend style={cells={align=left}},
      xtick=data,
        axis y line*=right,
      nodes near coords align={vertical},
      ymin=0.65,
      ymax=0.9,
      ytick={0.65,0.675,0.7,0.725,0.75,0.775,0.8,0.825,0.85,0.875,0.90},
    yticklabels={0.65,,0.70,,0.75,,0.80,,0.85,,0.90},
      xticklabels={},
      xtick style={draw=none},
      yticklabel pos=right,
      ylabel style={yshift=-3em},
      xlabel style={xshift=4em, yshift=-0.5em},
      yticklabel style={/pgf/number format/fixed,/pgf/number format/fixed},
    axis on top=false,
    nodes near coords, 
    nodes near coords align={vertical},
    axis on top=true,
    every node near coord/.append style={/pgf/number format/fixed,                       
    /pgf/number format/precision=2,                 
    /pgf/number format/fixed zerofill,font=\tiny,xshift=-0.5pt,yshift=-0.6pt}, 
      legend style={draw=none,
        line width=1pt,
        at={(0.5,1.0)},
        anchor=south},
        xtick=data,
        axis on top=false,
      ]
         \addplot[fill=ucyan,draw=uedgecyan, area legend] coordinates { 
(0, 0.80)
(1, 0.83)
(2, 0.79)
(3, 0.81)
(4, 0.80)
(5, 0.81)
(6, 0.83)
(7, 0.85)
(8, 0.87)
 };
        \addplot[fill=upink, draw=uedgepink!50, area legend] coordinates { 
(0, 0.82)
(1, 0.85)
(2, 0.83)
(3, 0.84)
(4, 0.84)
(5, 0.831)
(6, 0.84)
(7, 0.87)
(8, 0.90)
};

    \end{axis}
    \node [rectangle,draw=uedgecyan,fill=ucyan,inner sep=2pt,minimum height=0.5em,minimum width=1em,font=\small,anchor=north,align=center,] (label1) at (9.7em,9.5em){};
    \node [rectangle,draw=uedgepink!50,,fill=upink,inner sep=2pt,minimum height=0.5em,minimum width=1em,font=\small,anchor=north,align=center,] (label3) at (13.3em,9.5em){};
    \node [align=center] (label1_1) at ([xshift=1.8em,yshift=-0.25em]label1.north){\scriptsize before};
    \node [align=center] (label1_3) at ([xshift=1.6em,yshift=-0.25em]label3.north){\scriptsize after};

\begin{axis}[
global scale = 0.63,
   at={(0.42em,1em)},
    legend entries={before,after},
    grid=major,
    enlarge x limits=0.08,
    ymajorgrids=true,
    xmajorgrids=true,
    major grid style={dotted},
    legend style={draw=none,
    line width=1pt,
    at={(5.4em,12.5em)},
    column sep=3pt,
    yshift=-5pt,
    anchor=south,
    },
   axis y line*=left,
   scaled y ticks=false,
    xtick style={draw=none},
    yticklabel pos=left,
    xtick=data,
   xticklabels={1,2,3,4,5,6,7,8,9},
   ymin=0, ymax=0.20,
   ytick={0,0.02,0.04,0.06,0.08,0.10,0.12,0.14,0.16,0.18,0.20},
   yticklabels={0,,0.04,,0.08,,0.12,,0.16,,0.20},
   yticklabel style={/pgf/number format/fixed,/pgf/number format/fixed zerofill,/pgf/number format/precision=2,rotate=0}
   ]
   \addplot[sharp plot,ublue,smooth,thick,line width=0.5pt,mark=*,mark size=2pt,thick,mark options={fill=ublue,draw=ublue,line width=0.5pt}] plot coordinates{
(0, 0.169)
(1, 0.138)
(2, 0.132)
(3, 0.131)
(4, 0.123)
(5, 0.106)
(6, 0.103)
(7, 0.102)
(8, 0.114)
      };
   \addplot[sharp plot,ured,smooth,thick,line width=0.5pt,mark=*,,mark size=2pt,thick,mark options={fill=ured,draw=ured,line width=0.5pt}] plot coordinates {
(0, 0.033)
(1, 0.024)
(2, 0.021)
(3, 0.018)
(4, 0.020)
(5, 0.022)
(6, 0.016)
(7, 0.010)
(8, 0.003)

      };

   \end{axis}

\node at (8.5em,-0.6em){\footnotesize Layer Index};
   \node  [rotate=90] at(-1.5em,4.9em){\scriptsize Mean Squared Deviation};
   \node  [rotate=270] at(18.5em,5em){\scriptsize Overlap Ratio};
   
\end{tikzpicture}
    \caption{The comparison of the average overlap ratio (columns) and the $\mathrm{MSD}$ of activated neurons (curves) per layer in the LLaVA-1.5-7B model.}
    \label{overlap_before_after}
\end{figure}
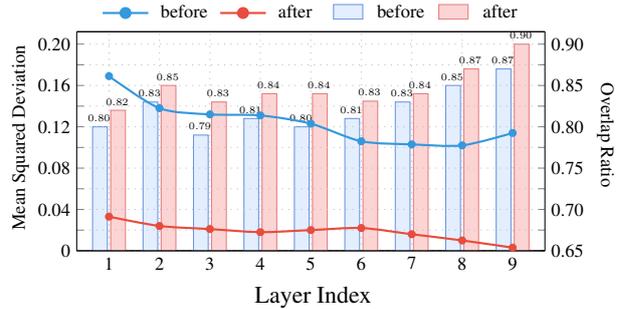

\begin{figure*}[t!]
    \centering
    \includegraphics[width=0.9\textwidth]{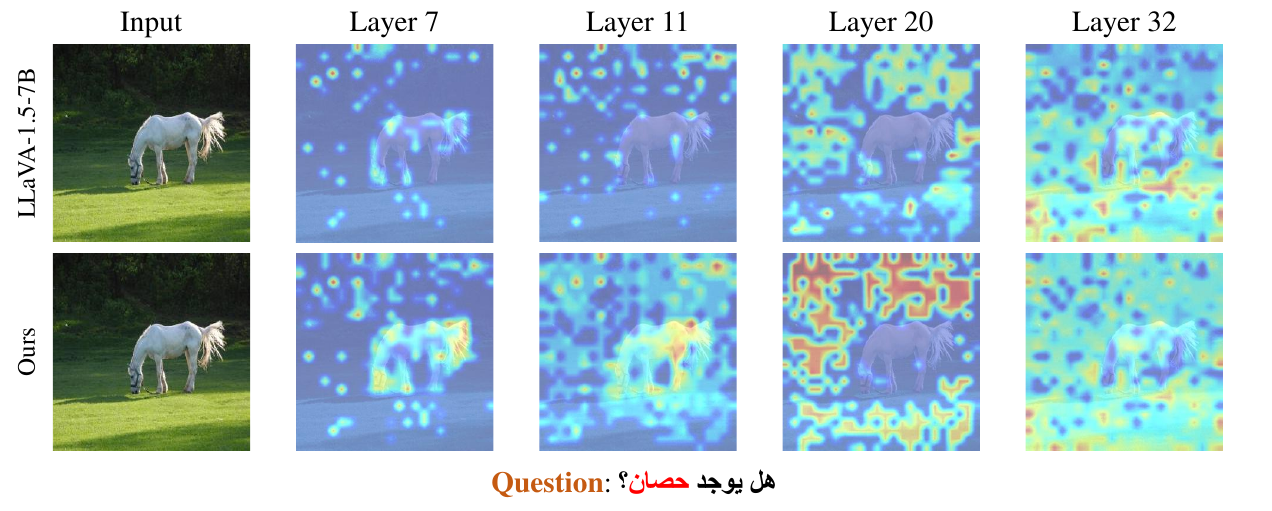}
    \caption{We compare the recognition of the object ``horse'' in images before and after training in LLaVA-1.5-7B using LLaVA-CAM \citep{zhang2024redundancy}, which reveals how attention scores guide the model to focus on relevant image regions during forward propagation based on the given questions. The case comes from the MMBench test sets, and the Arabic question in English means ``\textbf{Is there any \textcolor{red}{horse}}?''. For the visualization of the question in Turkish, and some qualitative case studies of \ours and other baselines, please refer to the Appendix~\ref{subsec:case_study}.}
    \label{fig:attention_score-arabic}
\end{figure*}

\paragraph{Comparison of neuron activation and overlap ratio before and after training.}
To further investigate neuron activation dynamics before and after training, we compute the $\mathrm{MSD^{\textit{i}}}$ using Equation (\ref{eq:msd}) and the average overlap ratio across all non-English languages as specified in Equation (\ref{eq:neuron-overlap}). 
As shown in Figure~\ref{overlap_before_after}, after training, the overlap ratio between non-English and English activated neurons shows a trend of an initial increase, followed by oscillating in the middle, and continuing to rise, which is consistent with the trend before training, representing that the model's representations gradually align with English representations.
Notably, the overlap ratio across layers is significantly higher after training, indicating that \ours effectively facilitates the transition from language-specific representations to English-centered representations in shallow layers.
Moreover, the substantial decrease in $\mathrm{MSD^{\textit{i}}}$ demonstrates that \ours promotes more consistent neuron activation patterns, suggesting enhanced stability during multilingual processing.

\paragraph{Visual attention and semantic representation analysis before and after training.}
To assess the impact of multilingual alignment, we first utilize LLaVA-CAM \citep{zhang2024redundancy} to visualize the information flow from multilingual questions to corresponding image tokens. 
As shown in Figure~\ref{fig:attention_score-arabic}, when processing an Arabic question explicitly referencing a specific visual element (\eg, ``horse''), models trained with \ours demonstrate significantly enhanced attention allocation to the relevant regions in the $7^{\text{th}}$ and $11^{\text{th}}$ shallow layers. 
This indicates that by encouraging alignment between visual inputs and multilingual queries, \ours enables the model to precisely capture language-specific visual features at early layers. 
Furthermore, we also investigate the transformation of linguistic representations through the model using t-SNE projections of language embeddings across various layers. Figure~\ref{fig:Scatter} (see Appendix~\ref{ssec:appendix_tsne}) reveals that the semantic space becomes notably more unified after training with \ours, thereby enabling more effective capability sharing across languages.

\section{Related Work}
With the acceleration of globalization, multilingual LVLMs~\citep{chen2023PaLI,li2023m3it} have gained great attention for their ability to handle multiple languages comprehensively. 
However, due to the training corpora being mainly English-centric, these models perform significantly better in English than in other languages, leading to an imbalanced performance in multilingual scenarios. 

Numerous approaches have been proposed to enhance the multilingual abilities of LVLMs, primarily categorized into prompting-based and training-based methods. 
Prompting-based methods leverage models' inherent understanding capabilities to translate non-English questions into English before generating final responses. For instance, \citet{qin2023cross, zhang2024multimodal} employ either implicit or explicit prompting to guide the model to solve tasks step-by-step in English. 
Conversely, training-based methods focus on synthesizing multilingual training data via machine translation for multilingual visual instruction tuning \citep{2024-Sun-hailong,PALO2024Mu,Geigle2025CenturioOD}.
For example, \citet{2024-Sun-hailong} constructs multilingual image-text pairs to train additional Mixture-of-Experts modules to convert English-biased features to language-specific features for multilingual alignment.

Unlike these approaches, our work focuses on achieving efficient multilingual capability enhancement. Rather than relying on extensive translated multilingual image-text pairs and full-parameter fine-tuning, \ours precisely identifies the specific layers responsible for multilingual understanding, enabling more efficient multilingual alignment while maintaining superior performance.

\section{Conclusion}
This work proposes \ours, a novel training recipe for efficient multilingual enhancement of LVLMs.
\ours first identifies layers predominantly engaged in multilingual understanding by monitoring language-specific neuron activations. These critical layers are then precisely fine-tuned with translation pairs to achieve multilingual alignment.
Extensive evaluations show that \ours significantly enhances LVLM multilingual performance.
Moreover, compared to full-parameter tuning methods, \ours achieves superior performance and efficiency with only 14\% of parameters tuned.
Further analysis confirms its effectiveness across both low-resource languages and more complex visual reasoning tasks, demonstrating its broad applicability in diverse multilingual scenarios.

\section*{Limitations}
\label{sec:limitations}
This work exhibits several limitations worth noting. 
First, although our experiments cover a variety of model types (\eg, the LLaVA series and Qwen-VL-Chat) and model scales (7B/13B), we do not conduct experiments on larger-scale models (larger than 13B) due to limited computing resources. 
We believe that \ours still has great potential and is worth exploring on larger-scale models in future work. 
Second, while \ours achieves comprehensive and efficient multilingual enhancement across on the MMBench, MMMB, MaXM, and M5-VGR benchmarks, the performance still involves trade-offs across different languages. We hypothesize that these trade-offs may arise from imbalances in multilingual training data during the pre-training and visual instruction tuning stages. 
However, our work starts from the perspective of language-specific layers to explore the efficient enhancement of multilingual abilities in LVLMs, rather than from the data-level. In future work, we will explore data-centric strategies for improving multilingual capabilities, especially data selection and data augmentation.

\section*{Ethics Statement}
This work does not require ethical considerations. All the data used in this paper is sourced from open-source materials. Throughout the experimental process, all data and models were strictly utilized following their intended purposes and respective licenses.  
Additionally, this paper may contain offensive text related to the case study. We have all referenced them elliptically and will not present the complete harmful content within the paper.
\label{sec:Ethics}

\section*{Acknowledgements}
This work was supported in part by the National Science
Foundation of China (Nos. 62276056 and U24A20334),
the Fundamental Research Funds for the Central Universities, the Yunnan Fundamental Research Projects (No.
202401BC070021), and the Program of Introducing Talents of Discipline to Universities, Plan 111 (No.B16009).
We would like to thank anonymous reviewers for their valuable comments. 
\bibliography{custom}

\newpage
\appendix
\section{Details of the number of activated neurons and overlap ratio for all models across all layers}
\label{sec:full_visualization}
We respectively present the complete visualization across all layers of LLaVA-1.5-7B/13B, LLaVA-1.6-7B/13B, and Qwen-VL-Chat in Figure~\ref{1.5-7B}-\ref{fig:Qwen-VLchat}.
\section{Experimental Details of baseline methods}
\label{sec:all_baselines}
We conduct experiments on the visual instruction-tuned LLaVA-1.5-Vicuna-7B/13B \citep{2024-Liu} and LLaVA-1.6-Vicuna-7B/13B models \citep{liu2024llavanext}, and Qwen-VL-Chat. The LLaVA series models are equipped with the CLIP Vit-L/336px \citep{Radford2021Learning} as the vision encoder and the Vicuna-1.5 \citep{vicuna2023} as the LLM backbone. And the Qwen-VL-Chat model is equipped with Openclip's ViT-bigG \citep{ilharco2021openclip} as the vision encoder and Qwen-7B-Chat \citep{bai2023qwen} as the LLM backbone.
The detailed baseline implementation is as follows:

\begin{table*}[ht]
    \centering
\resizebox{1.0\linewidth}{!}{
  \begin{tabular}{l|p{0.7\textwidth}}
    \toprule  
    \multirow{2}{*}{\textbf{Training Prompt}} &   Translate this from  [\{\textit{source\_lang}\}] to [\text{English}]:\texttt{\string\n}[\{\textit{source\_lang}\}]: \{\textit{source sentence}\}\texttt{\string\n}[\text{English}]: \{\textit{English sentence}\}  	\\
    
    \bottomrule
  \end{tabular}

}
    \caption{The prompt used to train the decoder layers selected by \ours.}
     \label{tab:Trans-template}
\end{table*}

\subsection{Prompting-based-methods}

\label{sec:prompting-based-methods}

\paragraph{Implicit Translation Prompting (ITP)~\citep{shao2024visual}: }
\begin{table*}[!t]
    \centering
\resizebox{1.0\linewidth}{!}{
  \begin{tabular}{c| l| p{0.7\textwidth}}
    \toprule  
    \multirow{5}{*}{\textbf{MMBench \& MMMB}} 
      & \multirow{2}{*}{\textbf{ITP Prompt}} 
      & Translate this question from [\{\textit{source\_lang}\}] to English and then answer with the option's letter from the given choices directly. \\
    \cmidrule(lr){2-3}
      & \multirow{3}{*}{\textbf{ETP Prompt}} 
      & Stage 1: Translate this question from [\{\textit{source\_lang}\}] to English. \\
      &                                      
      & Stage 2: \{\textit{Translation result}\}\texttt{\string\n}Answer with the option’s letter from the given choices directly. \\
       \midrule                                           
    \multirow{5}{*}{\textbf{MaXM}} 
      & \multirow{2}{*}{\textbf{ITP Prompt}} 
      & Translate this question from [\{\textit{source\_lang}\}] to English. \texttt{\string\n}And then only output the short answer in \{language\}: \\
    \cmidrule(lr){2-3}
      & \multirow{3}{*}{\textbf{ETP Prompt}} 
      & Stage 1: Translate this question from [\{\textit{source\_lang}\}] to English. \\
      &                                      
      & Stage 2: \{\textit{Translation result}\}\texttt{\string\n}And only output the short answer in \{language\}: \\
      \midrule
      \multirow{6}{*}{\textbf{M5-VGR}} 
      & \multirow{2}{*}{\textbf{ITP Prompt}} 
      & Translate the question from [\{\textit{source\_lang}\}] to English. Based on the two images, is it correct? Yes or no? One word answer in English: \\
    \cmidrule(lr){2-3}
      & \multirow{3}{*}{\textbf{ETP Prompt}} 
      & Stage 1: Translate this question from [\{\textit{source\_lang}\}] to English.  	\\
      &                                      
      & Stage 2: Based on the two images, is it correct to say \{\textit{Translation result}\} ? Yes or no? One word answer in English: \\
    \bottomrule
  \end{tabular}
}




    \caption{The prompt used for \textbf{ITP} and \textbf{ETP} baselines of MMBench, MMMB, MaXM, and M5-VGR test sets.}
     \label{tab:ITP-ETF-template}
\end{table*}



\begin{table*}[t!]
    \centering
    \resizebox{\linewidth}{!}{
%
%
\begin{tabular}{lccc|ccccccc|ccccccc}
\toprule[1.0pt]
\multirow{2}[2]{*}{\textbf{Method}} & \multirow{2}[2]{*}{ \shortstack{\\ \textbf{Training} \\ \textbf{Cost}}} & \multirow{2}[2]{*}{ \shortstack{\\ \textbf{Training} \\ \textbf{Layers}}} & \multicolumn{1}{c}{\multirow{2}[2]{*}{ \shortstack{\\ \textbf{Trained} \\ \textbf{Param.}}}} & \multicolumn{7}{c}{\textbf{MMBench}} & \multicolumn{7}{c}{\textbf{MMMB}}\\
    \cmidrule(lr){5-11}
    \cmidrule(lr){12-18}
&  & & \multicolumn{1}{c}{} & \multicolumn{1}{c}{\multirow{1}{*}{Ar}} & \multicolumn{1}{c}{\multirow{1}{*}{Tr}} & \multicolumn{1}{c}{\multirow{1}{*}{Ru}}& \multicolumn{1}{c}{\multirow{1}{*}{Pt}}& \multicolumn{1}{c}{\multirow{1}{*}{Zh}} & \multicolumn{1}{c}{\multirow{1}{*}{En}}& \multicolumn{1}{c}{\multirow{1}{*}{Avg.}} 
& \multicolumn{1}{c}{\multirow{1}{*}{Ar}} & \multicolumn{1}{c}{\multirow{1}{*}{Tr}} & \multicolumn{1}{c}{\multirow{1}{*}{Ru}}& \multicolumn{1}{c}{\multirow{1}{*}{Pt}}& \multicolumn{1}{c}{\multirow{1}{*}{Zh}} & \multicolumn{1}{c}{\multirow{1}{*}{En}}& \multicolumn{1}{c}{\multirow{1}{*}{Avg.}} \\ 
\midrule

   \textbf{\textit{LLaVA-1.6-7B}}    &    &    &   &37.2 & 46.0 & 57.9 & \ul{62.3} & \textbf{60.6} & 68.0 & 55.3 & 40.5 & 44.5 & \ul{62.4} & \ul{60.6} & \textbf{60.1} & 70.1 & 56.4\\
\ \ \ \ \ + ITP &   -&-  &-  &10.6 & 29.3 & 41.6 & 44.0 & 49.4 & 68.0 & 40.5 & 27.2 & 31.0 & 42.4 & 41.6 & 31.7 & 70.1 & 40.7  \\ 
\ \ \ \ \ + ETP &  -&-  &-  &35.3 & 40.2 & \textbf{59.5} & 58.7 & 56.1 & \textbf{68.0} & 53.0 & 43.5 & 47.1 & \textbf{62.8} & \textbf{62.5} & 59.9 & \textbf{70.1} & \ul{57.7} \\ 
\ \ \ \ \ + M-SFT   &  \phantom{0}7.3$\times$  &1-32  &100.0\%  &\ul{41.7} & \textbf{52.0} & 56.6 & 61.8 & 59.6 & 65.6 & 56.2 & \ul{47.4} & \ul{47.9} & 59.8 & 59.4 & 58.3 & 68.4 & 56.8 \\ 
\ \ \ \ \ + \textsc{QAlign}   &   \phantom{0}2.8$\times$  &1-32  &100.0\%  &35.7 & 43.3 & 53.1 & 54.1 & 51.1 & 58.8 & 49.4 & 40.3 & 42.1 & 51.2 & 47.9 & 47.7 & 59.6 & 48.1 \\ 
\headercolorSLAM
\ \ \ \ \ + \ours   &  \phantom{0}1.0$\times$  &1-5\phantom{0}  &\phantom{0}15.6\%  &\textbf{43.3} & \ul{51.9} & \ul{58.1} & \textbf{63.0} & \ul{59.8} & \ul{66.3} & \textbf{57.1} & \textbf{50.5} & \textbf{48.5} & 61.5 & 57.5 & \ul{59.3} & \ul{69.3} & \textbf{57.8} \\ 
\midrule
   \textbf{\textit{LLaVA-1.6-13B}} &  &   &  &45.4 & 52.9 & \textbf{61.9} & 64.1 & \textbf{64.5} & 70.9 & 59.9 & 45.4 & 50.6 & \textbf{67.5} & \textbf{65.6} & \ul{66.8} & 73.5 & 61.6  \\
\ \ \ \ \ + ITP  & - &- &- &23.7 & 36.2 & 49.5 & 43.8 & 58.2 & 70.9 & 47.0 & 31.3 & 28.7 & 40.3 & 36.3 & 46.6 & 73.5 & 42.8  \\ 
\ \ \ \ \ + ETP &  -&-  &-  &42.0 & 49.3 & 60.5 & 62.9 & 60.3 & \textbf{70.9} & 57.7 & 49.0 & \textbf{54.7} & \ul{67.3} & 62.7 & \textbf{67.4} & \textbf{73.5} & \textbf{62.4}  \\
\ \ \ \ \ + M-SFT   &  12.5$\times$  &1-40  &100.0\%  &\ul{47.3} & \ul{54.1} & 60.8 & \ul{64.7} & \ul{63.4} & \ul{69.7} & \ul{60.0} & \ul{50.9} & 51.4 & 67.1 & 64.2 & 64.9 & 72.4 & \ul{61.8} \\ 
\ \ \ \ \ + \textsc{QAlign}    & \phantom{0}4.9$\times$ &1-40  &100.0\%  &39.2 & 49.9 & 53.6 & 55.2 & 51.7 & 59.5 & 51.5 & 48.8 & 48.3 & 57.2 & 51.5 & 52.2 & 62.9 & 53.5 \\ 
\headercolorSLAM
\ \ \ \ \ + \ours    &  \phantom{0}1.0$\times$  &1-5\phantom{0} &\phantom{0}12.5\%  &\textbf{49.4} & \textbf{56.3} & \ul{61.5} & \textbf{65.1} & 62.8 & 68.5 & \textbf{60.6} & \textbf{53.7} & \ul{53.5} & 66.1 & \ul{64.6} & 63.5 & \ul{72.9} & \textbf{62.4} \\

\bottomrule[1.0pt]
\end{tabular}
  }
  \caption{The Accuracy (\%) on the MMBench and MMMB benchmarks. ``Avg.'' denotes the average accuracy across six languages. ``Training Cost'' refers to the time required to train the models. ``Training Layers'' specifies the decoder layers selected for training. ``Trained Param.'' indicates the proportion of trainable parameters in the LLM backbone. \textbf{Bold} and \ul{underline} numbers indicate the best performance and second performance among each group.}
  \label{table:LLaVA-1.6-7-13B}
\end{table*}

\begin{table*}[t!]
    \centering
    \resizebox{\linewidth}{!}{
%
%
\begin{tabular}{lccc|ccccccc|ccccccc}
\toprule[1.0pt]
\multirow{2}[2]{*}{\textbf{Method}} & \multirow{2}[2]{*}{ \shortstack{\\ \textbf{Training} \\ \textbf{Cost}}} & \multirow{2}[2]{*}{ \shortstack{\\ \textbf{Training} \\ \textbf{Layers}}} & \multicolumn{1}{c}{\multirow{2}[2]{*}{ \shortstack{\\ \textbf{Trained} \\ \textbf{Param.}}}} & \multicolumn{7}{c}{\textbf{MMBench}} & \multicolumn{7}{c}{\textbf{MMMB}}\\
    \cmidrule(lr){5-11}
    \cmidrule(lr){12-18}
& & & \multicolumn{1}{c}{} & \multicolumn{1}{c}{\multirow{1}{*}{Ar}} & \multicolumn{1}{c}{\multirow{1}{*}{Tr}} & \multicolumn{1}{c}{\multirow{1}{*}{Ru}}& \multicolumn{1}{c}{\multirow{1}{*}{Pt}}& \multicolumn{1}{c}{\multirow{1}{*}{Zh}} & \multicolumn{1}{c}{\multirow{1}{*}{En}}& \multicolumn{1}{c}{\multirow{1}{*}{Avg.}} 
& \multicolumn{1}{c}{\multirow{1}{*}{Ar}} & \multicolumn{1}{c}{\multirow{1}{*}{Tr}} & \multicolumn{1}{c}{\multirow{1}{*}{Ru}}& \multicolumn{1}{c}{\multirow{1}{*}{Pt}}& \multicolumn{1}{c}{\multirow{1}{*}{Zh}} & \multicolumn{1}{c}{\multirow{1}{*}{En}}& \multicolumn{1}{c}{\multirow{1}{*}{Avg.}} \\ 
\midrule

\textbf{\textit{LLaVA-1.5-7B}}     &  - & -  & -&34.6 & 42.4 & 54.8 & \ul{61.1} & \ul{58.1} & \textbf{64.7} & 52.6 & 41.7 & 43.1 & \ul{55.1} & \textbf{59.2} & \textbf{57.7} & \textbf{66.2} & 53.8 \\
   
\ \ \ \ \ + LoRA (r=512) & \phantom{0}4.9$\times$  & 1-32 & 19.8\%  &\ul{37.5} & \ul{48.3} & \ul{53.1} & 56.2 & 56.9 & 62.0 & \ul{52.3} & \ul{43.8} & \ul{46.6} & 55.2 & 56.3 & 54.6 & 62.8 & \ul{53.2}  \\ 

\ \ \ \ \ + \ours & \phantom{0}1.0$\times$  & 1-5\phantom{0} & 15.6\%  &\textbf{44.4} & \textbf{51.9} & \textbf{58.4} & \textbf{62.3} & \textbf{59.4} & \ul{64.2} & \textbf{56.8} & \textbf{46.7} & \textbf{50.1} & \textbf{59.4} & \ul{57.1} & \ul{56.8} & \ul{65.1} & \textbf{55.8} \\
\midrule
\textbf{\textit{LLaVA-1.5-13B}}   &  - & -  & - &\ul{46.6} & 53.2 & \ul{61.6} & \ul{63.0} & \textbf{63.2} & \textbf{69.0} & \ul{59.4} & \ul{45.9} & \ul{50.7} &\textbf{62.6} & \textbf{61.7} & \textbf{61.6} & \textbf{69.8} & \ul{58.7} \\
   
\ \ \ \ \ + LoRA (r=512) & \phantom{0}6.5$\times$  & 1-40 & 15.8\% &45.8 & \ul{54.3} & 58.5 & 61.9 & 61.2 & 65.9 & 57.9 & 45.8 & 49.3 & 54.2 & 59.8 & 58.1 & 65.4 & 55.4 \\ 

\ \ \ \ \ + \ours &  \phantom{0}1.0$\times$  & 1-5\phantom{0} &12.5\%  &\textbf{51.5} & \textbf{58.7} & \textbf{62.2} & \textbf{64.3} & \ul{62.3} & \ul{67.6} & \textbf{61.1} & \textbf{49.7} & \textbf{53.1} & \ul{61.8} & \ul{61.5} & \ul{60.6} & \ul{69.2} & \textbf{59.3} \\ 

\midrule
   \textbf{\textit{LLaVA-1.6-7B}}        &  -  &  -  & -&37.2 & 46.0 & \ul{57.9} & \ul{62.3} & \textbf{60.6} &  \textbf{68.0} & 55.3 & 40.5 & \ul{44.5} & \textbf{62.4} & \textbf{60.6} & \textbf{60.1} & \textbf{70.1} & \ul{56.4}\\
   
\ \ \ \ \ + LoRA (r=512) & \phantom{0}4.9$\times$  & 1-32 & 19.8\% &\ul{41.6} & \ul{48.9} & 56.2 & 61.0 & 59.4 & \ul{66.4} & \ul{55.9} & \ul{46.3} & 44.8 & 59.1 & \ul{57.0} & 57.9 & 66.0& 55.2  \\ 

\ \ \ \ \ + \ours &  \phantom{0}1.0$\times$ & 1-5\phantom{0} &15.6\%  &\textbf{43.3} & \textbf{51.9} & \textbf{58.1} & \textbf{63.0} & \ul{59.8} & 66.3 & \textbf{57.1} & \textbf{50.3} & \textbf{48.5} & \ul{60.4} & 57.4 & \ul{59.3} & \ul{69.3} & \textbf{57.5} \\ 
\midrule
   \textbf{\textit{LLaVA-1.6-13B}}   &  - & -  & - &\ul{45.4} & \ul{52.9} & \textbf{61.8} & \ul{64.1} & \textbf{64.5} & \textbf{70.9} & \ul{59.9} & 45.4 & \ul{50.6} & \textbf{67.5} & \textbf{65.6} & \textbf{66.8} & \textbf{73.5} & \ul{61.6}  \\
   
\ \ \ \ \ + LoRA (r=512) & \phantom{0}6.5$\times$  & 1-40 & 15.8\% &45.7 & 52.4 & 56.8 & 63.8 & \ul{63.1} & \ul{68.1} & 58.3 & \ul{49.3} & 48.4 & \ul{66.2} & 61.9 & 60.4 & 72.1 & 59.7  \\ 

\ \ \ \ \ + \ours &  \phantom{0}1.0$\times$ & 1-5\phantom{0} &12.5\%  &\textbf{49.4} & \textbf{56.3} & \ul{61.5} & \textbf{65.1} & 62.8 & 68.5 & \textbf{60.6} & \textbf{53.1} & \textbf{53.5} & 66.1 & \ul{64.6} & \ul{63.5} & \ul{72.9} & \textbf{62.3} \\

\bottomrule[1.0pt]
\end{tabular}
  }
  \caption{The Accuracy (\%) on the MMBench and MMMB test sets of LoRA training strategy.}
  \label{table:Lora-detail-acc}
\end{table*}

\begin{table*}[t!]
    \centering
    \resizebox{\linewidth}{!}{
%
%
\begin{tabular}{l|ccccccc|ccccccc}
\toprule[1.0pt]
\multicolumn{1}{l}{\multirow{2}[2]{*}{\textbf{Method}}} & \multicolumn{7}{c}{\textbf{MMBench}} & \multicolumn{7}{c}{\textbf{MMMB}}\\
    \cmidrule(lr){2-8}
    \cmidrule(lr){9-15}
\multicolumn{1}{c}{} & \multicolumn{1}{c}{\multirow{1}{*}{Ar}} & \multicolumn{1}{c}{\multirow{1}{*}{Tr}} & \multicolumn{1}{c}{\multirow{1}{*}{Ru}}& \multicolumn{1}{c}{\multirow{1}{*}{Pt}}& \multicolumn{1}{c}{\multirow{1}{*}{Zh}} & \multicolumn{1}{c}{\multirow{1}{*}{En}}& \multicolumn{1}{c}{\multirow{1}{*}{Avg.}} 
& \multicolumn{1}{c}{\multirow{1}{*}{Ar}} & \multicolumn{1}{c}{\multirow{1}{*}{Tr}} & \multicolumn{1}{c}{\multirow{1}{*}{Ru}}& \multicolumn{1}{c}{\multirow{1}{*}{Pt}}& \multicolumn{1}{c}{\multirow{1}{*}{Zh}} & \multicolumn{1}{c}{\multirow{1}{*}{En}}& \multicolumn{1}{c}{\multirow{1}{*}{Avg.}} \\ 
\midrule

   \textbf{\textit{LLaVA-1.5-7B}}         &34.6 & 42.4 & 54.8 & \ul{61.1} & 58.1 & \textbf{64.7} & 52.6 & 41.6 & 43.1 & 55.1 & \textbf{59.2} & \textbf{57.7} & \textbf{66.2} & \ul{53.8}\\
   
\ \ \ \ \ + \ours  &\textbf{44.4} & \textbf{51.9} & \ul{58.4} & \textbf{62.3} & \textbf{59.4} & \ul{64.2} & \textbf{56.8} & \textbf{46.7} & \textbf{50.1} & \textbf{59.4} & \ul{57.1} & 56.8 & \ul{65.1} & \textbf{55.8} \\

\ \ \ \ \ + w/o MLP   &39.2 & 45.4 & 56.1 & 58.1 & 55.8 & 61.9 & 52.8 & 44.5 & \ul{46.8} & \ul{57.2} & 50.4 & 54.9 & 63.6 & 52.9  \\ 

\ \ \ \ \ + w/o Attention   &\ul{43.7} & \ul{49.4} & \textbf{58.6} & 60.9 & \ul{58.5} & 63.2 & \ul{55.7} & \ul{46.3} & 45.2 & 50.7 & 56.2 & \ul{56.9} & 64.1 & 53.3 \\

\bottomrule[1.0pt]
\end{tabular}
  }
  \caption{The Accuracy (\%) of training different sub-layers on MMMB and MMBench test sets across all languages.}
  \label{table:ablation-diff-module}
\end{table*}

For each question in the MMBench and MMMB test sets, ITP implicitly prompts LVLMs to first translate the non-English questions into English before reasoning in English. 
During evaluation, we utilize the VLMEvalKit from OpenCompass \citep{2023opencompass} to evaluate the MMBench and MMMB test sets, and adopt the greedy decoding strategy for all models, setting the maximum generation length to 256.  
The evaluation prompt template used for MMBench and MMMB datasets is shown in Table~\ref{tab:ITP-ETF-template}. In the template, \{\textit{source\_lang}\} can be replaced with any of the following languages: Arabic, Turkish, Russian, Portuguese, Chinese, Hindi, Hebrew, Romanian, and Thai.

\paragraph{Explicit Translation Prompting (ETP)~\citep{qin2023cross}: }
For each question in the MMBench and MMMB test sets, ETP explicitly prompts LVLMs to first translate non-English questions into English, and then solve multimodal tasks with the translated questions. The evaluation prompt template used for MMBench and MMMB datasets is shown in Table~\ref{tab:ITP-ETF-template}. In the template, \{\textit{source\_lang}\} can be replaced with any of the following languages: Arabic, Turkish, Russian, Portuguese, Chinese, Hindi, Hebrew, Romanian, and Thai.
\{\textit{Translation result}\} can be replaced with the translation result by the model itself in the first round of dialogue.

\subsection{Training-based-methods}
\label{sec:training-based-methods}

\paragraph{Multilingual Supervised Fine-tuning (\textsc{M-SFT}):}
This method involves directly full-parameter fine-tuning models with multilingual instruction-following data during the visual instruction tuning stage.
During training, we adopt the training hyperparameter from \citet{2024-Liu}, using the M-ShareGPT4V~\citep{2024-Sun-hailong} dataset. We keep the vision encoder frozen and train the projector and the LLM backbone for one epoch using eight NVIDIA A800 GPUs. The total batch size is set to 128, and the learning rate is maintained at 2e-5. The maximum input sequence length is set to 2,048 tokens. During decoding, we adopt the same decoding hyperparameter as the prompting-based method.

\paragraph{\textsc{QAlign}:}
\citet{zhu2024QALign} proposes a two-stage training strategy to enhance multilingual abilities. 
In the first stage, we train the visual instruction-tuned model using multilingual question translation data paired with images, referred to as M-ShareGPT4V-Q, to translate non-English questions into corresponding English questions that convey the same meaning. In the second stage, to recover the general capabilities that were compromised during the first stage, we employ the English image-text instruction-following data paired with images, used in the first stage, referred to as ShareGPT4V-Sub, for visual instruction fine-tuning.
In both two stages, we freeze the visual encoder layers and fine-tune the projection as well as all the decoder layers in the LLM backbone for one epoch using eight NVIDIA A800 GPUs. The total batch size is set to 128, and the learning rate is set to 2e-5. The maximum input sequence length is set to 2,048 tokens.

 
\section{Experimental Details of \ours}
\label{sec:ours_training_details}

\subsection{Training and Evaluation Datasets}
\label{ssec:appendix_training_dataset}

 \begin{table}[t!]
    \centering

\small
\resizebox{0.49\textwidth}{!}{%
\centering
\begin{tabular}{l l c c r}
\toprule
\textbf{Function} & \textbf{Dataset}  & \textbf{Usage} & \textbf{Lang} & \textbf{Size} \\
\midrule

\multirow{3}{*}{\centering\textbf{Training}}
& M-ShareGPT4V    & M-SFT & Ar, Tr, Ru, Pt, Zh, En & 138,000 \\
& M-ShareGPT4V-Q  & \ours, \textsc{QAlign} & Ar, Tr, Ru, Pt, Zh     & 60,000 \\
& ShareGPT4V-Sub  & \textsc{QAlign} & En                     & 12,000 \\
\midrule
\addlinespace[2pt]
\multirow{4}{*}{\textbf{Evaluation}} 
& MMBench & - & Ar, Tr, Ru, Pt, Zh, En & 25,205  \\
& MMMB    & - & Ar, Tr, Ru, Pt, Zh, En & 11,879  \\
& MaXM    & - & Hi, Iw, Ro, Th & 1092 \\
& M5-VGR & - & Hi, Th, Ru, En & 478 \\
\bottomrule
\end{tabular}
}

  \caption{Dataset Statistics used for training and evaluation. 
  “Usage” indicates the training data used by each method, “Lang” denotes the languages covered, and “Size” denotes the total number of samples.
  ``M-ShareGPT4V'' refers to the multilingual ShareGPT4V dataset~\citep{2024-Sun-hailong}, and ``ShareGPT4V-Sub'' is a subset sampled from ShareGPT4V~\citep{chen2024sharegpt4v}. ``M-ShareGPT4V-Q'' contains question-translation data from ``ShareGPT4V-Sub'' and ``M-ShareGPT4V''.}
  
  \label{table:dataset}
\end{table}
To avoid translation errors caused by overly complex image-text pairs, we select the relatively simple sentence structures from the coco and GQA datasets within the ShareGPT4V dataset. Then we sort these data by length, extracting 7,200 and 4,800 image-text pairs from coco and GQA, respectively, which we refer to as ShareGPT4V-Sub. Subsequently, we translate the questions in ShareGPT4V-Sub into Arabic, Turkish, Russian, Portuguese, and Chinese using GPT-4 \citep{2023-ChatGPT}, followed by manual calibration. Finally, we use these translated questions paired with images to construct X-English question-translation data, referred to as M-ShareGPT4V-Q for \ours training. We provide detailed statistics of the training set in Table~\ref{table:dataset}, including the training data adopted by each method, the total number of samples, and the languages involved. For the main experiments, we select MMBench and MMMB as the evaluation benchmarks. Table~\ref{table:dataset} presents the data volumes and the languages included in MMBench and MMMB test sets. For the analysis experiment, we chose MaXM and M5-VGR as evaluation benchmarks. The specific data volume and included languages in these test sets are shown in Table~\ref{table:dataset}.

\subsection{Training Prompts}
\label{sec:training_prompt}
The prompt template employed for training is shown in Table~\ref{tab:Trans-template}. The prompt explicitly trains models to translate multilingual questions into English. 
In the template, \{\textit{source\_lang}\} can be replaced with any of the following languages: Arabic, Turkish, Russian, Portuguese, and Chinese. The placeholder \{\textit{source sentence}\} is substituted with the multilingual questions, and \{\textit{English sentence}\} is replaced with the corresponding English questions that convey the same meaning.

\subsection{Training Details}

We use LLaVA project\footnote{\href{https://github.com/haotian-liu/LLaVA}{https://github.com/haotian-liu/LLaVA}} as our training framework. Training is conducted on eight NVIDIA A800 GPUs using Deepspeed stage 2 \citep{deepspeed2} for efficient multi-GPU distribution, with training precision set to Bfloat16. We maintain a total batch size of 128, a learning rate of 2e-5, and a maximum input sequence length of 2,048 tokens.
Both the LLaVA-1.5-7B/13B and LLaVA-1.6-7B/13B models are trained over 2 epochs.

\section{Experimental Results of LLaVA-1.6}
\label{sec:appendix_llava_1.6}
We present the complete results of LLaVA-1.6-7B and LLaVA-1.6-13B in Table~\ref{table:LLaVA-1.6-7-13B}.

\section{Additional Analysis}
\label{sec:appendix_further_analysis}
FFN sub-layers in LLMs have been recognized as storing the multilingual knowledge \citep{Knowledge2022Dai,zhao2024how,Tang2024LanguageSpecific}. 
Therefore, we investigate the function of FFN and Attention sub-layers of selected layers by separately trainig them. 
\begin{figure}[t!]
    \centering
    \definecolor{c1}{HTML}{81BECE}
\definecolor{c2}{HTML}{378BA4}
\definecolor{ublue}{HTML}{5fa0d1}  
\definecolor{ured}{HTML}{a9c1f7}   
\definecolor{udpblue}{HTML}{0419fb}
\definecolor{usemiblue}{HTML}{17becf}
    \begin{tikzpicture}
    \tiny{
    \begin{axis}[
      at={(0,12em)},
      legend entries={de2en},
      ymajorgrids,
      xmajorgrids,
      grid style=dashed,
      xbar,
      legend image code/.code={%
                    \draw[#1, draw=none] (0cm,-0.1cm) rectangle (0.6cm,0.1cm);
                }, 
      height=.3\textwidth,
      width=.45\textwidth,
      bar width=1em,
      xlabel={\footnotesize{Accuracy (\%)}}, 
      xticklabels={52.0,53.0,54.0,55.0,56.0,57.0,58.0,59.0},
      xmax=59, 
      legend style={at={(21.2em,11em)}},
      symbolic y coords={ PLAST, {PLAST \\ w/o Attention}, {PLAST \\ w/o MLP}, {LLaVA-1.5-7B}},
      yticklabel style={align=right,font=\tiny},
      ytick=data,
      nodes near coords,
      nodes near coords align={horizontal},
      enlarge y limits=0.2,xticklabel style={/pgf/number format/fixed,/pgf/number format/fixed zerofill,/pgf/number format/precision=1},
      enlarge x limits=0.3,xticklabel style={/pgf/number format/fixed,/pgf/number format/fixed zerofill,/pgf/number format/precision=1},]
    \addplot[fill=ublue!40,draw=ublue,area legend] coordinates {
(52.6,{LLaVA-1.5-7B})
(52.8,{PLAST \\ w/o MLP})
(55.7,{PLAST \\ w/o Attention})
(56.8,PLAST)
};

    \addplot[fill=ured!40,draw=ured,area legend] coordinates {
(53.8,{LLaVA-1.5-7B})
(52.900,{PLAST \\ w/o MLP})
(53.3,{PLAST \\ w/o Attention})
(55.8,PLAST)
};
      \addlegendentry{MMBench}
      \addlegendentry{MMMB\phantom{ch}}
      
    \end{axis}

  }

    \end{tikzpicture}
    \caption{Comparison of the average accuracy of training different sub-layers on MMBench and MMMB. The detailed accuracy of all languages is shown in Table~\ref{table:ablation-diff-module}.}
    \label{fig:diff-sub-layers}
\end{figure}
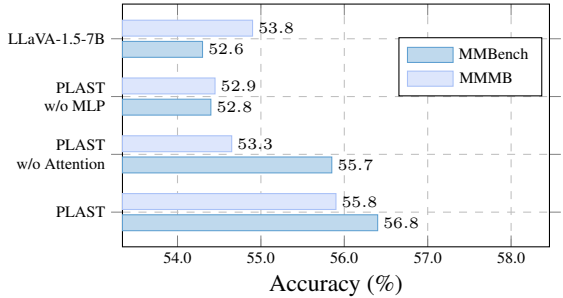
As indicated in Figure~\ref{fig:diff-sub-layers}, compared with only training the FFN and Attention sub-layers, \ours further improves average accuracy by 5.1\% and 4.8\% across MMBench and MMMB test sets, respectively. 
This indicates that \ours, by training the entire selected layer, not only improves the fusion of language information and visual information but also enhances the multilingual understanding abilities of LVLMs.

\section{Experimental Details of Ablation Studies and Analysis}
\subsection{Experimental Details of MaXM Benchmark}
\label{ssec:appendix_maxm}
We follow the main experimental settings and utilize GPT-4 to translate the “ShareGPT4V-Sub” dataset into four low-resource languages included in the MaXM test set (Hindi, Hebrew, Romanian, and Thai) for mixed-language training. The evaluation prompt used for MaXM test set is shown in Table~\ref{tab:ITP-ETF-template}. To ensure a fair comparison and mitigate the risk of hallucinations \cite{Hallucination-lhuang,Citation-lhuang,Faithful-lhuang,Misalignment-lhuang,Self-Improving-lhuang} that can arise from complex multilingual questions, we use evaluation scripts\footnote{\label{fn:scripts}\href{https://github.com/floschne/m5b}{https://github.com/floschne/m5b}} provided by \citet{schneider2024m5benchmark}.

\subsection{Experimental Details of Complex Reasoning Tasks}
\label{ssec:appendix_m5}
Following the main experimental settings, we utilized GPT-4 to translate the “ShareGPT4V-Sub” dataset into Hindi and Thai as included in the M5-VGR benchmark, and combined these translations with Russian training data from “M-ShareGPT4V” for mixed-language training. 
The evaluation prompt used for the M5-VGR test set is shown in Table~\ref{tab:ITP-ETF-template}. For a fair comparison, we use evaluation scripts\footref{fn:scripts} provided by \citet{schneider2024m5benchmark}.

\subsection{Experimental Details of LoRA Training Strategy}
\label{sec:LoRA_training-details}
We use Low-Rank Adaptation (LoRA) \citep{Hu2022LoRA} as an alternative to \textsc{\textbf{M-SFT}}. 
For the LoRA training, we use a rank of 512, and the LoRA target modules are `q\_proj, k\_proj, v\_proj, o\_proj, up\_proj, down\_proj, gate\_proj'. We set a total batch size of 128, a learning rate of 2e-4 for the LLM backbone, a learning rate of 2e-5 for the projector, with a 0.03 linear warmup ratio, and a maximum input sequence length of 2,048 tokens. All the models are trained over 1 epoch. 

\subsection{Experimental Details of t-SNE}
\begin{figure*}[t!]
    \centering
     \includegraphics[height=7cm,width=1\textwidth]{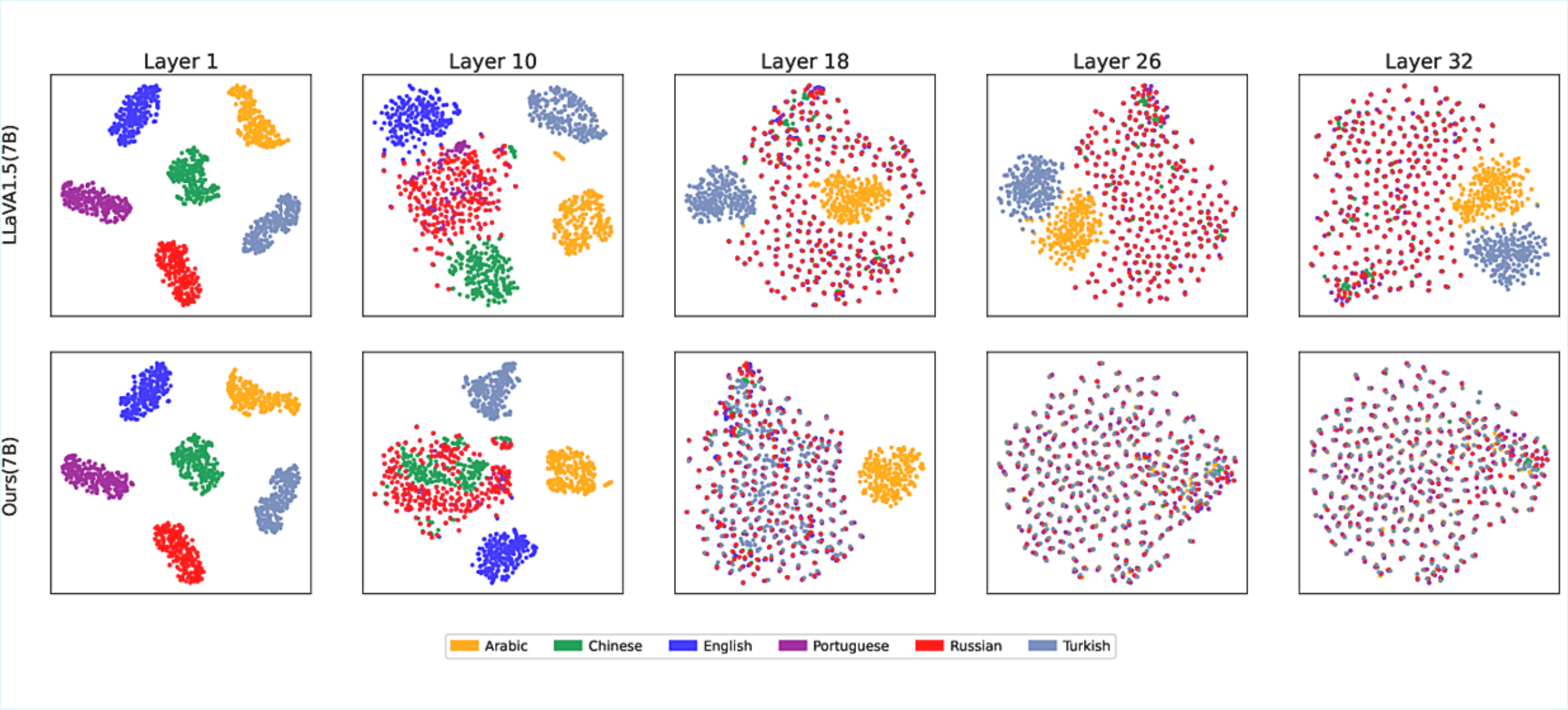}
    \caption {The final token representations from the MMBench test set are visualized using t-SNE for dimensionality reduction. The distributions in LLaVA1.5-7B at the 1\textsuperscript{th}, 10\textsuperscript{th}, 18\textsuperscript{th}, 26\textsuperscript{th}, and 32\textsuperscript{th} layers are compared before and after training. Different colors represent different languages.}
    \label{fig:Scatter}
\end{figure*}

To better understand the model's multilingual capability, we visualize the hidden state representations of the final token in multilingual questions, as it plays a crucial role in guiding the model’s subsequent output \citep{Last-token-important}. Specifically, we randomly sample 250 instances from each of the five languages in the MMBench test set and extract 4096-dimensional hidden state representations from each layer of LLaVA-1.5-7B, both before and after training. 
These representations are then projected into a 2D space using t-SNE for visualization, as illustrated in Figure~\ref{fig:Scatter}.
\label{ssec:appendix_tsne}

\section{Case Study}
\begin{figure*}[t!]
    \centering
    \includegraphics[width=\textwidth]{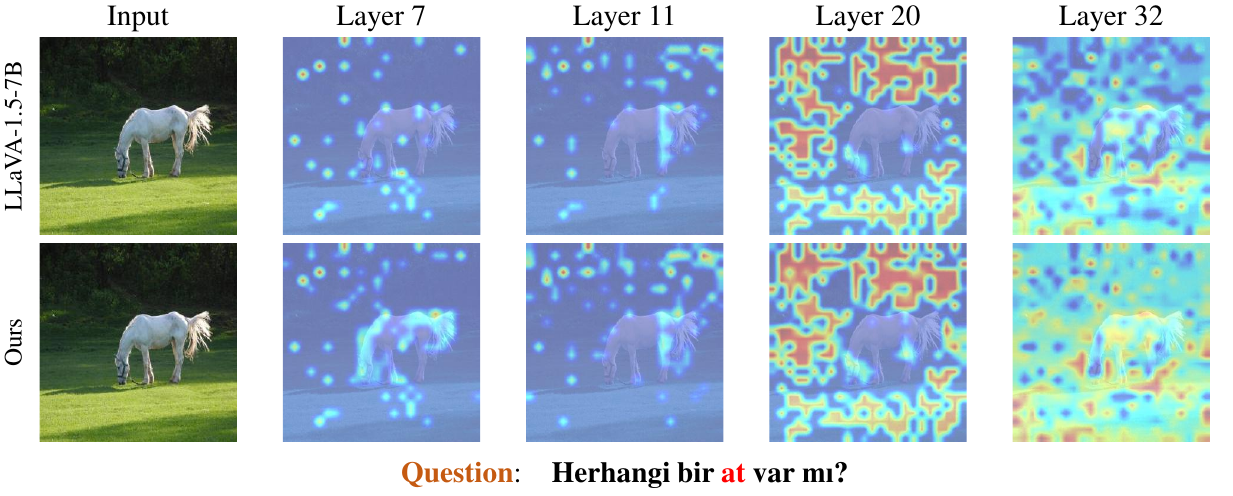}

    \caption{We compare the recognition of the object ``horse'' in images before and after training in LLaVA-1.5-7B using LLaVA-CAM \citep{zhang2024redundancy}, which reveals how attention scores guide the model to focus on relevant image regions during forward propagation based on the given questions. The case comes from the MMBench test sets, and the question in English means ``\textbf{Is there any \textcolor{red}{horse}}?''}
    \label{fig:attention_score-turkish}
\end{figure*}

\begin{figure*}[t!]
    \centering
    \includegraphics[width=\textwidth]{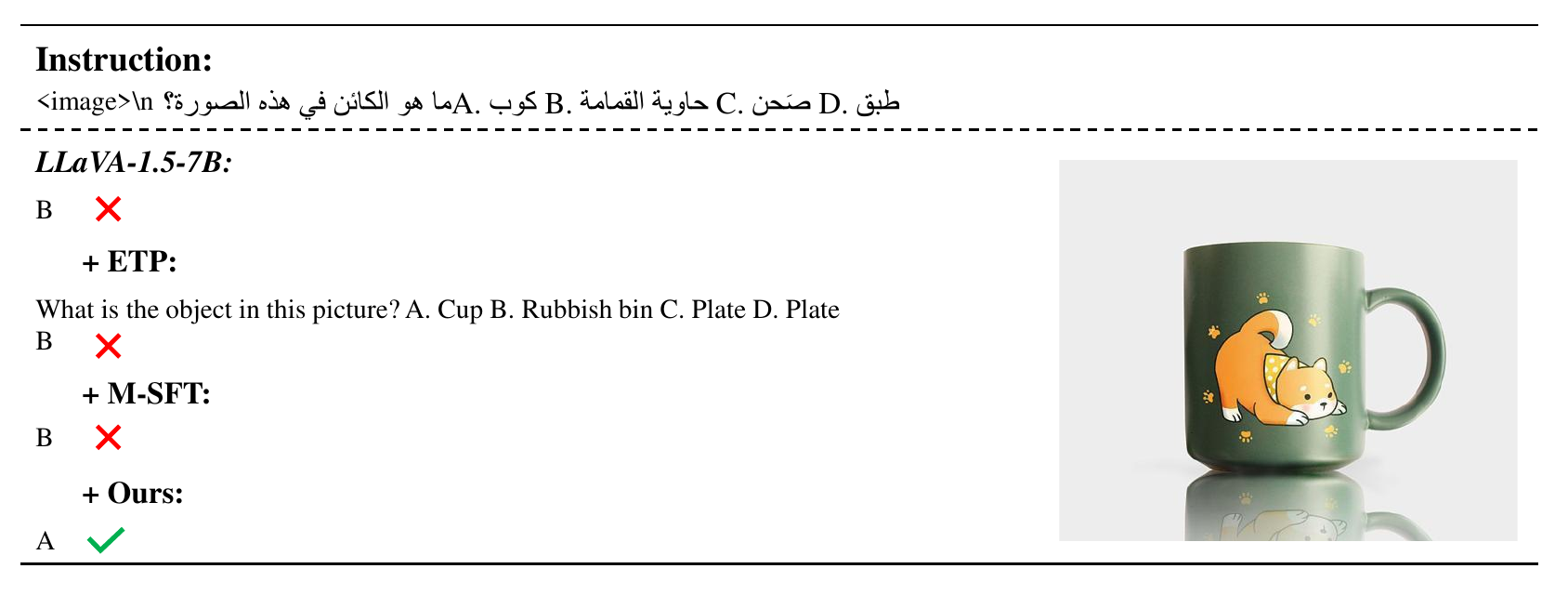}

    \caption{Example where \ours can yield a correct answer compared to other baselines. The case comes from the MMBench Arabic test sets, and the question in English means ``What is the object in this picture? A. Cup B. Trash can C. Bowl D. Plate''.}
    \label{fig:case-study-arabic}
\end{figure*}

\begin{figure*}[t!]
    \centering
    \includegraphics[width=\textwidth]{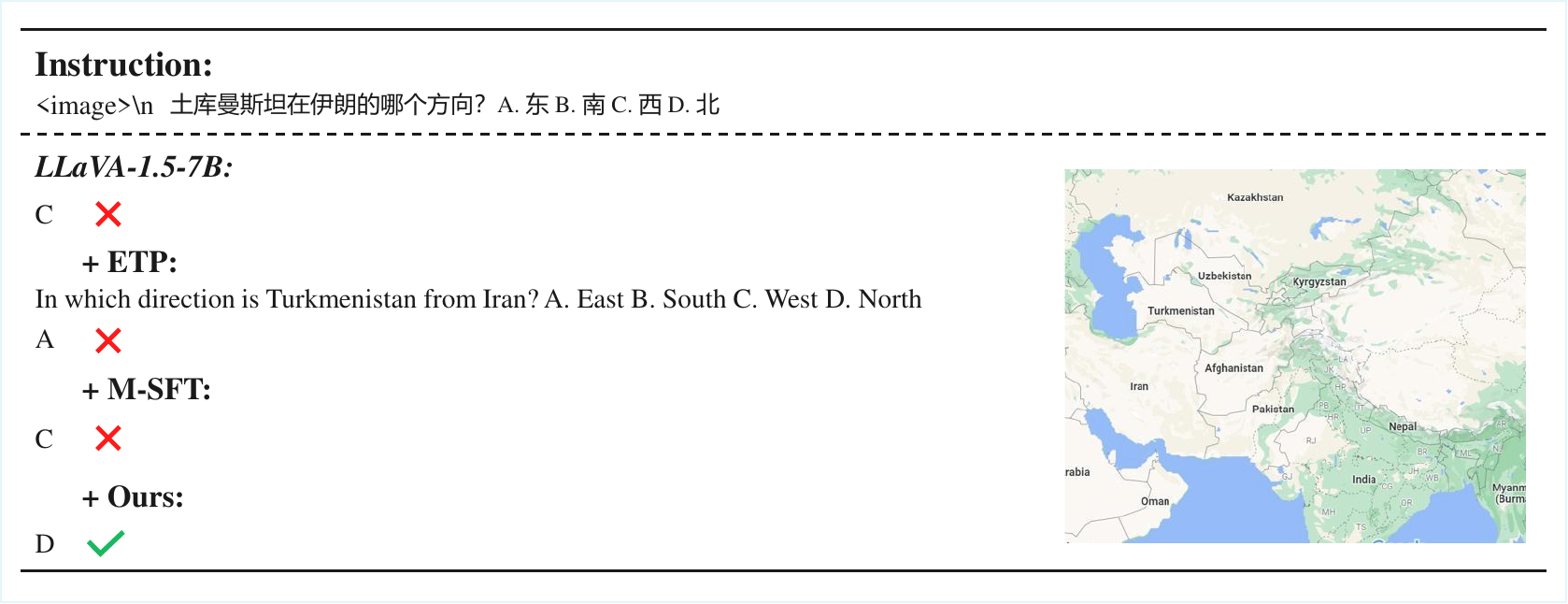}
    \caption{Example where \ours can yield a correct answer compared to other baselines. The case comes from the MMBench Chinese test sets, and the question in English means ``In which direction is Turkmenistan from Iran? A. East B. South C. West D. North''.}
    \label{fig:case-study-chinese}
\end{figure*}

\begin{figure*}[t!]
    \centering
    \includegraphics[width=\textwidth]{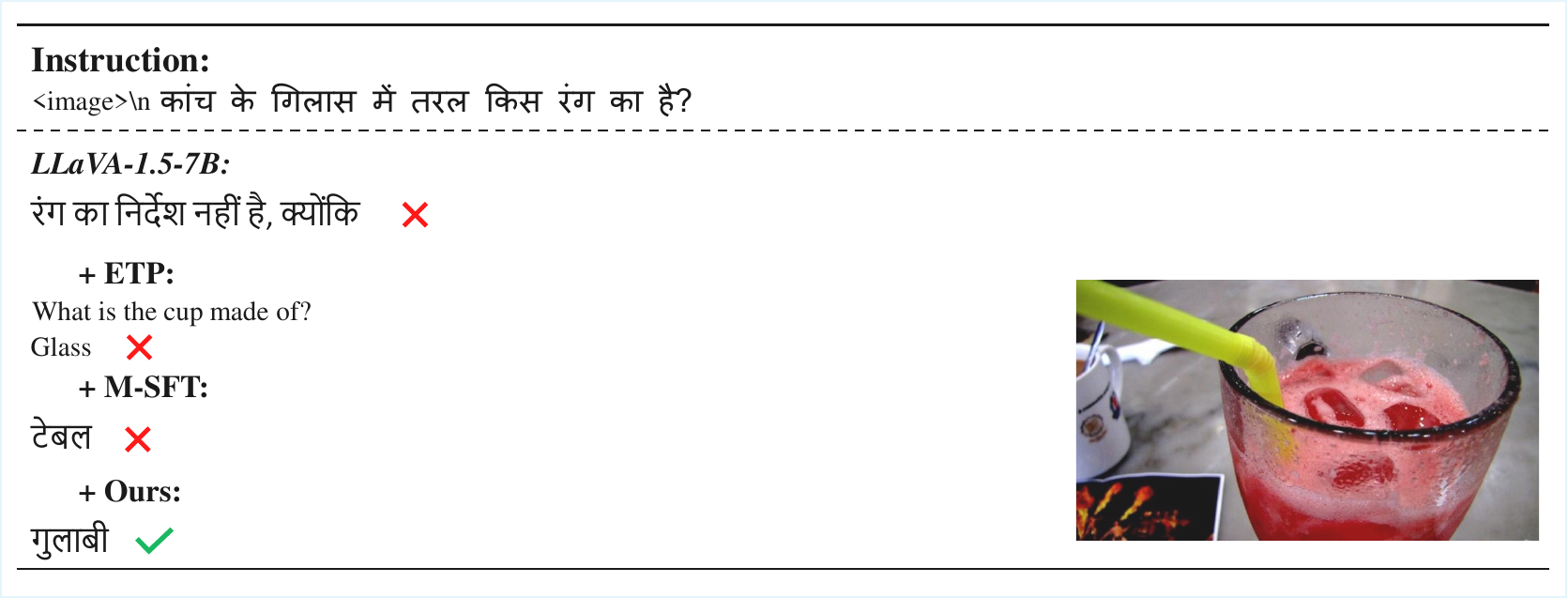}
    \caption{Example where \ours can yield a correct answer compared to other baselines. The case comes from the MaXM Hindi test sets. The question in English means ``What is the color of the liquid in the glass?'', and the correct answer in English means ``pink''.}
    \label{fig:case-study-hindi}
\end{figure*}

\begin{figure*}[t!]
    \centering
    \includegraphics[width=\textwidth]{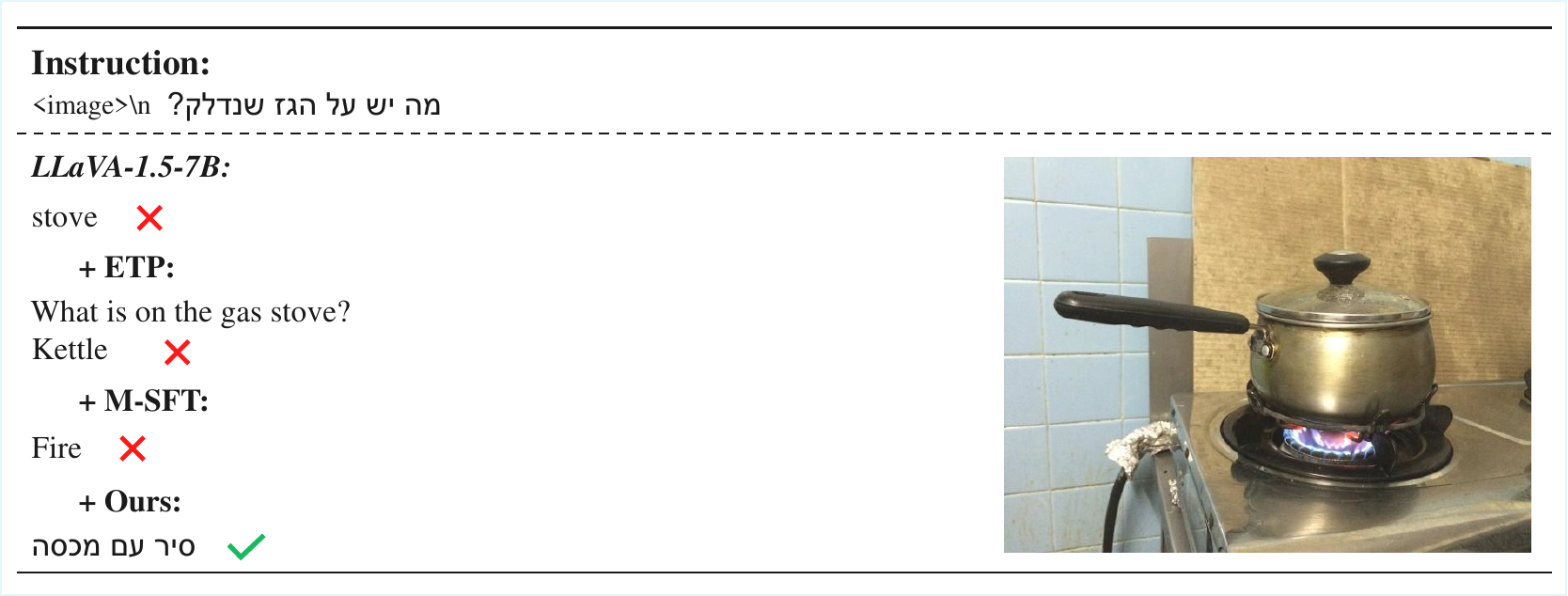}

    \caption{Example where \ours can yield a correct answer compared to other baselines. The case comes from the MaXM Hebrew test sets. The question in English means ``What is on the lit stove?'', and the correct answer in English means ``A pot with a lid''.}
    \label{fig:case-study-iw}
\end{figure*}

The visualization of attention scores for the question in Turkish is shown in Figure~\ref{fig:attention_score-turkish}. Furthermore, we provide several qualitative examples from MMBench test sets in Figure~\ref{fig:case-study-arabic} and Figure~\ref{fig:case-study-chinese} to compare different methods of enhancing the multilingual abilities on LLaVA-1.5-7B. 
As shown in Figure~\ref{fig:case-study-arabic}, ~\ref{fig:case-study-chinese}, the ITP often leads the model to produce mere translations under low-resource language conditions, failing to follow instructions to provide the final answer. In contrast, \ours achieves efficient alignment of multilingual capabilities in LVLMs by facilitating shallow-layer understanding of multilingual instructions and integrating this information with visual features. 
In addition, we provide examples from the MaXM test sets in Figure~\ref{fig:case-study-hindi} and ~\ref{fig:case-study-iw} respectively. As shown in Figure~\ref{fig:case-study-hindi} and ~\ref{fig:case-study-iw}, PLAST not only performs well in multilingual VQA types of multiple-choice questions, but also can follow instructions and answer correctly in multilingual non-multiple-choice question types.

\label{subsec:case_study}
\begin{figure*}[t!]
    \centering
    \includegraphics[width=\textwidth]{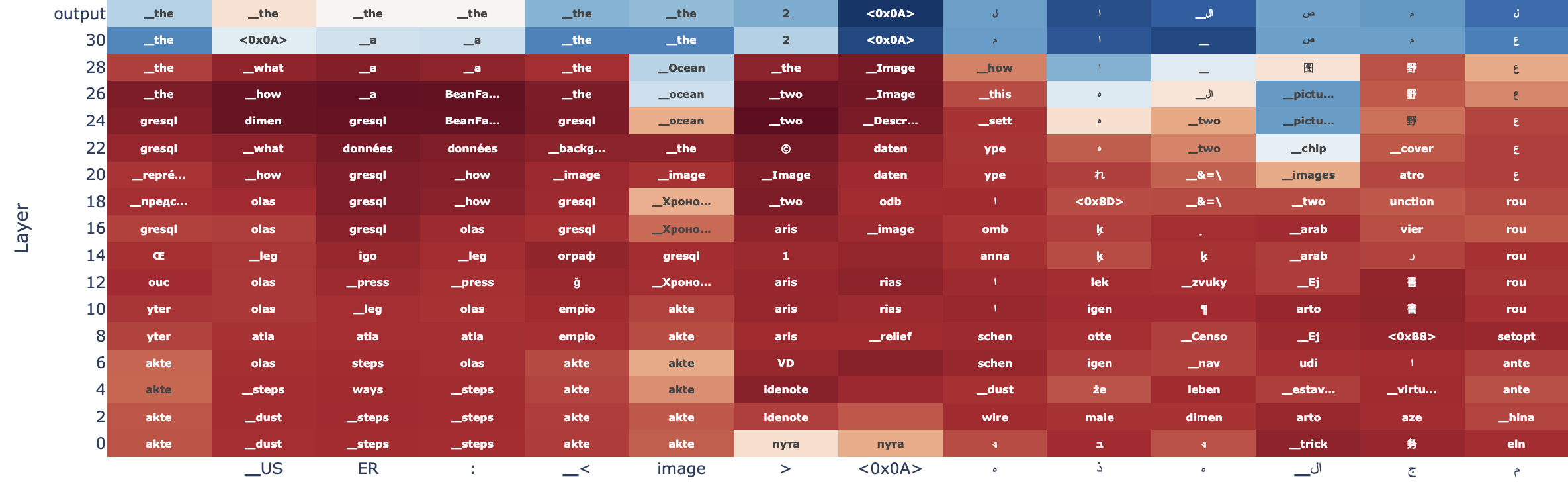}
    \label{fig:f13101}
    
\end{figure*}

\begin{figure*}[t!]
    \centering
    \includegraphics[width=\textwidth]{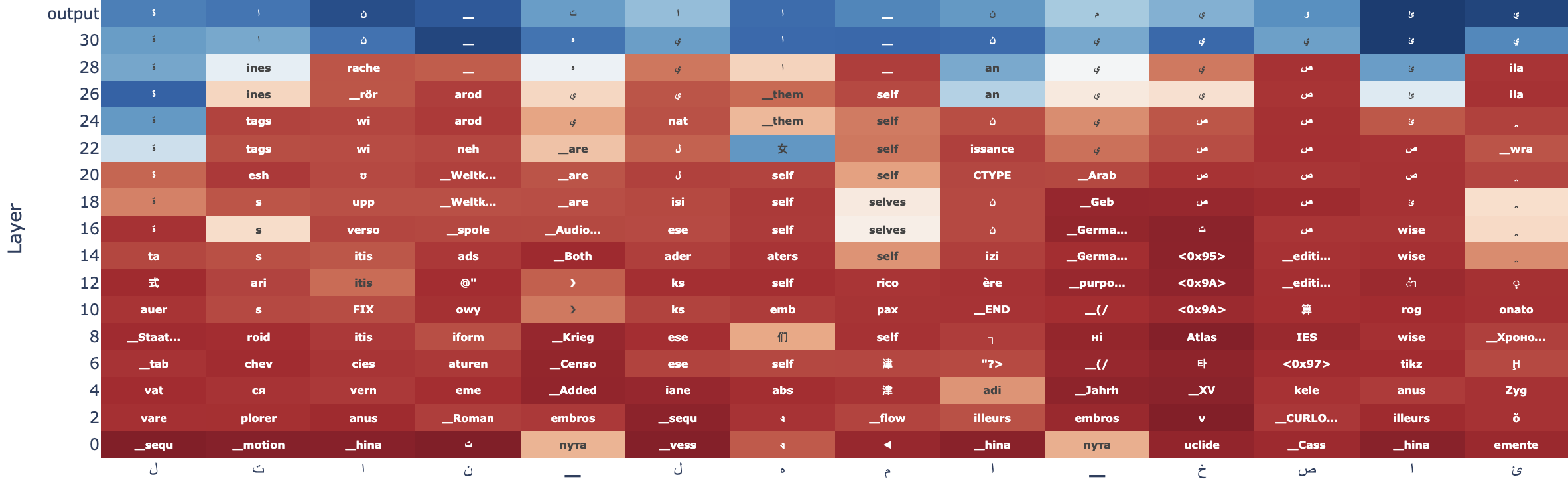}
    \label{fig:f13102}
    
\end{figure*}

\begin{figure*}[t!]
    \centering
    \includegraphics[width=\textwidth]{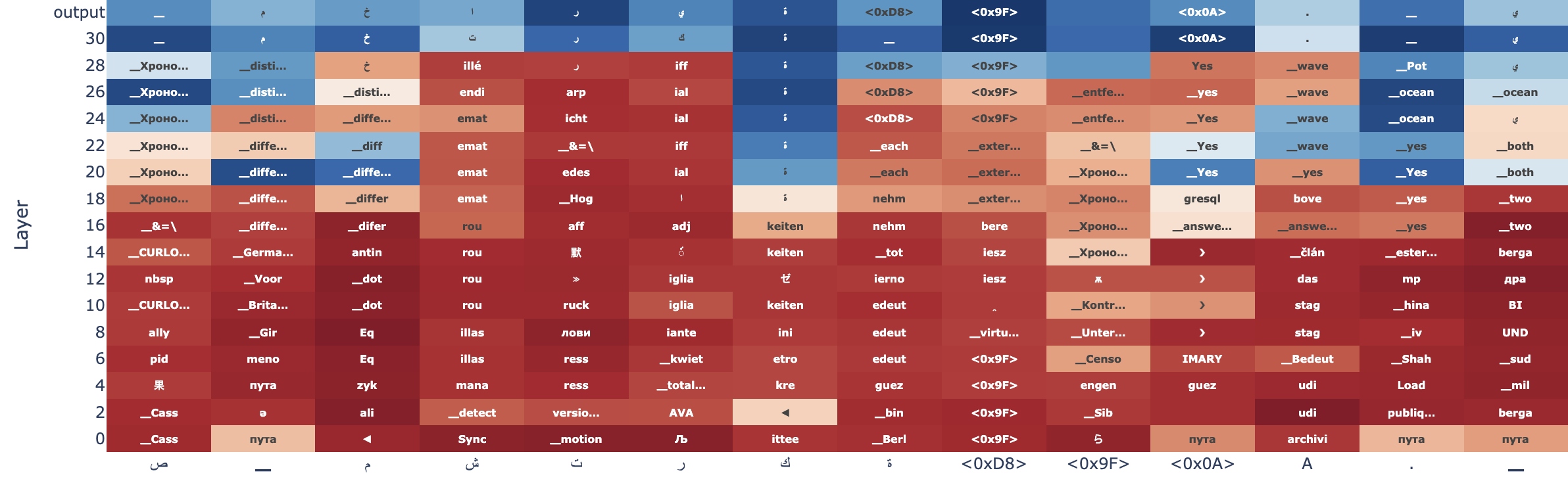}
    \label{fig:f13103}
    
\end{figure*}

\begin{figure*}[t!]
    \centering
    \includegraphics[width=\textwidth]{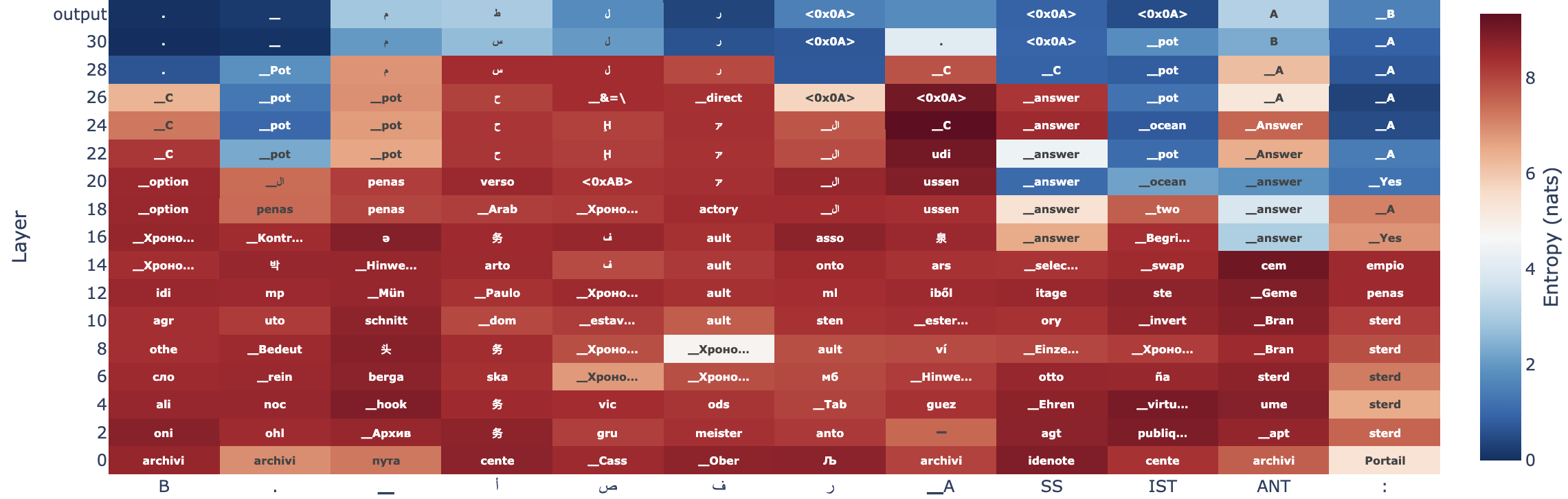}
      
     \caption{The complete visualization of next-token distributions in Arabic.}
     \label{fig:arabic} 
\end{figure*}

\begin{figure*}[t!]
    \centering
    \includegraphics[width=\textwidth]{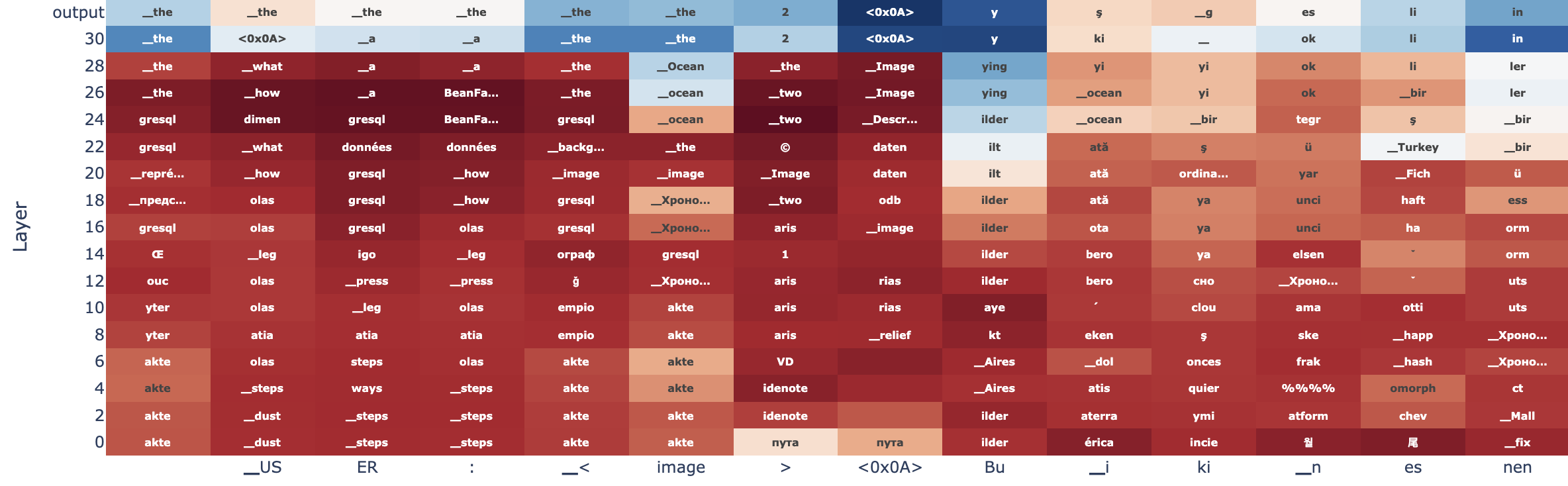}
    \label{fig:f13104}
    
\end{figure*}

\begin{figure*}[t!]
    \centering
    \includegraphics[width=\textwidth]{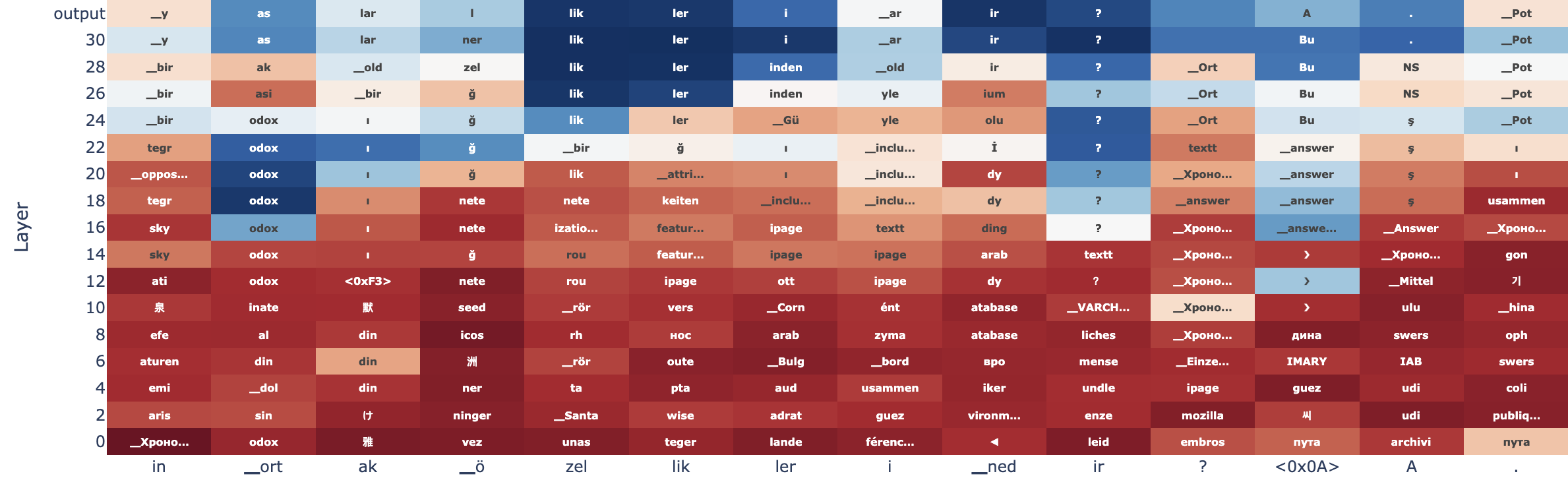}
    \label{fig:f13105}
    
\end{figure*}

\begin{figure*}[t!]
    \centering
    \includegraphics[width=\textwidth]{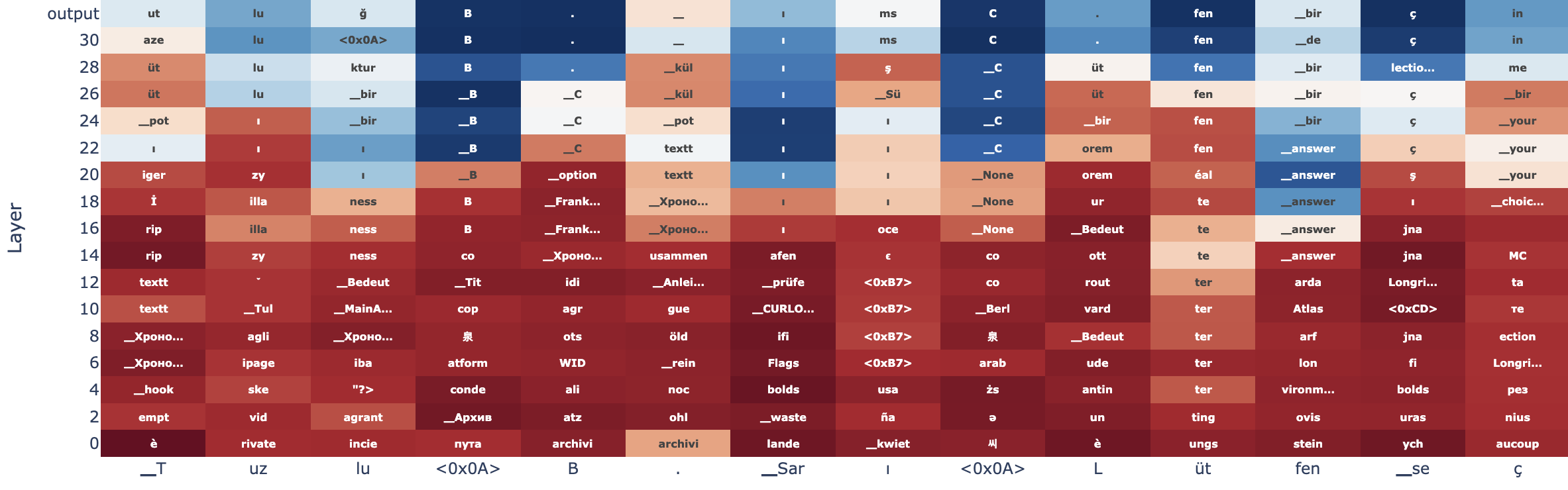}
    \label{fig:f13106}
    
\end{figure*}

\begin{figure*}[t!]
    \centering
    \includegraphics[width=\textwidth]{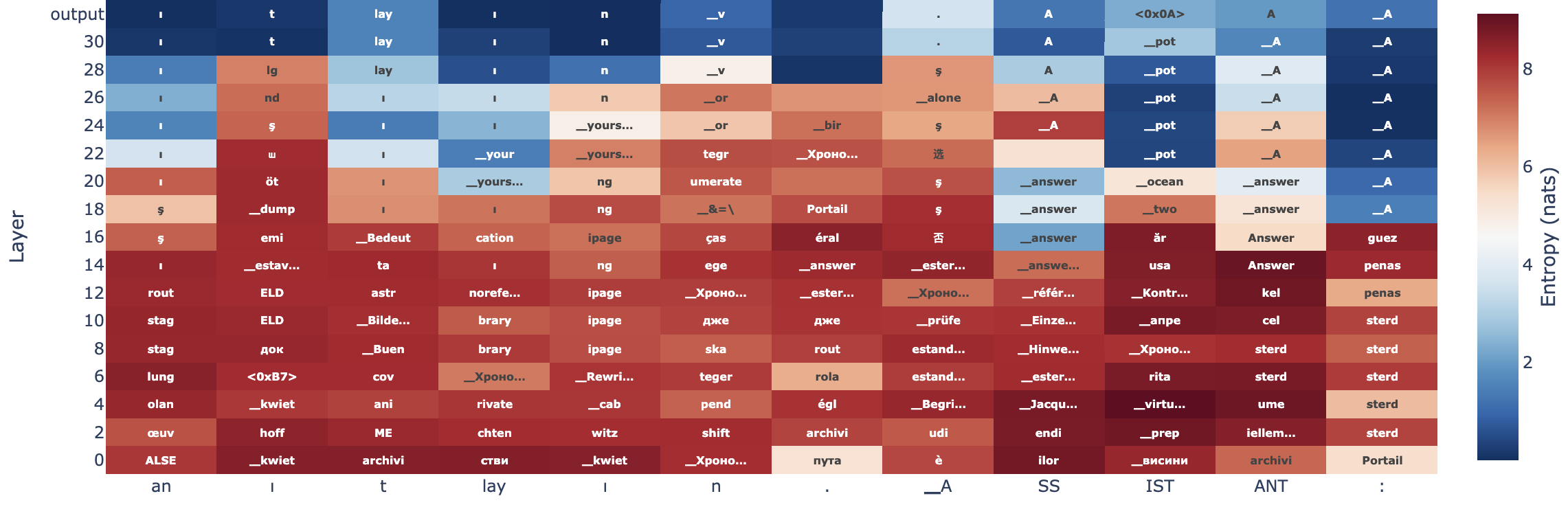}

     \caption{The complete visualization of next-token distributions in Turkish.}
     \label{fig:turkish}
\end{figure*}

\begin{figure*}[t!]
    \centering
    \includegraphics[width=\textwidth]{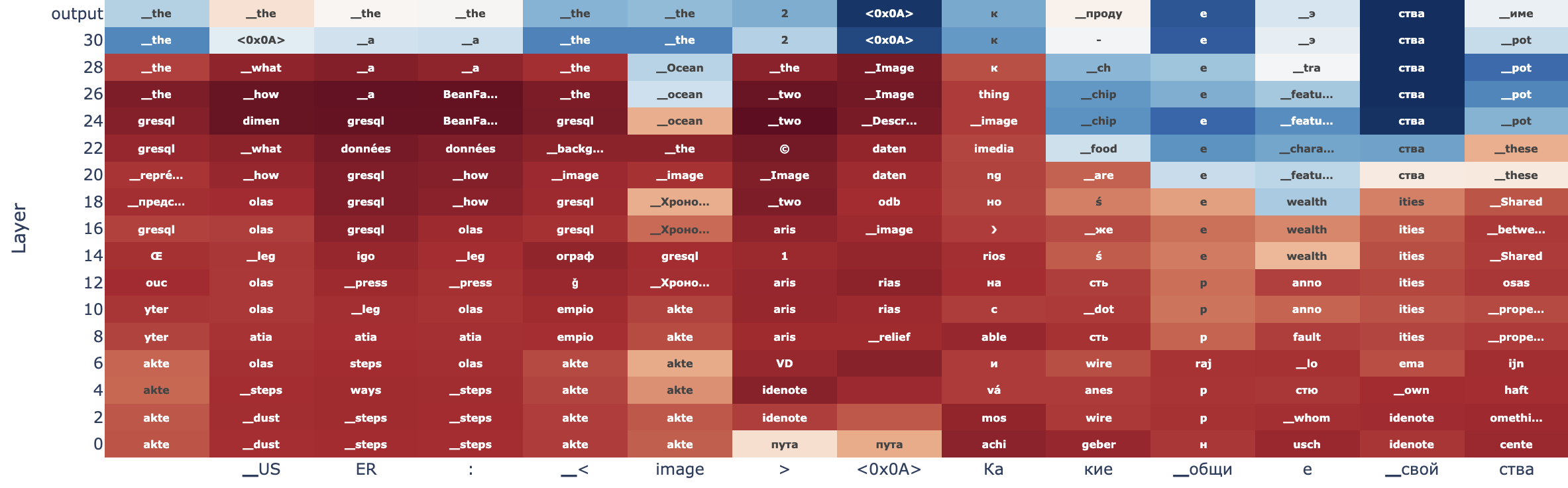}
    \label{fig:f13107}
    
\end{figure*}

\begin{figure*}[t!]
    \centering
    \includegraphics[width=\textwidth]{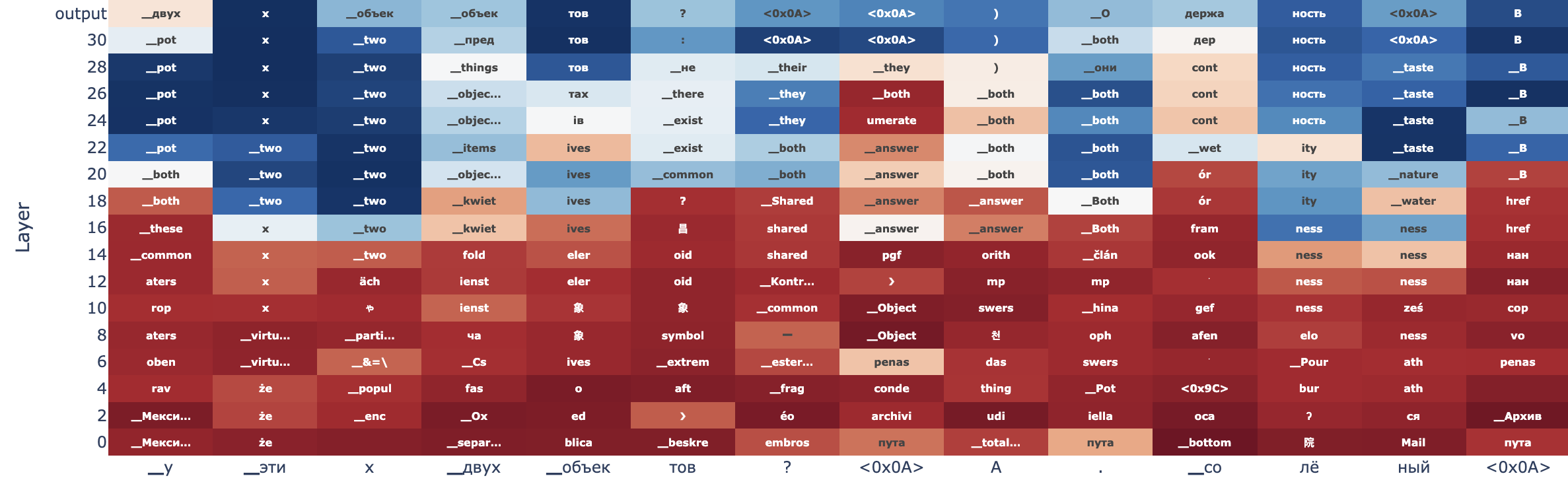}
    \label{fig:f13108}
    
\end{figure*}

\begin{figure*}[t!]
    \centering
    \includegraphics[width=\textwidth]{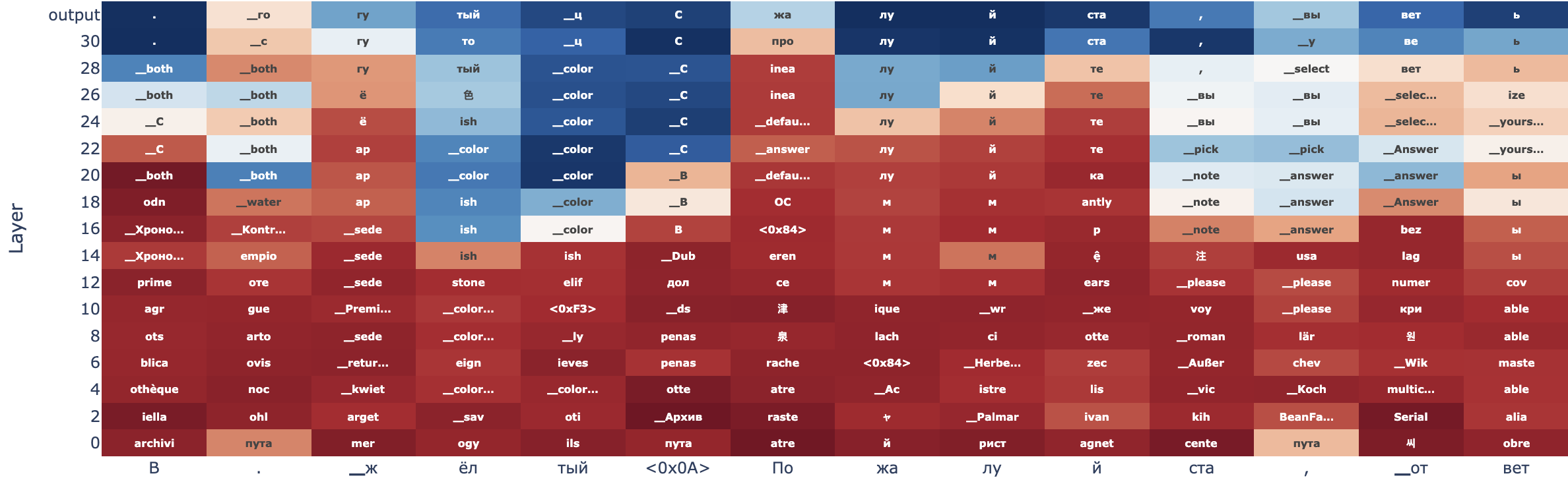}
    \label{fig:f13109}
    
\end{figure*}

\begin{figure*}[t!]
    \centering
    \includegraphics[width=\textwidth]{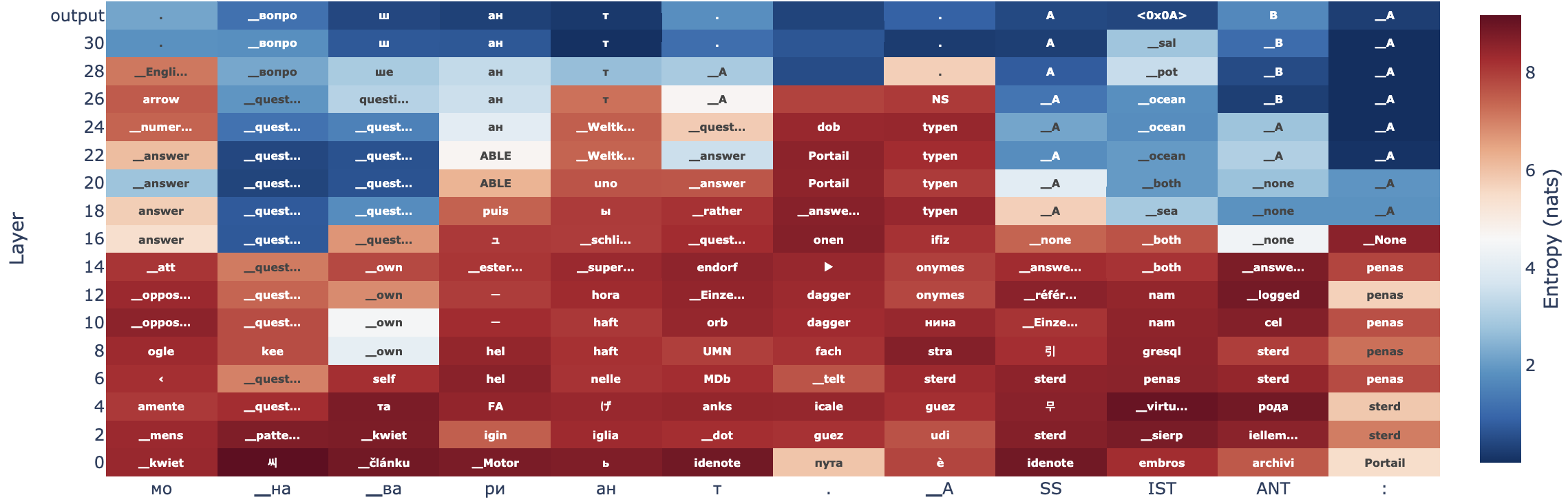}

     \caption{The complete visualization of next-token distributions in Russian.}
     \label{fig:russian}
\end{figure*}

\begin{figure*}[t!]
    \centering
    \includegraphics[width=\textwidth]{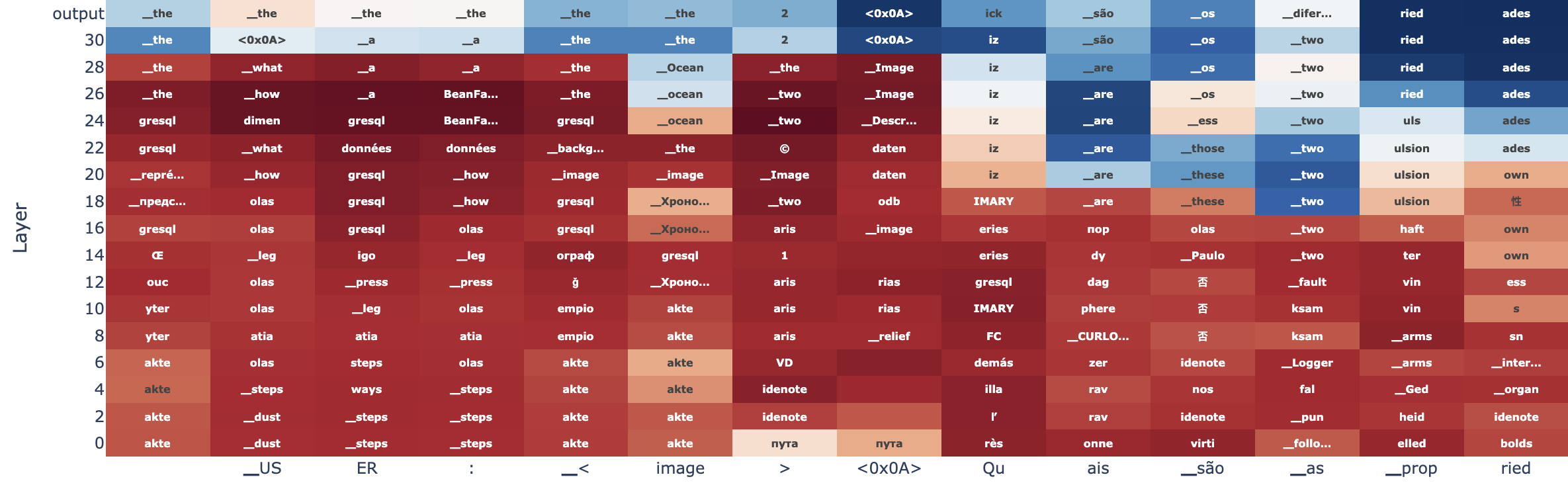}
    \label{fig:f131010}
    
\end{figure*}

\begin{figure*}[t!]
    \centering
    \includegraphics[width=\textwidth]{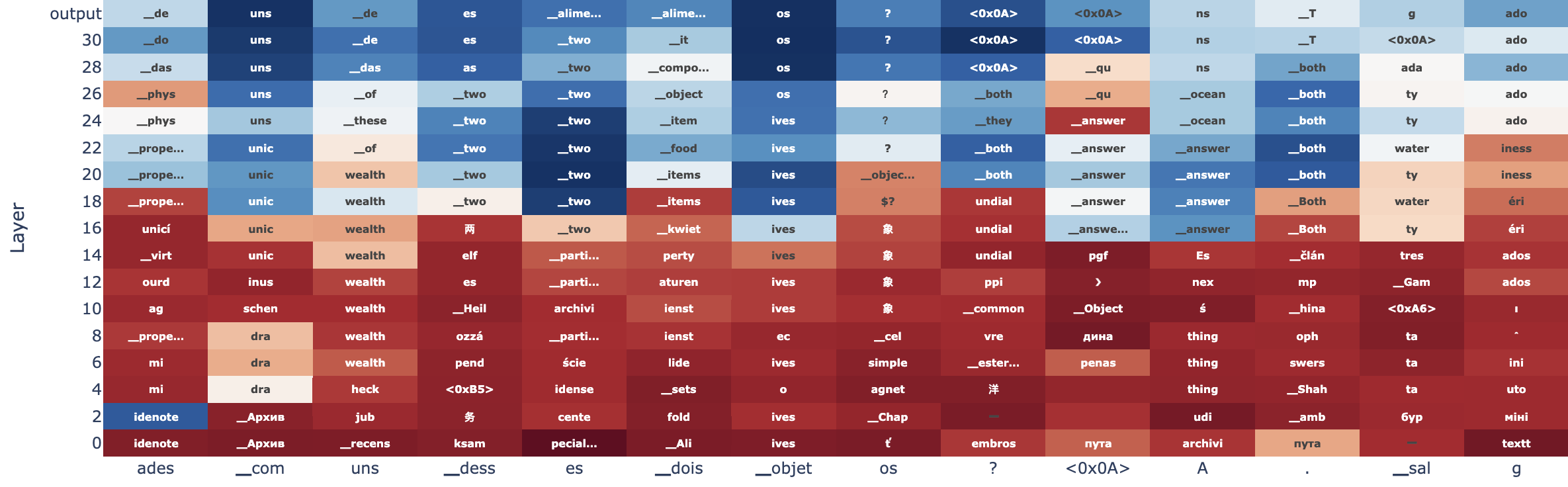}
    \label{fig:f131011}
    
\end{figure*}

\begin{figure*}[t!]
    \centering
    \includegraphics[width=\textwidth]{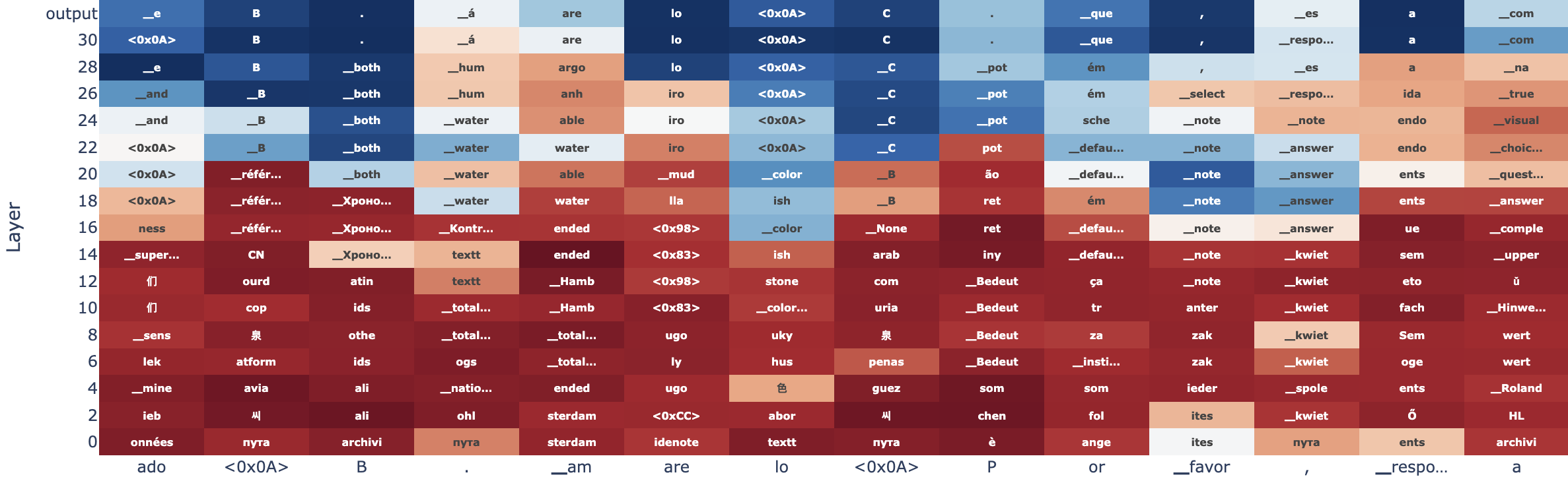}
    \label{fig:f131012}
    
\end{figure*}

\begin{figure*}[t!]
    \centering
    \includegraphics[width=\textwidth]{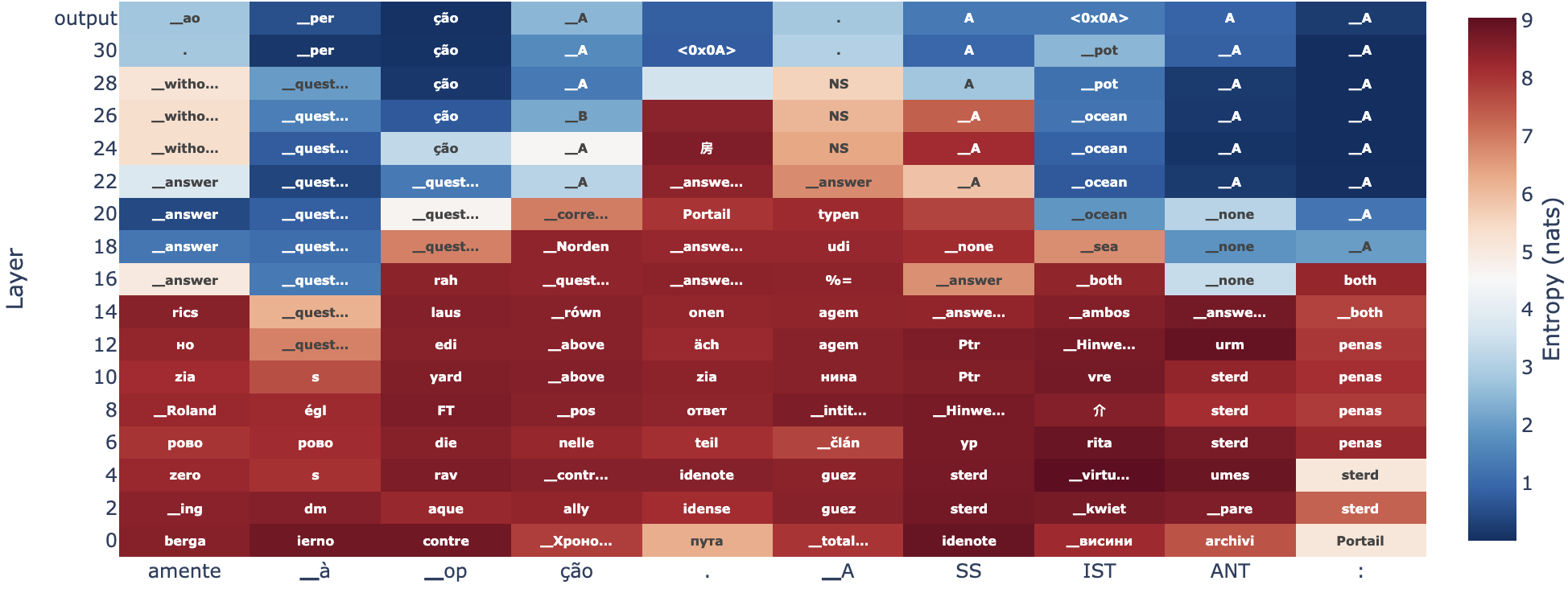}

     \caption{The complete visualization of next-token distributions in Portuguese.}
     \label{fig:portuguese}
\end{figure*}

\begin{figure*}[t!]
    \centering
    \includegraphics[width=\textwidth]{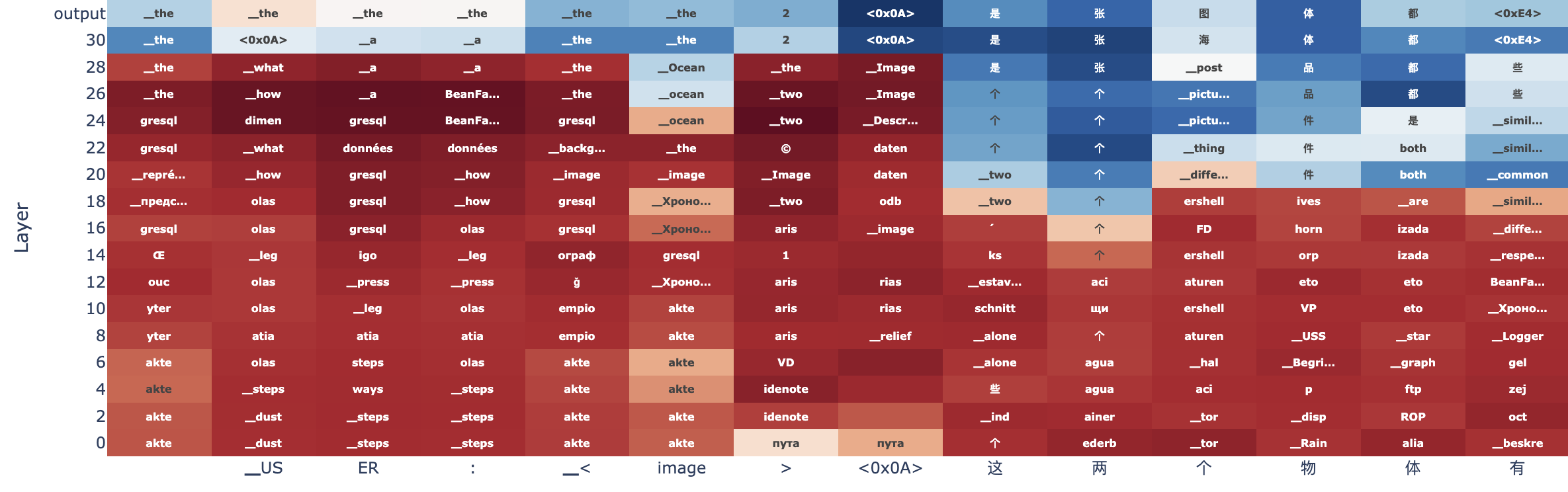}
    \label{fig:f131013}

\end{figure*}

\begin{figure*}[t!]
    \centering
    \includegraphics[width=\textwidth]{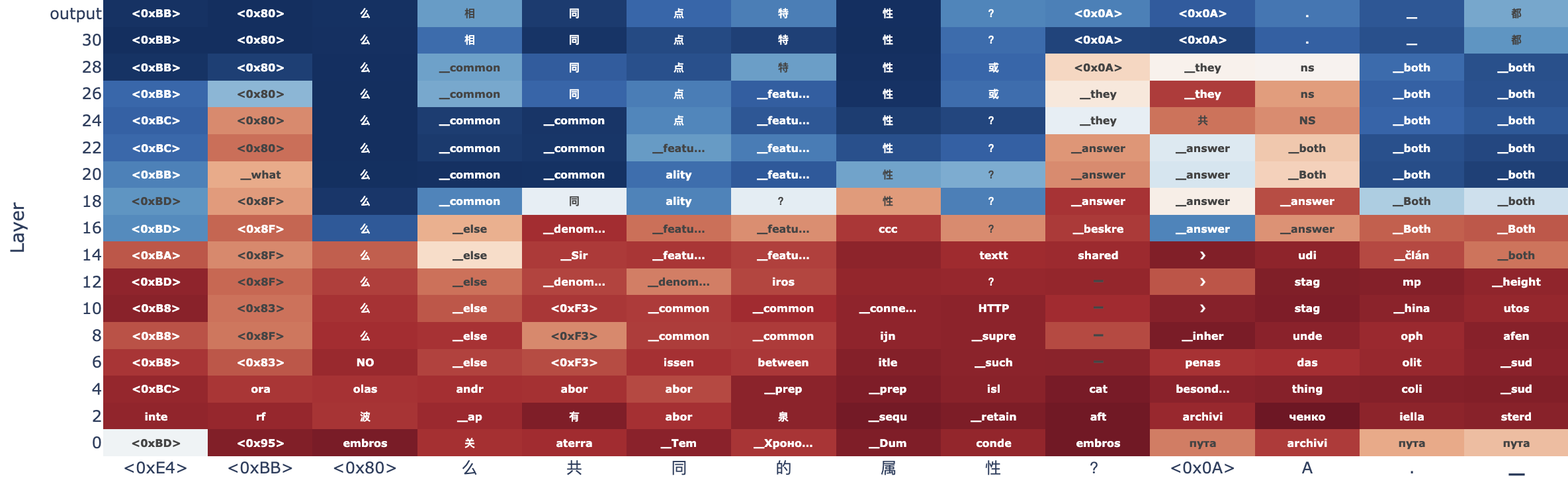}
    \label{fig:f131014}
    
\end{figure*}

\begin{figure*}[t!]
    \centering
    \includegraphics[width=\textwidth]{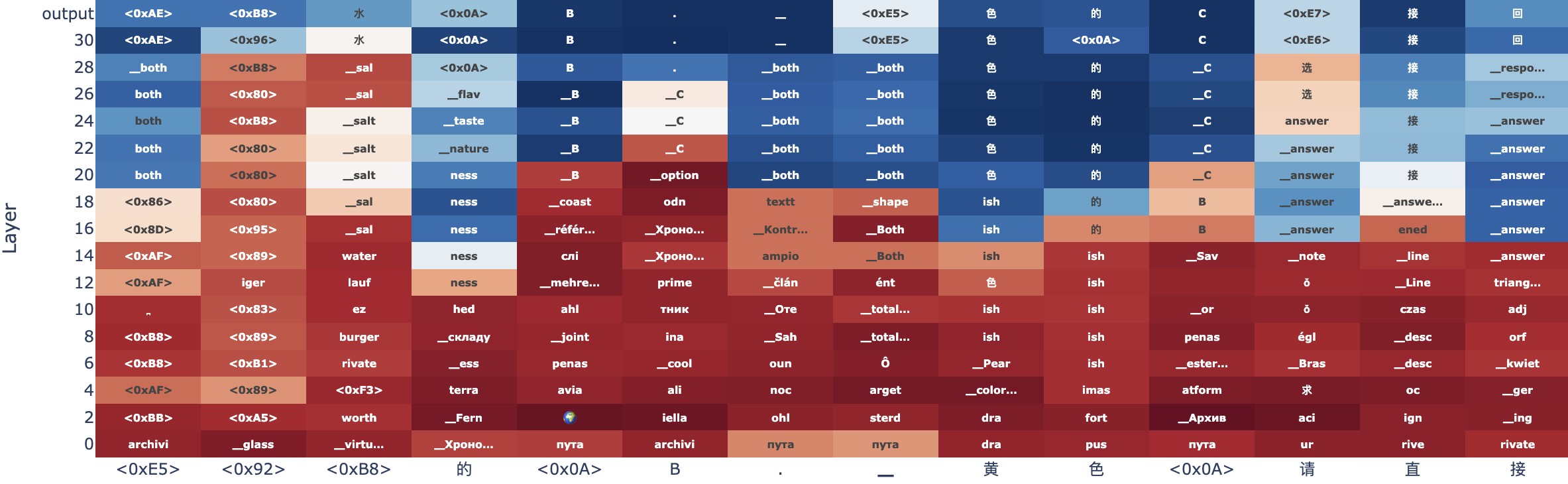}
    \label{fig:f131015}
    
\end{figure*}

\begin{figure*}[t!]
    \centering
    \includegraphics[width=\textwidth]{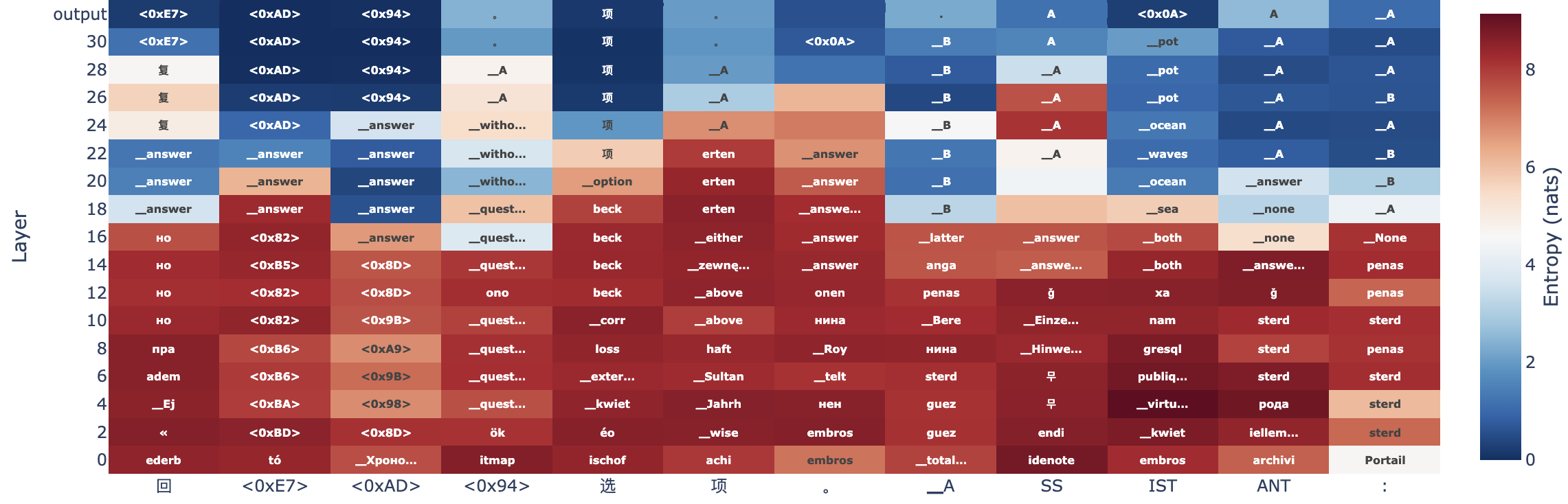}

     \caption{The complete visualization of next-token distributions in Chinese.}
     \label{fig:chinese}
\end{figure*}

\begin{figure*}[t!]
    \centering
    \input{figures/1.5-7B-LLaVA-Overlap-neuron}

     \caption{The overlap ratio between non-English and English activated neurons and the normalized number of activated
neurons across non-English languages in the LLaVA-1.5-7B model.}
\label{1.5-7B}
\end{figure*}

\begin{figure*}[t!]
    \centering
    \input{figures/1.5-13B-LLaVA-Overlap-neuron}

    \caption{The overlap ratio between non-English and English activated neurons and the normalized number of activated
neurons across non-English languages in the LLaVA-1.5-13B model.}
    \label{1.5-13B}
\end{figure*}

\begin{figure*}[t!]
    \centering
    \input{figures/1.6-7B-LLaVA-Overlap-neuron}

    \caption{The overlap ratio between non-English and English activated neurons and the normalized number of activated
neurons across non-English languages in the LLaVA-1.6-7B model.}
    \label{fig:1.6-7B}
\end{figure*}

\begin{figure*}[t!]
    \centering
    \input{figures/1.6-13B-LLaVA-Overlap-neuron}

    \caption{The overlap ratio between non-English and English activated neurons and the normalized number of activated
neurons across non-English languages in the LLaVA-1.6-13B model.}
    \label{fig:1.6-13B}
\end{figure*}

\begin{figure*}[t!]
    \centering
    \input{figures/Qwen-VL-chat-Overlap-neuron}

    \caption{The overlap ratio between non-English and English activated neurons and the normalized number of activated
neurons across non-English languages in the Qwen-VL-Chat model.}
    \label{fig:Qwen-VLchat}
\end{figure*}

\end{document}